\documentclass[a4paper,12pt,numbered,print,index]{Classes/PhDThesisPSnPDF}
\usepackage{afterpage}
\usepackage{epigraph}
\usepackage{adjustbox}
\usepackage{graphicx}
\usepackage{listings}
\usepackage{listings}

\newcommand\blankpage{%
    \null
    \thispagestyle{empty}%
    \addtocounter{page}{-1}
    \newpage}

\input{preamble}

\begin{document}

\frontmatter

\begin{titlepage}
\begin{center}

\vspace*{\fill}
\centering

{\Huge{An Embarrassingly Pragmatic Introduction to Vision-based Autonomous Robots}}\\[1cm]
\LARGE {Applications, Datasets and State of the Art}\\[3cm]

\Large {Marcos V. Conde}\\[0.5cm]
\Large {December 2021}

\vfill

\end{center}
\end{titlepage}

\begin{dedication} 

\vfill
\epigraph{I learned very early the difference between knowing the name of something and knowing something.}{\textit{Richard P. Feynman}}

\vfill
\epigraph{To develop the skills of correct thinking it is first and foremost necessary to learn what to disregard.}{\textit{Kurt Gödel}}

\vfill
\epigraph{I am thankful for all of those who said NO to me. It's because of them I'm doing it myself.}{\textit{Albert Einstein}}

\end{dedication}
\begin{abstract}

Autonomous robots are currently one of the most popular Artificial Intelligence problems, having experienced significant advances in the last decade, from Self-driving cars and humanoids to delivery robots and drones.\\

Part of the problem is to get a robot to emulate the perception of human beings, our sense of sight, replacing the eyes with cameras and the brain with mathematical models such as Neural Networks. Developing an AI able to drive a car without human intervention and a small robot to deliver packages in the city may seem like different problems, nevertheless from the point of view of perception and vision, both problems have several similarities. \\


The main solutions we currently find focus on the environment perception through visual information using Computer Vision techniques, Machine Learning, and various algorithms to make the robot understand the environment or scene, move, adapt its trajectory and perform its tasks (maintenance, exploration, etc.) without the need for human intervention.\\

In this work, we develop a small-scale autonomous vehicle from scratch, capable of understanding the scene using only visual information, navigating through industrial environments, detecting people and obstacles, or performing simple maintenance tasks. We review the state-of-the-art and demonstrate that many of our approaches are similar to the ones employed in real Self-driving cars from companies like Tesla or Lyft. Also, we study possible solutions to COVID-19 situations in Industry.\\

Finally, we discuss the current state of Robotics and autonomous driving and the technological and ethical limitations that we can find in this field.\\

\end{abstract}



\tableofcontents

{\small
\listoffigures
}

\listoftables



\mainmatter


\chapter{Introduction}  
\label{introduction}

\ifpdf
    \graphicspath{{Chapter1/Figs/Raster/}{Chapter1/Figs/PDF/}{Chapter1/Figs/}}
\else
    \graphicspath{{Chapter1/Figs/Vector/}{Chapter1/Figs/}}
\fi




Self-Driving cars imply an AI revolution, to understand and model human perception while doing a highly skilled task as driving is one of the most challenging problems of the decade, and is not even close to be solved. Moreover, this AI application opens the debate about Ethical implications of AI systems and the Human-AI interaction.
However, other autonomous vehicles (AV) or also referred in this project as Autonomous Robots (AR) are more feasible projects with direct real-world impact and easier Human-AI behaviour understanding (less Ethical implications).
These can be seen as a simplification of the complete autonomous vehicle since the domain of the robot (size, tasks, impact, ethical implications) is simpler and ''smaller``. In our opinion this argument is true, yet many technologies and solutions are essentially the same.\\
In this project we aim to prove some similarities between real-scale and small-scale autonomous robots, by developing a small AR that performs essential tasks. Additionally, we give the robot useful technologies for industrial environments under COVID-19 regulations.

\section{Related Work} 

\begin{figure}[htbp!] 
\centering    
\includegraphics[width=1.0\textwidth]{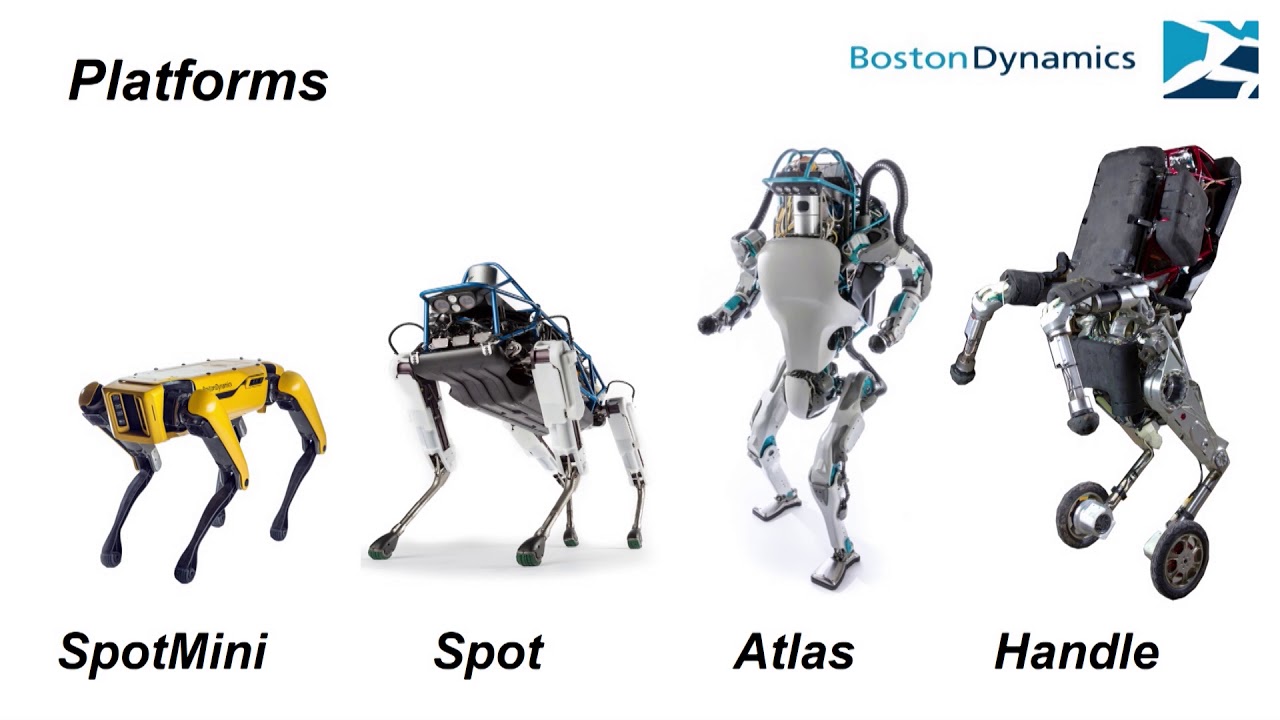}
\caption[Boston Dynamics Robots]{Boston Dynamics Autonomous Robots.}
\label{fig:boston-dyn}
\end{figure}

\begin{figure}[htbp!] 
\centering    
\includegraphics[width=1.0\textwidth]{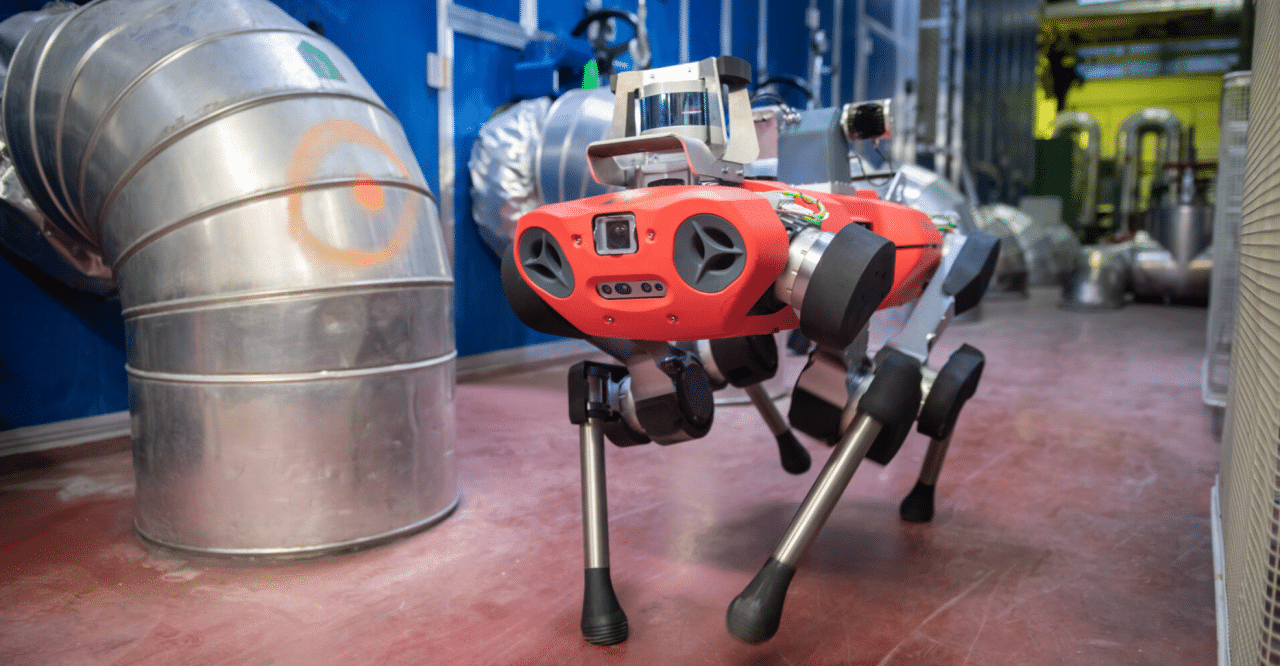}
\caption[ANYmal]{ANYmal from ANYbotics (ETH) using Nvidia Jetson AGX Xavier GPU.}
\label{fig:anymal-nvidia}
\end{figure}

If we talk about Autonomous Robots, the first in the list is Boston Dynamics project \ref{fig:boston-dyn}. These robots perform extreme difficult tasks that imply a deep understanding of human perception and motion: autonomous navigation through industrial plants, open doors or pick up objects, collaborate with other robots, acrobatic movements, etc. Our main inspiration is the \textit{SpotMini} \cite{spotmini}, who has proved its positive impact in real-world industry.
Another similar robot is ANYmal (ANYbotics, ETH Zurich), shown in Figure \ref{fig:anymal-nvidia}. This quadruped robot uses the Nvidia Jetson AGX Xavier board as embedded computer, it can explore and map factories, navigate autonomously, perform monitoring and maintenance tasks, etc. 

\begin{figure}[htbp!] 
\centering    
\includegraphics[width=1.0\textwidth]{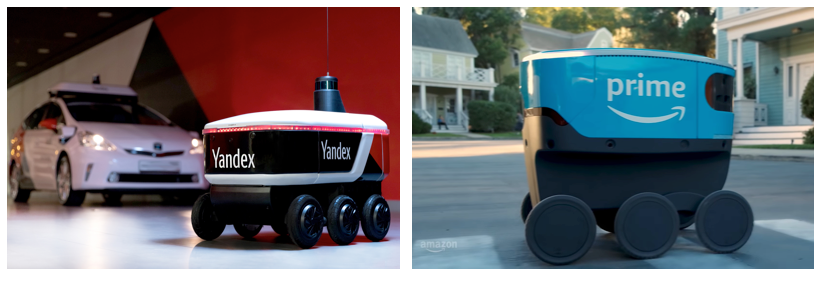}
\caption[Yandex - Amazon Delivery Robots]{(Left) Yandex self-driving car and delivery robot. (Right) Amazon delivery robot.}
\label{fig:delivery}
\end{figure}

Besides complete and multi-task ARs, some companies like Amazon and Yandex are developing Delivery Robots (see Figure \ref{fig:delivery}), their task is simpler, to transport food or objects from point A (local store) to a point B (delivery location), therefore, these sort of robot require mainly navigation and perception systems (path planning, avoid pedestrians, etc.). 
We take these robots as realistic and feasible baselines since their mechanical solution, the robot itself, is simple and easy to build.

\begin{figure}[htbp!] 
\centering    
\includegraphics[width=0.8\textwidth]{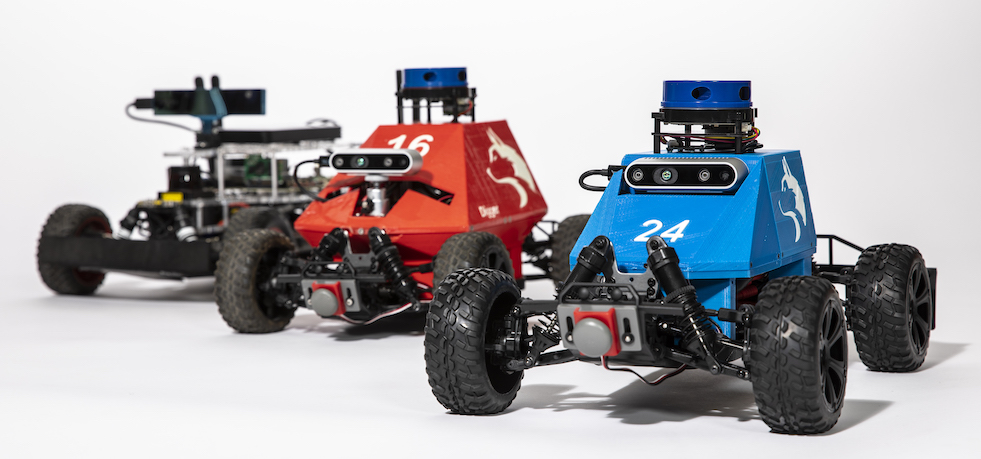}
\caption[MuSHR]{Left to right: MIT RACECAR-inspired, MuSHR V2-V3. Source: \url{https://mushr.io/}}
\label{fig:mushr}
\end{figure}

In Figure \ref{fig:delivery} (left), we can see on top of the robot an antenna (black color), this is sensor is called LiDAR and is an essential sensor for 3D perception and mapping used by many autonomous robots (including self-driving cars). In Figures \ref{fig:anymal-nvidia} , \ref{fig:boston-dyn} and \ref{fig:mushr} we can see bigger variants of this sensor on top of the robots.

\begin{figure}[htbp!] 
\centering    
\includegraphics[width=0.9\textwidth]{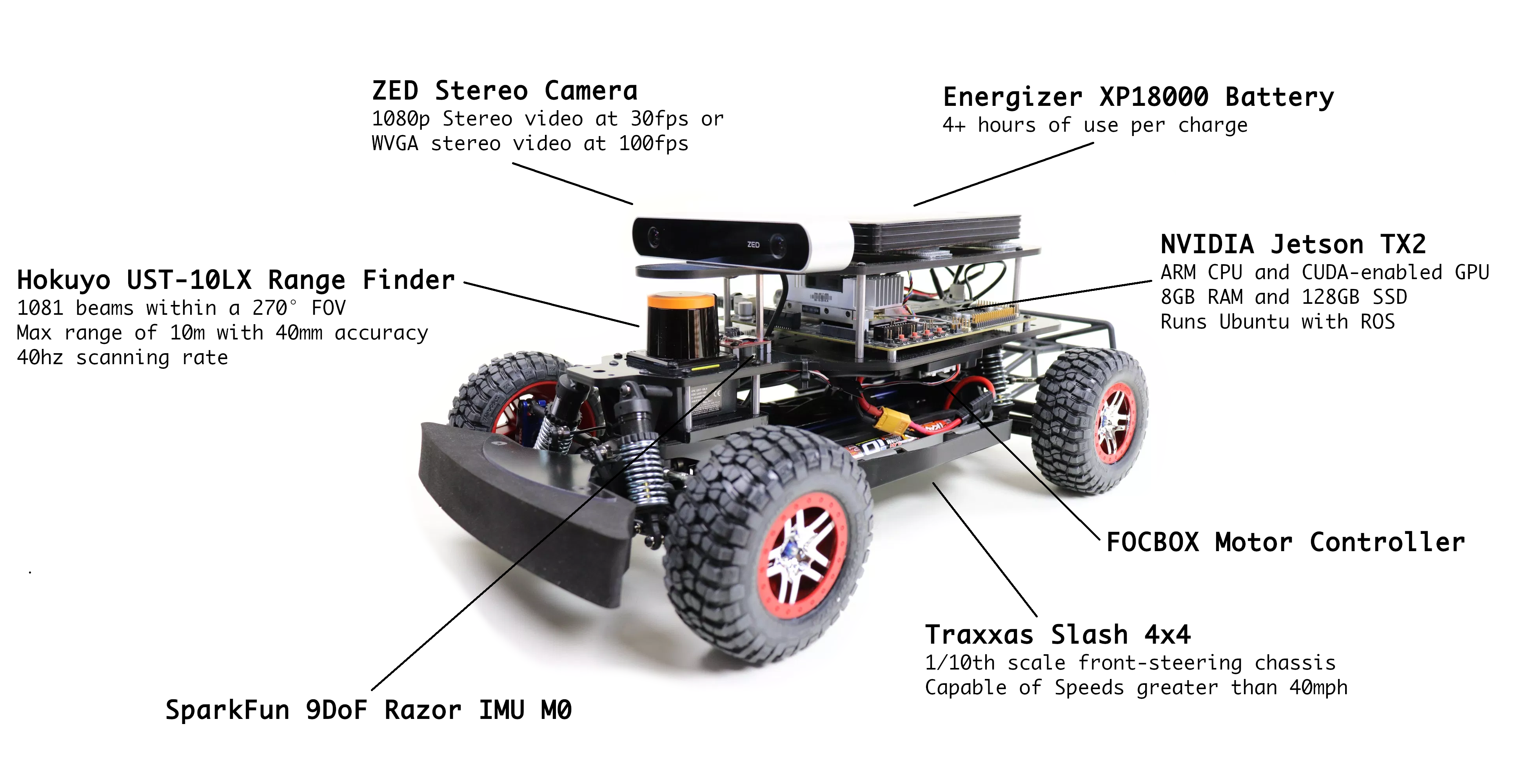}
\caption[MIT RACECAR]{MIT RACECAR $ICRA$ 2019}
\label{fig:mit-racecar}
\end{figure}

We aim to extend the idea of easy-to-build and practical Delivery Robots by improving its multi-task capabilities. 
The \emph{MIT RACECAR} \cite{mit-racecar-icra-2019} is an open-source powerful platform for robotics research and education, the platform houses state-of-the-art sensors and computing hardware, placed on top of a powerful 1/10-scale mini race car. The research time organized a workshop at the International Conference on Robotics and Automation (IEEE ICRA) 2019, which serves as introduction to this project and tutorial. 
The MIT Racecar Model-N uses the NVIDIA Jetson Nano board as embedded computer allowing it to run complex deep learning models in real-time applications (e.g. detect persons) \cite{zhang2020empowering-jetson}. 

The \emph{MuSHR} (Multi-agent System for non-Holonomic Racing) \cite{srinivasa2019mushr} is a low-cost, open-source robotic racecar platform for education and research, developed by the Personal Robotics Lab in the Paul G. Allen School of Computer Science \& Engineering at the University of Washington. 
Researchers designed and developed the robot taking inspiration from the MIT RACECAR platform \cite{mit-racecar-icra-2019}, aiming to create an affordable full-stack robotics system which not only can support our own research and teaching demands, but those of the community at-large, for this reason, MuSHR is considered a valuable asset for academic research labs, robotics instructors, and robotics enthusiasts.\\

Besides pure robotic solutions, there are research projects focused on solving real-world Autonomous Driving problems using small-scale simulations. The \emph{Duckietown project} \cite{duckitown, duckietown-icra} was conceived in 2016 at MIT, its goal was to build a platform that was small-scale and cute yet still preserved the real scientific challenges inherent in a full-scale real autonomous robots and self-driving cars \cite{zilly2019ai-nips-ducky, tani2020integrated-ducky, sazanovich2020imitation-ducky}. This platform has been used at several universities around the world: NCTU in Taiwan, Tsinghua in China, Université de Montréal, TTI Chicago, and ETH Zürich among many others.

The Duckietown Foundation \cite{duckitown} organized the AI Driving Olympics (AI-DO) \cite{zilly2019ai-nips-ducky}, the first competition focused on AI for self-driving cars with real robots, in 2018 at the Conference on Neural Information Processing Systems conference (NeurIPS) in Montreal. The scond edition of this challenge, AI-DO2, was held at the International Conference on Robotics and Automation (ICRA) in 2019.

\begin{figure}[htbp!] 
\centering    
\includegraphics[width=1.0\textwidth]{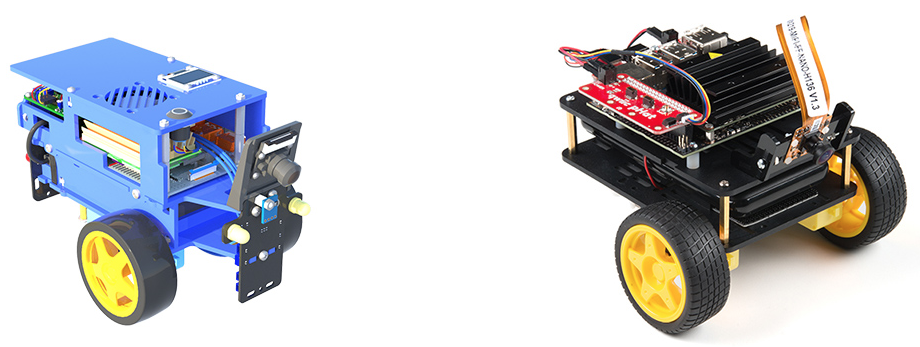}
\caption[Nvidia Open Source Educational robots ]{(Left) Duckiebot. (Right) Nvidia JetBot.}
\label{fig:jetbots}
\end{figure}

\begin{figure}[htbp!] 
\centering    
\includegraphics[width=1.0\textwidth]{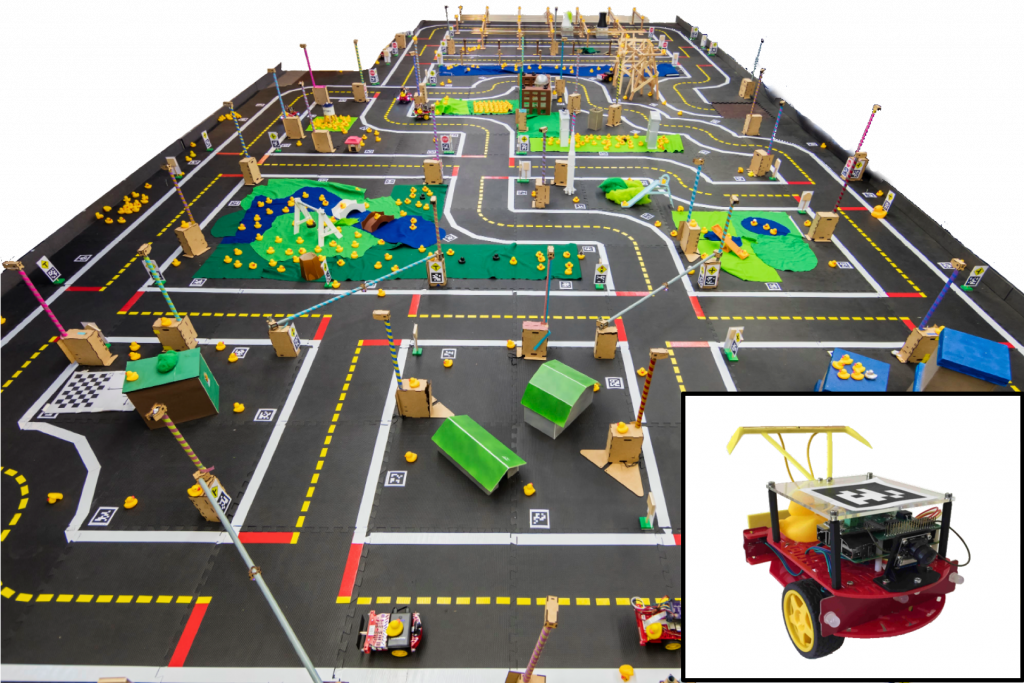}
\caption[Duckietown Lab]{Duckietown Autolab. Source \url{https://www.duckietown.org/}.}
\label{fig:duckietown}
\end{figure}

\subsection*{Our approach: JetRover}

\begin{figure}[htbp!] 
\centering    
\includegraphics[width=1.0\textwidth]{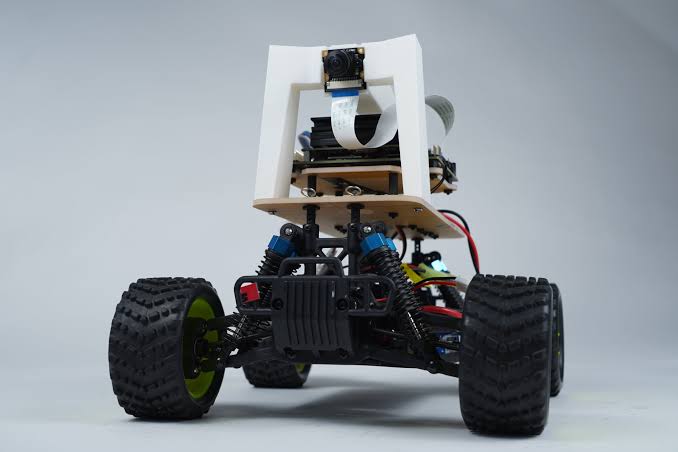}
\caption[JetRover (Donkey Car)]{Donkey Car powered by the NVIDIA Jetson Nano board. Source ''Robocar Store``.}
\label{fig:our-jetrover}
\end{figure}

After exploring the related work in AI robotics, we decide to build our robot inspired by MIT RACECAR \cite{mit-racecar-icra-2019} and MuSHR \cite{srinivasa2019mushr}. The complete robot costs less than 250\$ and is easy to build (less than 2hr). This robot allows us to experiment the same technologies as the Delivery Robots, expand its devices and sensors, and use it for other technical tasks as MuSHR and ANYmal (mapping, object detection, etc.) or other industrial AGV (Automatic Guided Vehicles). We use the Jetson Nano board, a powerful computer that allow us to run multiple neural networks in parallel for applications like image classification, object detection, segmentation, and speech processing. Moreover, NVIDIA provides an open-sourced software called ''JetPack SDK``, which includes the latest APIs for AI Edge application development.

Additionally, we also developed JetBot (see Figure \ref{fig:jetbots}), and educational open-source robot based on Jetson Nano that includes a set of tutorials which cover basic robotics concepts like programatic motor control, to more advanced topics like training a custom AI for avoiding collisions.
More information about the robots, hardware and cost is provided in the Appendix \ref{app:robot}.

\paragraph{What is a vision-based agent?}

In the context of artificial intelligence, we can define a rational agents or intelligent agent as an autonomous entity which acts and makes decisions directing its activity towards achieving goals. Their actions are constrained to an environment, and guided by observation through sensors, their perception. Reinforcement Learning study how agents can learn or use knowledge to achieve their goals, and the impact of past actions.
The environment may have an inherent uncertainty and variability, moreover, many agents may interact in the environment, each one trying to do a different action or achieve different goals, this is known as a multi-agent scenario. In this project, we assume the inherent uncertainty that autonomous robots (as intelligent agents) may find in industrial environments, however, we limit the scope of this project to single agent operating in the environment.

Russell \textit{et.al.} \cite{ai-modern-approach-russell} groups intelligent agents into five types based on their degree of perceived intelligence and capabilities:
\begin{enumerate}

    \item Simple reflex agents: They act only on the basis of the current perception-status, based on condition-action rules, and ignoring the past and history.
    
    \item Goal-based agents: These agents take decision considering specific goals (desirable situations), every action tries to achieve such goal or minimize the ''distance`` to it, usually these agents are reward if they achieve their goals. In the case where multiple actions are possible, the agent acts consequently maximizing the probability of achieving the goal.
    
    \item Utility-based agents: Agents based on uses, not goals. Sometimes achieving the desired goal is not enough, in cases where the states are also important, we can define a ''utility function`` that measures how desirable a particular state is. These agents choose actions based on a preference (utility) for each state, also known as agent's ''happiness``.

    \item Model-based reflex agents: These agents require a model about the world or environment. They work by finding a rule whose condition matches the current situation, and thus, they can handle partially observable environments. The agent remembers internal states (past history) that are used to adjust current-future actions.
    
    \item Learning agents: These agents can learn from its past experiences or online, meaning they have active learning capabilities. They start to act with basic knowledge and then they are able to act and adapt automatically through learning, this has the advantage that it allows the agents to initially operate in unknown environments and to improve consistently.
\end{enumerate}

Our robots are then, \textit{Model-based reflex agents}, however, rules are learned by a machine learning model with little prior knowledge about the environment, therefore, the agents are also \textit{Learning agents}.
Moreover, in Reinforcement Learning \cite{Fran_ois_Lavet_2018_intro_rl}, agents are usually trained to optimize a certain task, so many agents are also considered \textit{Goal-based agents} \cite{florensa2018automatic-goal-rl}. 

\begin{figure}[]
\includegraphics[width=\textwidth]{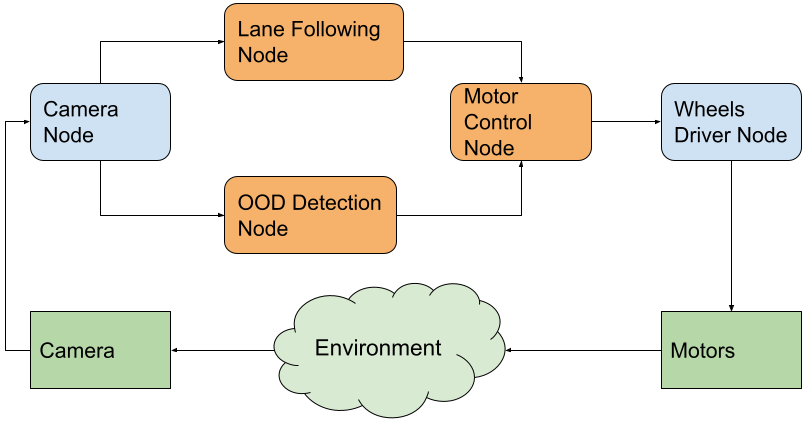}
\centering
\caption{Example of visual-based agent from Duckietown OOD detector project.}
\label{fig:duckie-agent-diagram}
\end{figure}

In the context of Computer Vision and the scope of this project, a visual-based agent is an intelligent agent that makes decisions relaying on visual information (see Figure \ref{fig:duckie-agent-diagram}), and its main perception system are camera sensors.
This clarification is necessary since many autonomous driving systems combine information from different sensors, this process is called: Sensor Fusion \cite{sensor-fusion-ad}. For instance, besides camera sensors (monucular RGB cameras), the other main sensor is the LiDAR (Light Detection and Ranging or Laser Imaging Detection and Ranging), which allows real-time and high-quality 3D perception.
In Figures \ref{fig:anymal-nvidia}, \ref{fig:mushr} and \ref{fig:delivery} we can see on top of the robots a this device with cylindrical shape.

The most famous visual-based system, meaning, perception is based only on images (camera sensors) is Tesla Autopilot, this Full Self-Driving (FSD) system does not use pre-mapped data and LiDAR which makes scaling difficult, since vehicles’ autonomous capabilities are practically tied to a geofenced areas. Moreover, the LiDAR is very expensive.

\begin{figure}[]
\includegraphics[width=\textwidth]{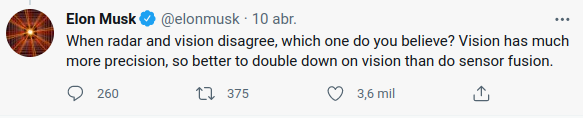}
\centering
\caption{Elon Musks's public comment about vision vs radar.}
\label{fig:elon-on-vision}
\end{figure}

\begin{figure}[]
\includegraphics[width=\textwidth]{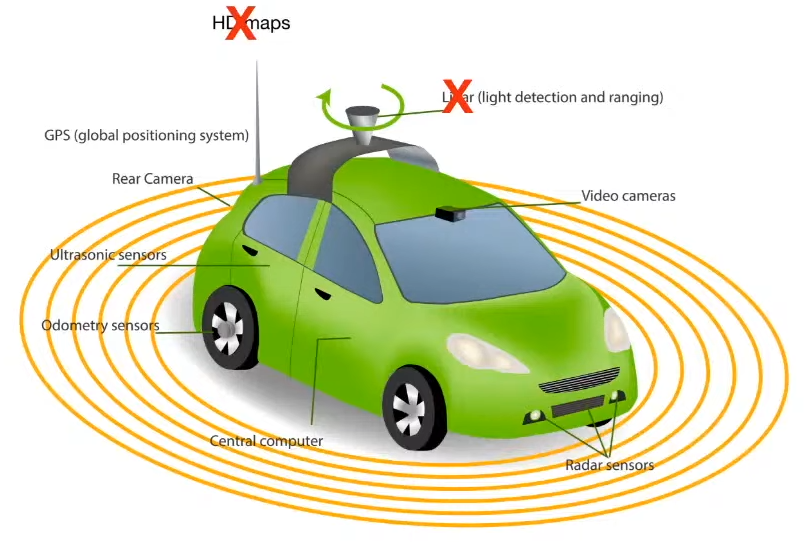}
\centering
\caption[New Tesla system relaying only on visual information (no radar)]{New Tesla system relaying only on visual information (no radar). Image from Andrej Karpathy's presentation at CVPR'21 WAD.}
\label{fig:tesla-no-lidar}
\end{figure}

\clearpage


\section{Motivation and Objectives}  

Industry 4.0 \cite{Hermann2015-industry40, Zhong2017-industry40} demands AI solutions for automatizing industrial processes like monitoring, maintenance, prognosis (prevent failures or accidents). However, there is a huge gap between research and industry-level implementation.
For instance, Object Detection is a well-known high-level computer vision task \cite{Jiao_2019-objectdetc-survey} and a hot research area in this field, yet it is difficult to use in the industry those technologies developed in laboratories, some of the reasons are: (i) models cannot infer fast enough, so they cannot be used in real-time or streaming applications, (ii) models fail to generalize or generalize but only under certain conditions, (iii) models are not robust against adversarial conditions (\textit{e.g.} sensor occlusion), noisy inputs, etc. In this project we aim to propose and solve simple (yet realistic) industrial applications implemented using a small-scale autonomous robot.
Furthermore, we want to find and prove the similarities between real-scale Autonomous Cars and small Autonomous Robots (SpotMini, JetBot, etc.) in terms of problem resolution, technologies, and deep learning solutions.\\

The main objective of this work is to understand the fundamental problems in computer vision that allow autonomous navigation and constitute a visual perception system. 
We explore the state-of-the-art for each technical problem, the related work, applications, and some implementations. This work focuses on the experimental and pragmatic aspects. 
Thus, our work aims to achieve the following specific objectives:
    

\begin{enumerate}

    
    \item Develop a starting \emph{low-cost} prototype robotic solution (JetRover) for basic industry-level tasks, based on computer vision and deep learning applications. We test this robot in real environment.
    
    \item Describe the technical similarities between real autonomous vehicles and small-scale robots from the perception point of view.
    
    \item Study possible solutions to COVID-19 situations in Industry, example use cases: detect and count persons in closed areas like a factory plant, and detect if persons wear facemasks. These applications can be used under ``normal'' situations too, and bring a huge value to an Industry 4.0 environment. We also participate in the ``Face Bio-metrics under COVID (Masked Face Recognition Challenge \& Workshop ICCV 2021)''.
    
    \item Introduce briefly 3D perception and future applications like 3D scene reconstruction.
    
\end{enumerate}

Note that due to the time-cost constraints, other related tasks like Navigation (which is a research topic in itself) and safety are not covered in this work, and we have to use multiple pre-trained model.

\clearpage


\section{Planification and Costs}

\begin{figure}[]
\includegraphics[width=0.9\textwidth]{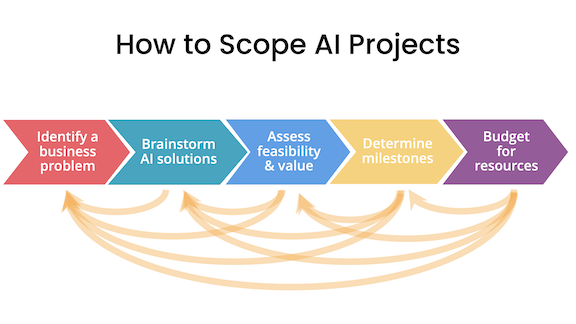}
\centering
\caption[Andrew Ng’s 5 Step Framework to Plan AI Projects Effectively]{Andrew Ng’s 5 Step Framework to Plan AI Projects Effectively \cite{andrewng-plan}.}
\label{fig:andrewng-plan}
\end{figure}

In this section we will cover the planification of the project following ''Andrew Ng's 5 Step Framework to Plan AI Projects Effectively`` \cite{andrewng-plan} (see Figure \ref{fig:andrewng-plan}). We understand AI projects are different from classical Software projects, therefore, the development process and planification is slightly different.
We will explain each step in the planification pipeline and how we develop deep learning models.
Finally, we describe the resources and cost of the project.

\begin{enumerate}
    \item \textbf{Identify a business problem (not an AI problem):} Control automatically certain conditions in factories: number of persons in the plant, social distancing, security (is the worker wearing a helmet?, is someone in a danger zone?), etc. These problems-solutions are part of an Industry 4.0 environment (AI integration into Industry).
    
    \item \textbf{Brainstorm AI solutions:} At the beginning of the project, we have 6 supervised brainstorming meetings to discuss the project's scope, ideas, robotic solutions, etc.
    After each meeting, milestones and feasibility are updated.
    
    \item \textbf{Assess the feasibility and value of potential solutions:} We determine that a solution based on related problems, and adapted to this \textit{Use case}, is a realistic and highly valuable starting point for real-world Industry 4.0 applications.
    
    \item \textbf{Determine milestones:} (i) Understand and propose a initial robotic solution based on existing solutions and state-of-the-art, (ii) solve at least one real-world industry problem, (iii) determine the scalability of the solution.
    
    \item \textbf{Budget for resources:} The initial budget is $250\$$ for building the robot, and $0\$$ for computing resources. We use open-sourced tools and free cloud resources (Kaggle and Google Collab). At the end of this section we describe the market cost of the resources and provide more details.
\end{enumerate}

\begin{figure}[!h]
\includegraphics[width=0.60\textwidth]{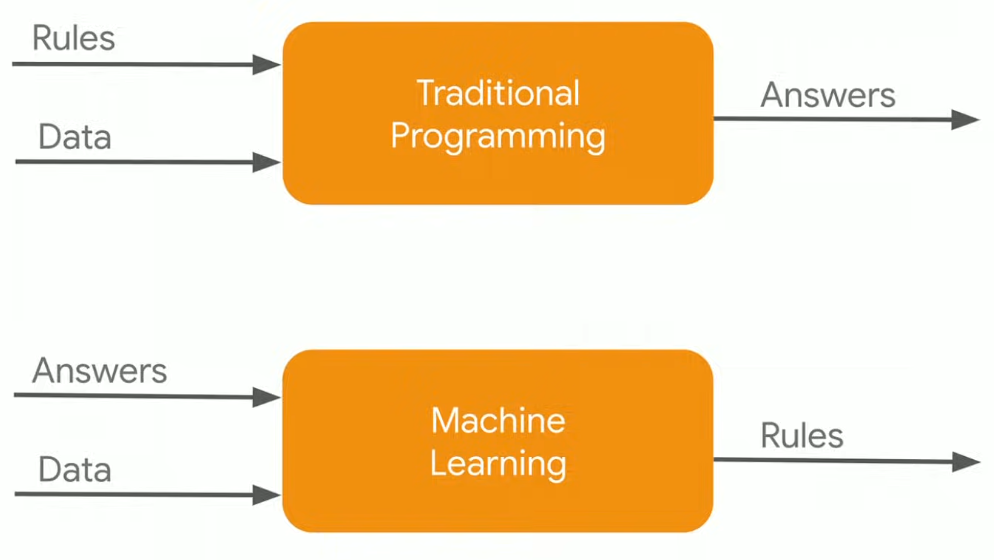}
\centering
\caption{''New`` paradigm of Machine Learning by Laurence Moroney.}
\label{fig:ai-vs-software}
\end{figure}

Note that in our opinion, and following the understanding of top ''AI voices`` like Andrew Ng and Laurence Moroney, AI projects are essentially different from classical Software projects. As we can see in Figure \ref{fig:ai-vs-software},
Traditional Programming express rules in a programming language like Python, Java, C++, these rules act on data, and we get answers (a sorted array, the output from a DB query, etc.). However, instead of expressing all these rules in code, Machine Learning assumes we have data and answers that we can label, and then the machine infers the rules.
This paradigm also affect traditional programming because recent work from \textit{Open AI} proves that specific rules represented as pure code (e.g. Python code) can be learned. An example is GitHub Copilot \cite{copilot}, an AI system that can learn to generate code and to assist developers.

\begin{figure}[]
\includegraphics[width=0.8\textwidth]{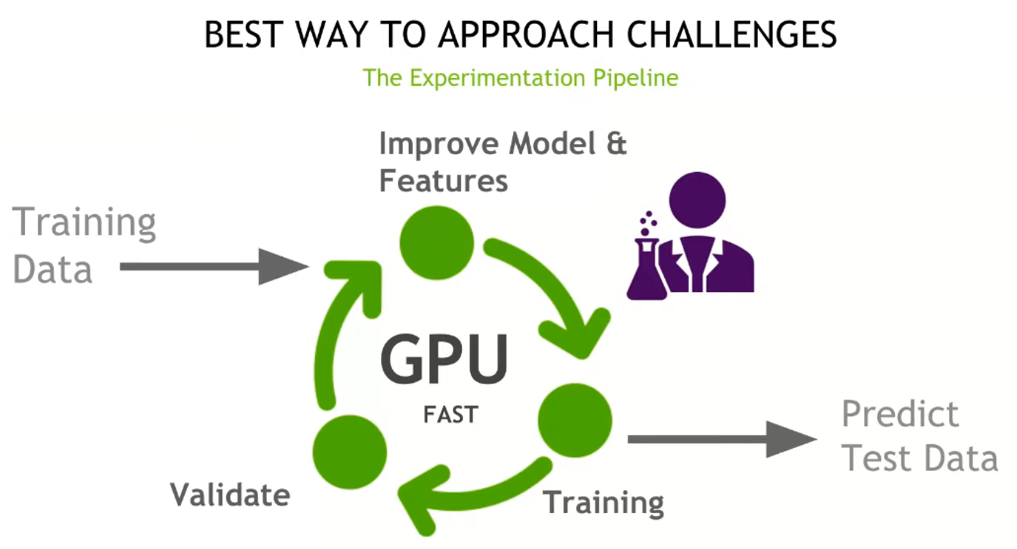}
\centering
\caption{Experimentation Pipeline from NVIDIA Kaggle Grandmaster Series.}
\label{fig:experiment}
\end{figure}

Also, a machine learning project requires training data (labeled answers), and the data quality highly determines the performance of the final solution. To find the best model that learns properly the rules is a very time-consuming and expensive task, and it is not directly linked to code, but to ideas. Therefore, it is very difficult to provide estimates regarding time and deliverables, and impossible without having a brief overview of the data and inputs.
Furthermore, a single iteration (Figure \ref{fig:experiment}) strictly depends on the results of the previous one, for this reason, detailed planning for a long time ahead does not make sense.


After determining the project's scope, we also determine a \textbf{flexible} time distribution of the main task, the deep learning solution development, so we can detect bottlenecks in the project. As shown in Table \ref{tab:time-dist} we establish two phases: 
\begin{enumerate}
    \item We develop the solution in a ''laboratory``: we find and explore datasets, prepare curated training data, study different evaluation metrics depending on the problem, train SOTA deep learning models for each problem, validate and test the models, etc. In Figure \ref{fig:experiment} we show the fast experimentation pipeline for developing deep learning solutions that we employ. The key idea is to run experiments as fast as possible, so we can try more models and ideas.
    
    \item We deploy the model created in the 'laboratory``, this means we need to integrate this AI into an industry environment. We use the NVIDIA Jetson Nano board as embedded computer to deploy these models and achieve the best performance in real-time applications \cite{zhang2020empowering-jetson, muller2021openbot-jetson}. We provide the complete information in Section \ref{sec:train-deploy}.
    
\end{enumerate}

\begin{table}[b]
    \caption{Time distribution for each task in the deep learning solution development cycle.}
    \centering
    \begin{adjustbox}{max width=\textwidth}
    \begin{tabular}{c c c}
         Task & \% of time & Description \\
         \toprule
         Dataset preparation   & 30 & Find public datasets or create a new one from scratch\\ 
         Experimentation cycle & 60 & Train-Validate-Test models (Figure \ref{fig:experiment})\\ 
         Interpretation        & 10 & Explore and interpret results, can we improve? \\ 
         \midrule
         \textbf{Phase 1}      & 100 & Find the best model-solution \\ 
         \toprule
         \toprule
         Deployment            & 40 & The best model is deployed (cloud, embedded system, etc.)\\ 
         Real-world test       & 60 & Implement the \textit{Use Case}, learn and improve (cycle)\\ 
         \midrule
         \textbf{Phase 2}      & 100 &  Test the solution in the industry, Human-AI interaction, etc. \\ 
         \bottomrule
    \end{tabular}
    \end{adjustbox}
    \label{tab:time-dist}
\end{table}

\paragraph{Why training deep learning models is so expensive?}

The reason is the number of matrix multiplications that we have to perform, the memory required to store and update weights and gradients, and the limitations of the backpropagation algorithm \cite{Rumelhart1986LearningRB-backprop}.
We need to use GPUs as standard devices for performing such operations in an optimized way \cite{wang2019benchmarking}. The upward trend in the cost of the GPUs, together with the increasing complexity of deep learning models due to AI needs, lead to a very expensive research cost.
Moreover, most recent language models have billions of parameters: Google's Switch Transformer model \cite{fedus2021switch-language-trill} has 1.6 trillion parameters, and OpenAI GPT-3 model \cite{brown2020language-gpt3} has 175 billion parameters, this implies a change in the deep learning paradigm. since we need new devices and strategies for training such massive models.


\paragraph{Methodology. }
We have introduced in our planification part of our research methodology for solving AI problems. Although the broader literature on deep leraning research approaches is comprehensive, very little has been published on appropriate ways of doing AI research considering ethical and climatic implications, an example is the work from D. Michie on ''Methodologies from Machine Learning in Data Analysis and Software`` \cite{ml-methodologies}. The core of our methodology is a fast experimentation pipeline as shown in Figure \ref{fig:experiment}. We consider our research method a combination of: 
\begin{itemize}
    \item Engineering method (evolutionary method): observe state-of-the-art solutions, improve SOTA or propose better solutions, develop, measure and analyze impact, and repeat until no further improvement is possible.
    \item Empirical method (revolutionary paradigm): propose models, develop mathematical methods, aply them to case studies, validate the model, measure its impcat-performance, and repeat.
\end{itemize}

\subsection*{Computational  Resources}

We use GPUs (Graphics Processing Unit) as standard device for training and deployment of deep learning models. The GPU is a specialized processor with dedicated memory that usually performs floating point operations required for rendering graphics. In this context, we use this device because it is optimized to process multiple operations in parallel, mainly matrix multiplications (standard commercial GPUs are 4-5 times faster than CPUs).

Additionally, Kaggle and Google Collab provide TPU resources. The Tensor Processing Unit (TPU) is an AI accelerator application-specific integrated circuit (ASIC) developed by Google \cite{jouppi2017indatacenter-tpu} specialized in deep learning tasks. Deployed in datacenters since 2015, it accelerates the inference and training of neural networks.
At approximately 50 cm, a TPU v3-8 board is a fairly sizeable piece of hardware. It sports 4 dual-core TPU chips for a total of 8 TPU cores. Each TPU core has a traditional vector processing part (VPU) as well as dedicated matrix multiplication hardware capable of processing 128x128 matrices. This is the part that specifically accelerates machine learning workloads. TPUs are equipped with 128GB of high-speed memory allowing larger inputs and larger models. \\

Deep Learning models are developed using the setups described in Table \ref{tab:pc_setups}.
Kaggle and Google Collab Virtual Machines (VMs) resources are \textbf{free} after registration in each site.
Multiple VMs can be created in parallel, and thus, we can train or test multiple models in parallel.
However, notebooks run by connecting to virtual machines that have maximum lifetimes that can be as much as 12 hours in Collab, and 9 hours in Kaggle.
This is the most extended practice in academy since the cost of these resources is really high (see Table \ref{tab:comp_cost}).

\begin{figure}[]
\includegraphics[width=\textwidth]{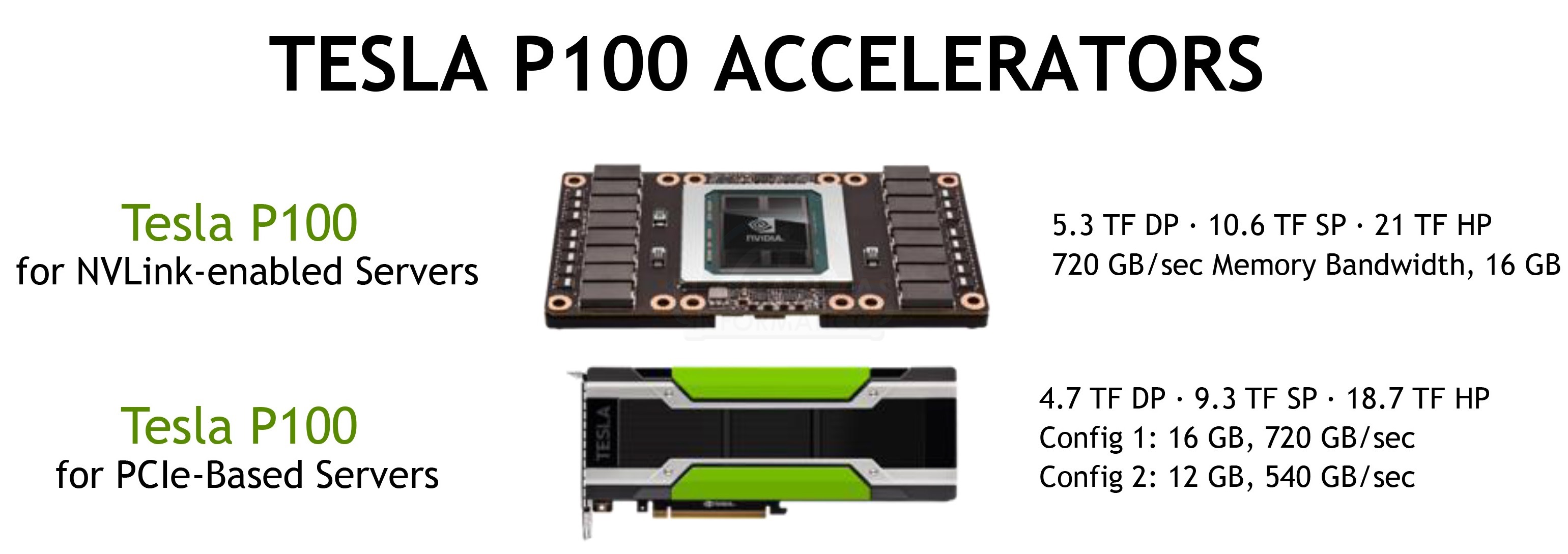}
\centering
\caption{Specifications of the Tesla P100 GPU used in this project. Source: NVIDIA.}
\label{fig:teslap100}
\end{figure}

\begin{figure}[]
\includegraphics[width=0.85\textwidth]{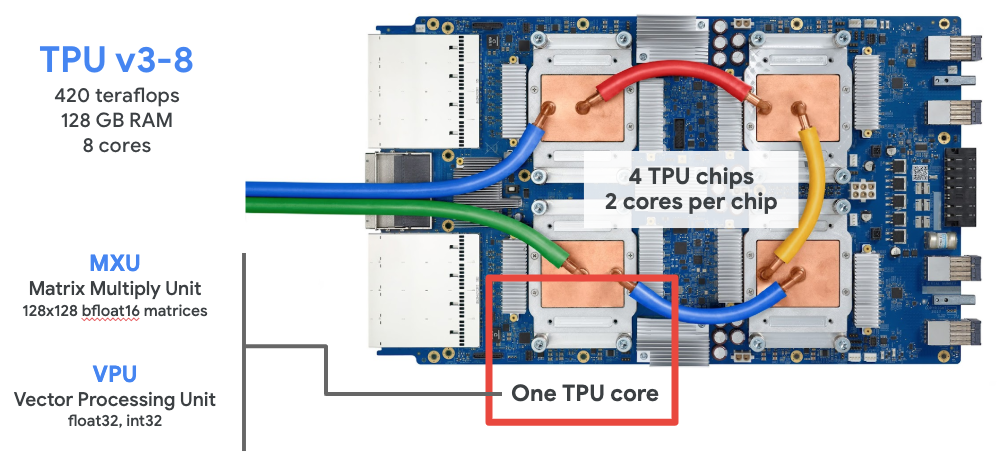}
\centering
\caption{Specifications of the TPU v3-8 board used in this project. Source: Kaggle.}
\label{fig:tpu-viz}
\end{figure}


\begin{table}[]
    \caption{Different setups for developing Deep Learning models}
    \centering
    \begin{tabular}{l l c l}
         Setup & Disk & RAM (GB) & GPU  \\
         \toprule
         Personal Computer & 1TB SSD & 32 & RTX 3070 \\
         Kaggle GPU        & 73GB    & 16 & Tesla P100 \\
         Kaggle TPU        & 73GB    & 16 & TPU v3-8 \\
         Google Collab GPU & 1008GB  & 12 & Tesla T4 \\
         \bottomrule
    \end{tabular}
    \label{tab:pc_setups}
\end{table}

\begin{table}[]
    \caption{Computational Resources and Cost}
    \centering
    \begin{tabular}{c c c c c}
    GPU & VRAM (GB) & Market cost (\$) & Platform & Final cost (\$) \\
    \toprule
     Tesla P100 & 16 & 2500 & Kaggle & 0 \\
     Tesla T4   & 15 & 2200 & Google Collab & 0 \\
     RTX 3070   & 16 & 800 & Own & 0 \\
     TPU v3-8   & 128& -   & Kaggle & 0 \\
    \bottomrule
    \end{tabular}
    \label{tab:comp_cost}
\end{table}

\subsection*{Tools}

\begin{figure}[]
\includegraphics[width=0.9\textwidth]{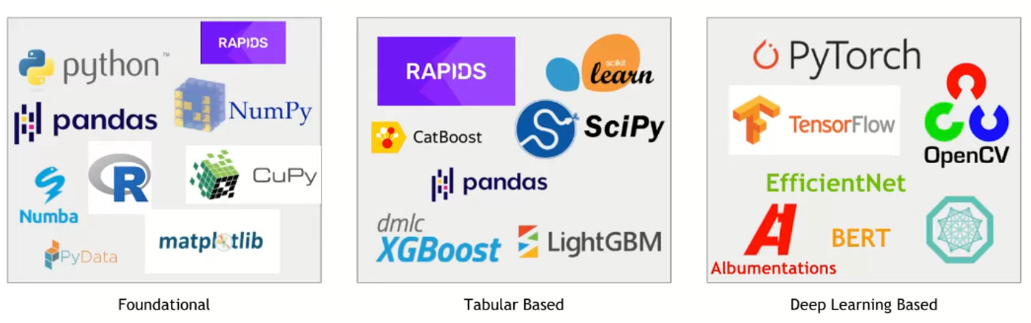}
\centering
\caption{Common Open-sourced Data Science Tools.}
\label{fig:tools}
\end{figure}

In this project, we use only open-source tools for preparing datasets, pre-processing images or tabular data, training and deploy deep learning models, etc.
The most common tools are shown in Figure \ref{fig:tools}, we use:
\begin{itemize}
    \item Programming Language: Python, C++
    \item General Data Science: Numpy, Pandas, Sklearn, SciPy, etc.
    \item Image Processing: OpenCV, Matplotlib, Albumentations.
    \item Deep Learning frameworks: Tensorflow 2.0 (Keras high-level API) and Pytorch
\end{itemize}

We will describe in detail the libraries, frameworks and software for developing Computer Vision applications in the upcoming Sections.

\subsection*{Robot Development Cost}

In the Appendix \ref{app:robot} we explain into detail the building process and cost of the robots that we use to materialize our solution and use it in industrial environments. The \textit{JetBot} costs less than 200\$ in parts, including the NVIDIA Jetson Nano board, our \textit{JetRover} (DonkeyCar) costs less than 250\$ in parts, and it takes $\approx$ 2 hours to assemble each one. The following table shows the ''Bill of Materials`` for building the basic JetBot robot:

\begin{figure}[!h]
\includegraphics[width=0.9\textwidth]{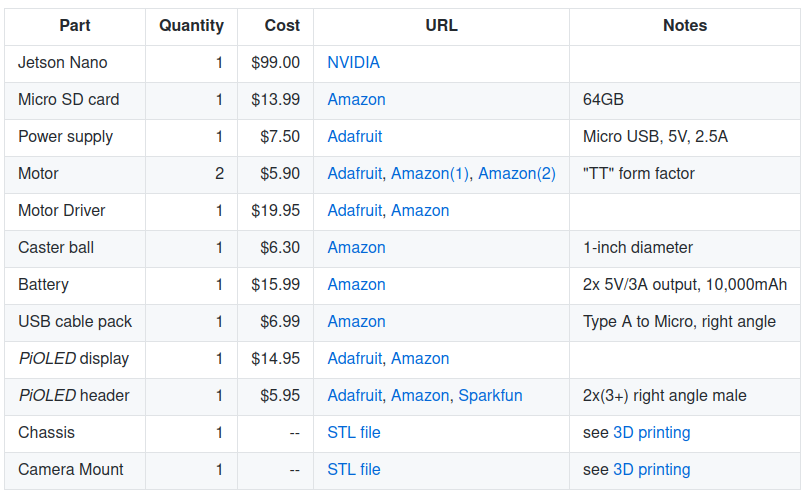}
\centering
\caption[The official Bill of materials for building the JetBot]{The official Bill of materials for building the JetBot. Source \url{https://github.com/NVIDIA-AI-IOT/jetbot/wiki/Bill-of-Materials}}
\label{fig:jetbot-bill}
\end{figure}

Some components as the Jetson Nano board or Raspberry PI Cam are reused to build the advanced version of the JetBot, our so-called JetRover (Donkey CAr-style) inspired by the MIT RACECAR and the JetRacer. 
We do not consider 3D printing or assembly hardware (screws, wires) as additional costs.


\chapter{Computer Vision}
\label{computer_vision}

\ifpdf
    \graphicspath{{Chapter2/Figs/Raster/}{Chapter2/Figs/PDF/}{Chapter2/Figs/}}
\else
    \graphicspath{{Chapter2/Figs/Vector/}{Chapter2/Figs/}}
\fi

In this Chapter, we introduce the main computer vision concepts and techniques used to build a Visual Perception system.

\section{High-Level Vision Tasks}

The three main computer vision tasks well-described and studied in the literature are: 
\begin{enumerate}
    \item Classification: for a given image predict to one of a set of predefined categories or classes the image belongs. There might be 1 class (binary classification) or multiple classes (multi-class).
    \item Detection: this task aims to find the location of a single (or multiple) objects in an image, this task is an extension of the classification. We provide the localization of the object using coordinates related to the image, usually this coordinates conform a box called ''bounding box``.
    \item Segmentation: is an extension of the detection task, in this case, we want to know where the objects are and also its particular area, not a wide bounding box. We will not focus on this task, as the model's require to perform it are too complex to train and integrate.
\end{enumerate}

By definition, the most important and common task is image classification. All these task have a common approach, Convolutional Neural Networks (CNNs or ConvNets) \cite{LeCun1999-cnns, imagenet-classification-cnn}. To explain CNNs architectures, functions and applications, we use as baseline the exceptional material from the ''CS231n Convolutional Neural Networks for Visual Recognition`` \footnote{\url{https://cs231n.github.io/}}.

\begin{figure}[!ht] 
\centering    
\includegraphics[width=0.8\textwidth]{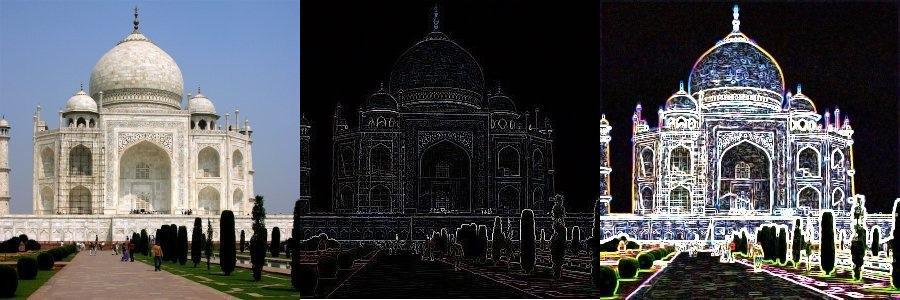}
\caption[Sobel Filter]{Sobel Filter Example.}
\label{fig:sobel}
\end{figure}

\begin{figure}[!ht] 
\centering    
\includegraphics[width=0.9\textwidth]{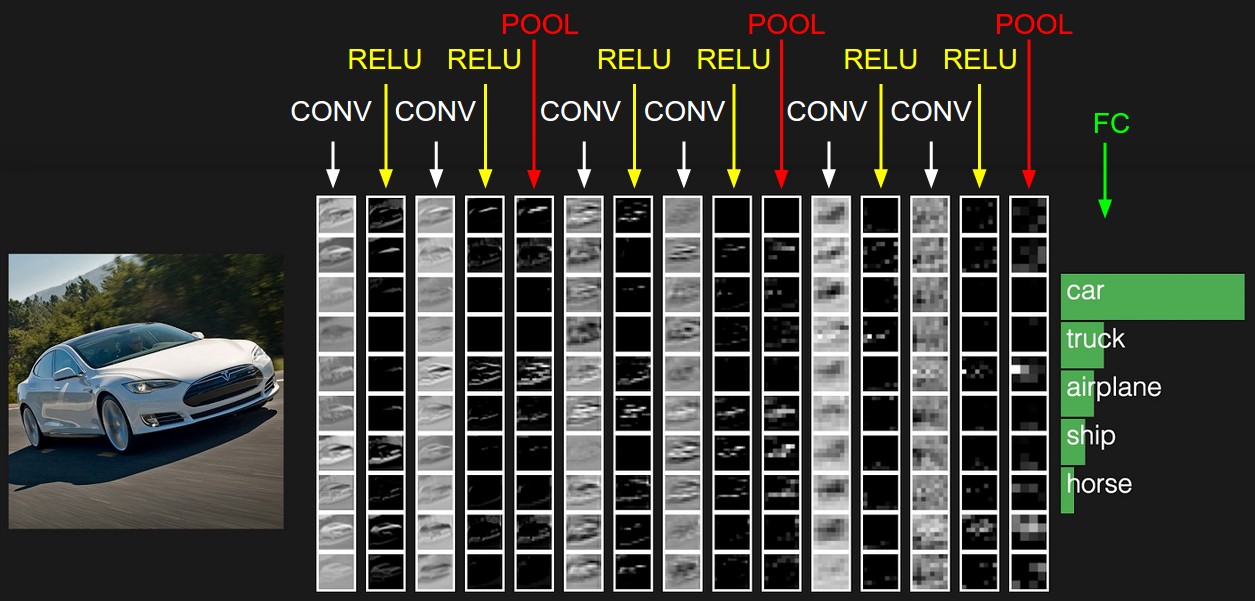}
\caption[CNN Demo]{CNN Demo. Source: CS231n.}
\label{fig:cnn-demo}
\end{figure}

CNNs are learnable feature extractors. For example, we can apply the well-known Sobel Filter ($3\times3$ matrix) \cite{996-sobel} for edge extraction as shown in Figure \ref{fig:sobel}. The edges are \textit{features} of the image, in some cases, using only the edges we can properly classify the image, for example flattening the edge's image (a transformation from 3-channel tensor to vector) and using classical models as SVM or Logistic Regression to infer the class. Let's suppose now than instead of having a fixed filter ($3\times3$ matrix), we can learn it in a way such that the filter's extracted features are optimal for classifying the image.
In computer vision, CNNs are made of blocks, the most important block is the convolutional layer, which have the learnable kernels (also called filters) that allow us learn different image's representations and extract features such that we can classify images easier. Figure \ref{fig:cnn-demo} shows such deep representations of the main image that the CNN learn and facilitate its classification. We will introduce briefly the main blocks that we need to build any CNN and the most common network architectures for each problem.

\clearpage

\begin{figure}[!ht] 
\centering    
\includegraphics[width=0.8\textwidth]{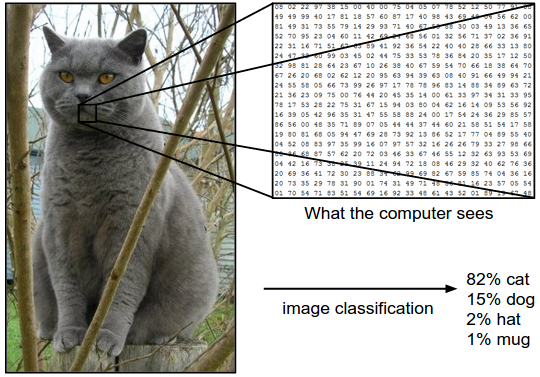}
\caption[Image representation in computer vision]{Image representation in computer vision. Source: CS231n}
\label{fig:image-representation}
\end{figure}

The main blocks of the CNNs are \footnote{\url{https://cs231n.github.io/convolutional-networks/}}:

\begin{enumerate}

\item The input image discretized, always represented in a vector space as a 3D tensor of shape $h \times w \times 3$ where $d$ represents the height, $w$ the width and $d$ the depth or number of channels, typically 3 as we work with RGB images. The images are stored as INT8, meaning the pixel have values from 0 to 255, however, when we train the networks, images are always normalized and converted to FLOAT32. See Figure \ref{fig:image-representation} as an example.

\item Convolutional layers will compute the output of neurons that are connected to local regions in the input, each computing a dot product between their weights (filter or kernel) and a small region they are connected to in the input volume (image or deep representation of it). This may result in volume with $F$ filters, since each filter will extract features from the input volume. The kernels or filters (and their bias), are the main learnable element in the CNNs. Moreover, the convolution operation only performs multiplications and addtions, therefore, the output generated for a given input is a linear combination of the input and the used kernel (assuming there not activations in between).

\item Activations, the most common one is RELU \cite{xu2015empirical-relu} , Sigmoid, Softmax etc. Activations will apply an elementwise activation function, such as the max(0,x) thresholding at zero. Activations do not change the dimension of the input (size of the volume). 

\item Pool layers will perform a downsampling operation along the spatial dimensions (width, height), resulting in volume reduction (dimensionality reduction). The pooling can be done attending to max or mean values of a given window of size ($s \times k$), typically ($2 \times 2$).

\item Dropout \cite{JMLR:v15:srivastava14a-dropout} is simple, yet powerful regularization technique for neural networks. This technique allows to randomly ignore neurons during training, they are “dropped-out” randomly.

\item FC (i.e. fully-connected) , Dense or MLP layer takes as input features and will return other dense features or typically the output of the network (class scores, regressed values, etc). As with ordinary Neural Networks and as the name implies, each neuron in this layer will be connected to all the numbers in the previous volume.

\end{enumerate}

Using only this blocks to build a CNN, we can achieve state-of-the-art results on every vision task.
\clearpage

\begin{figure}[!ht] 
\centering    
\includegraphics[width=1.0\textwidth]{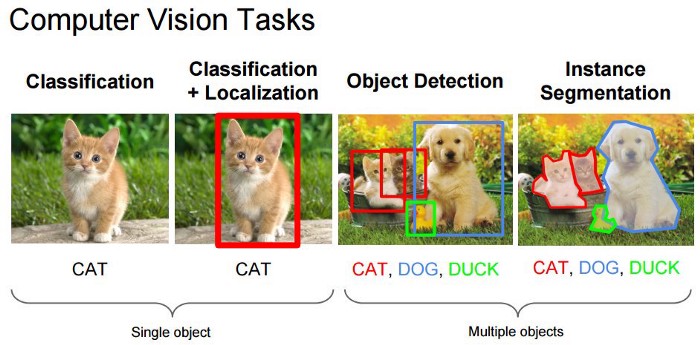}
\caption[Computer Vision 3 Basic Taks]{Computer Vision 3 main Taks}
\label{fig:3tasks}
\end{figure}

\clearpage

\subsection{Classification}
\label{sec:classification}

\begin{figure}[!ht] 
\centering    
\includegraphics[width=0.8\textwidth]{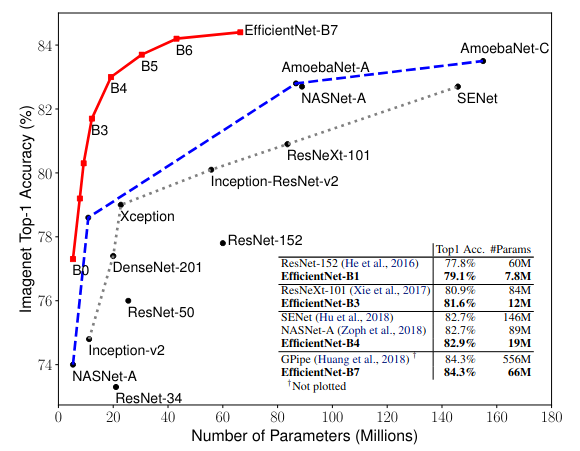}
\caption[Classification Benchmarks]{Classification SOTA Benchmarks\cite{tan2020efficientnet}.}
\label{fig:class-benchmark}
\end{figure}

The most common benchmark and dataset to train and test models is ImageNet \cite{imagenet, imagenet-classification-cnn}. Figure \ref{fig:class-benchmark} shows the performance of different models in this benchmark, this is usually a sign of model's generalization and learning capabilities.

\begin{figure}[!ht] 
\centering    
\includegraphics[width=\textwidth]{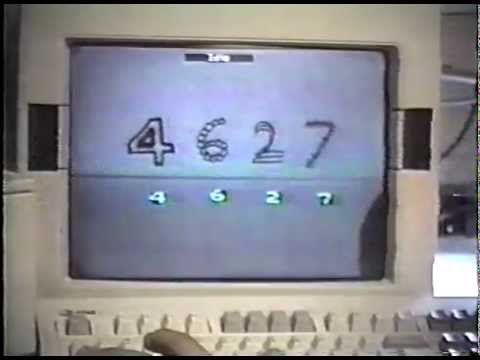}
\caption[First Digit Recognizer Application]{First Digit Recognizer Application by Yann Lecun.}
\label{fig:digitrecog}
\end{figure}

\begin{figure}[!ht] 
\centering    
\includegraphics[width=\textwidth]{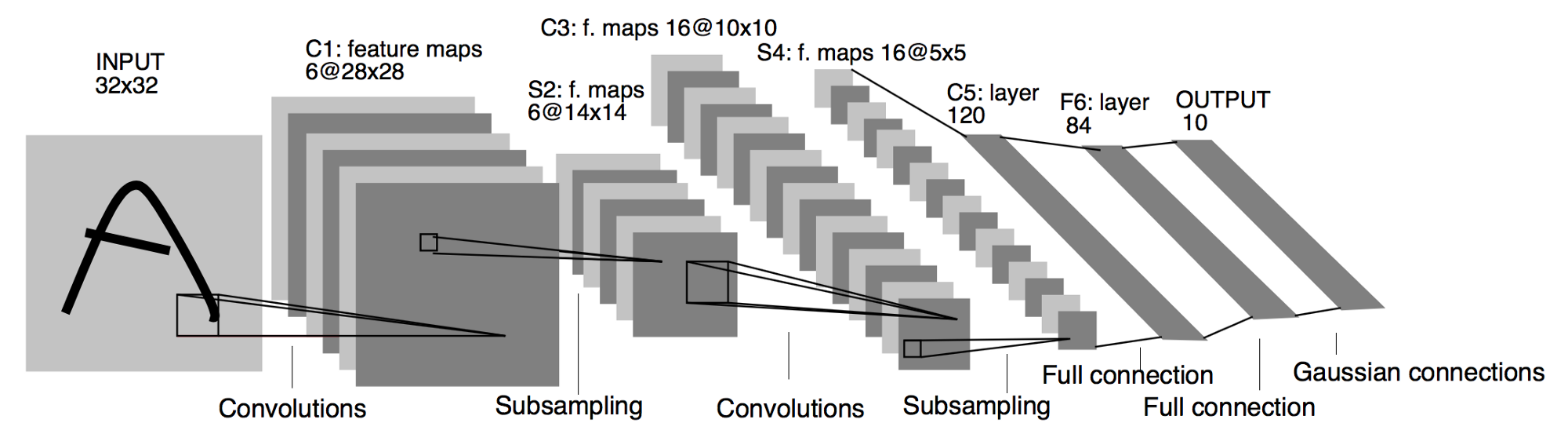}
\caption[LeNet Architecture]{LeNet Architecture.}
\label{fig:lenet}
\end{figure}

The most popular Convolutional Neural Network architectures are listed and described in CS231 \footnote{\url{https://cs231n.github.io/convolutional-networks/}}, ordered by its year of publication:

\begin{itemize}
    \item LeNet. The first successful applications of Convolutional Networks were developed by Yann LeCun in 1993 (see Figure \ref{fig:digitrecog}). Of these, the best known is the LeNet architecture that was used to read zip codes, digits, etc. The architecture is shown in Figure \ref{fig:lenet}.
    
    \item VGGNet \cite{vgg}. The runner-up in ILSVRC 2014 was the network from Karen Simonyan and Andrew Zisserman that became known as the VGGNet. Its main contribution was in showing that the depth of the network is a critical component for good performance. Their final best network contains 16 CONV/FC layers and, appealingly, features an extremely homogeneous architecture that only performs 3x3 convolutions and 2x2 pooling from the beginning to the end. A downside of the VGGNet is that it is more expensive to evaluate and uses a lot more memory and parameters (140M) due to the required FC layers, and it was since found that these FC layers can be removed with no performance downgrade (using Global Average Pooling for instance), significantly reducing the number of necessary parameters.
    
    \item ResNet \cite{resnet}. Residual Network developed by Kaiming He \textit{et al} was the winner of ILSVRC 2015. It features special skip connections and a heavy use of batch normalization \cite{ioffe2015batch}. The architecture is also missing fully connected layers at the end of the network, which reduces the number of parameters and model's complexity. The proposed novel residual block, for avoiding gradients vanishing and propagate error information improved notably CNNs performance. ResNets are still state of the art Convolutional Neural Network. 
    
\end{itemize}

\begin{figure}[!ht] 
\centering    
\includegraphics[width=\textwidth]{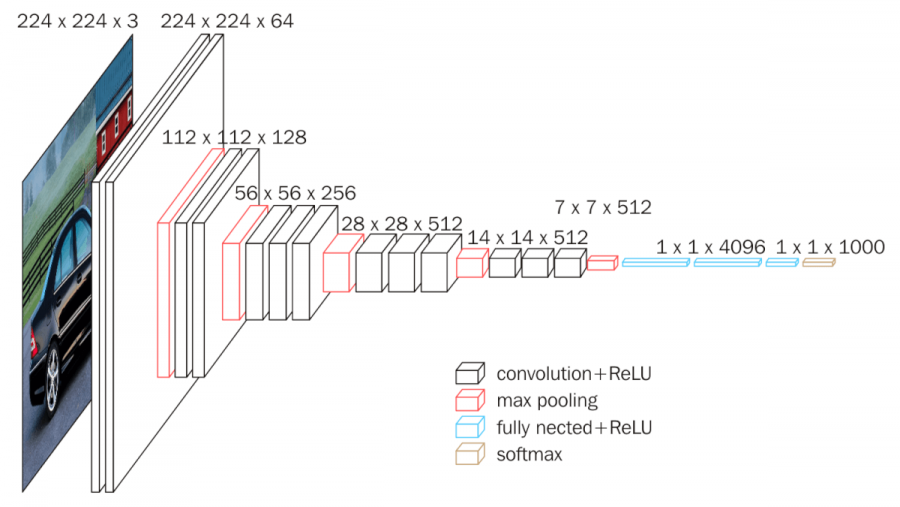}
\caption{VGG Architecture.}
\label{fig:vgg}
\end{figure}

\begin{figure}[!ht] 
\centering    
\includegraphics[width=\textwidth]{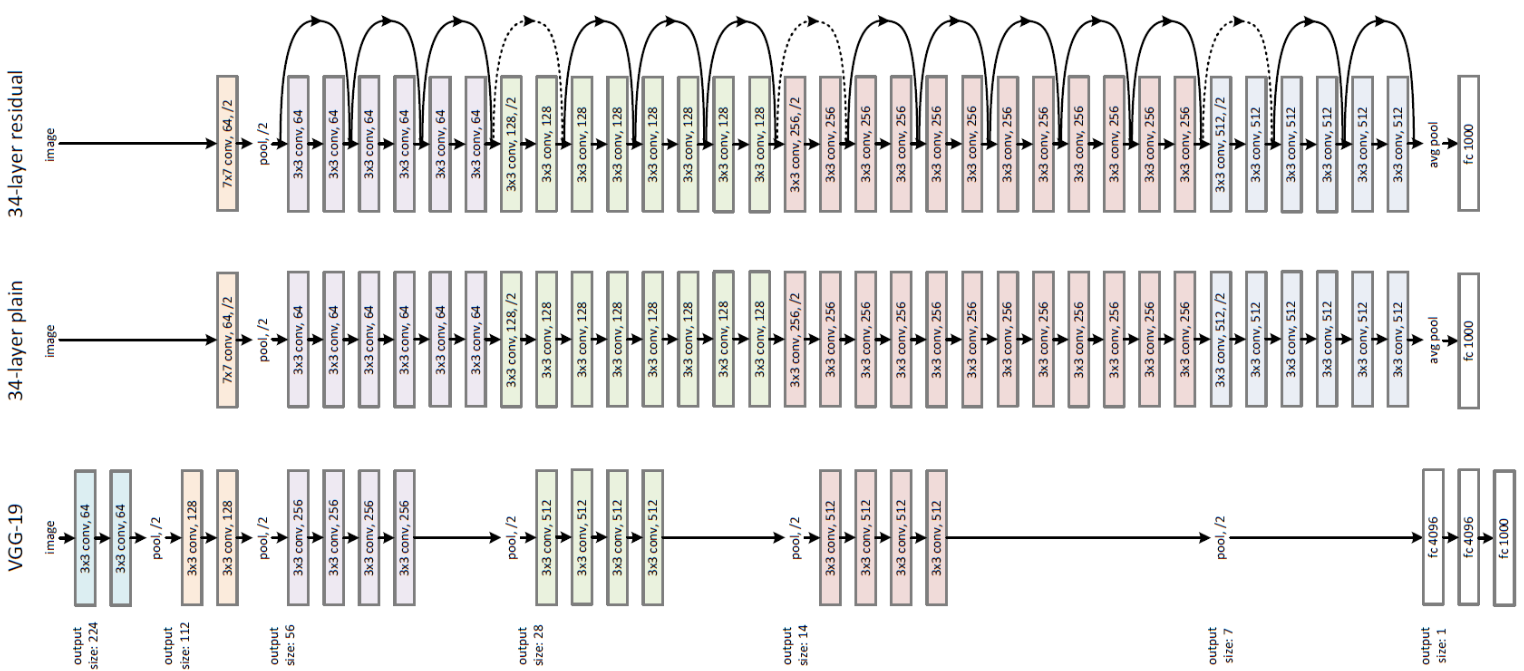}
\caption{ResNet architecture \cite{resnet}}
\label{fig:resnet}
\end{figure}

\clearpage

\subsection{Object Detection}
\label{sec:object-detecttion}

\begin{figure}[!ht] 
\centering    
\includegraphics[width=\textwidth]{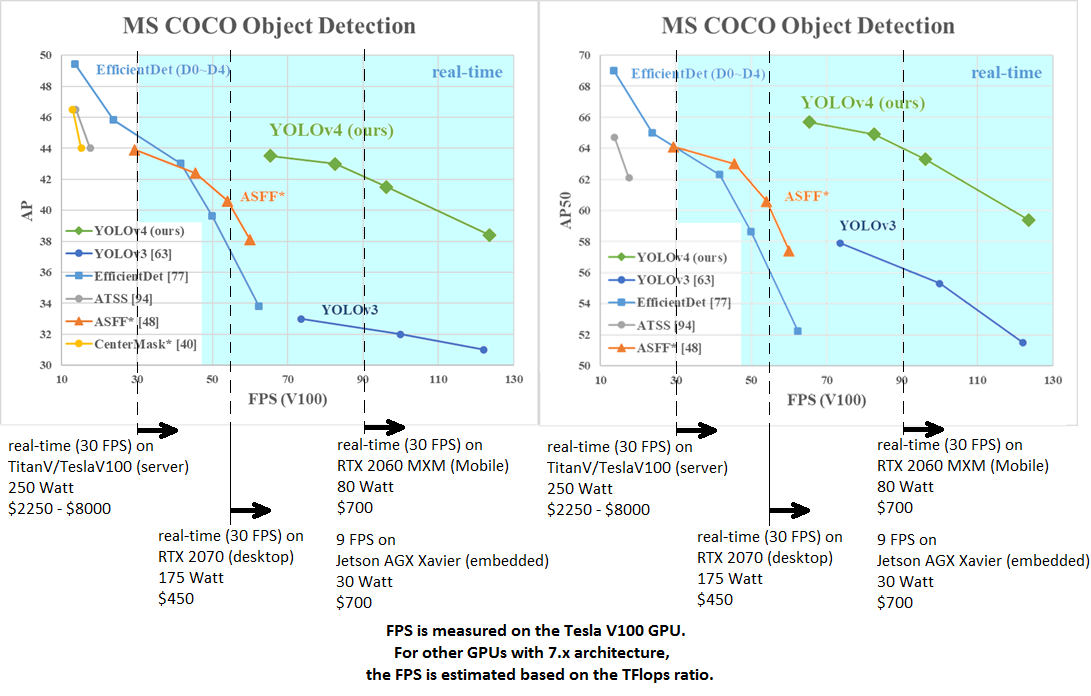}
\caption{MS-COCO SOTA Benchmark \cite{lin2015microsoft-coco-dataset}}
\label{fig:coco}
\end{figure}

The most common benchmark and dataset to train and test models is MS-COCO \cite{lin2015microsoft-coco-dataset}. Figure \ref{fig:coco} shows the performance of different models in this benchmark, this is usually a sign of model's generalization and learning capabilities.

The most widely used state of the art version of the R-CNN family — Faster R-CNN \cite{ren2016faster}.
These networks usually consist of: (i) A region proposal algorithm to generate “bounding boxes” or locations of possible objects in the image; (ii) A feature generation stage to obtain features of these objects, usually using a CNN; (iii) A classification layer (a network from the previous section) to predict which class this object belongs to; and (iv) A regression layer to make the coordinates of the object bounding box more precise.

The only stand-alone portion of the network left in Fast R-CNN was the region proposal algorithm. Both R-CNN and Fast R-CNN use CPU based region proposal algorithms, Eg- the Selective search algorithm which takes around 2 seconds per image and runs on CPU computation. The Faster R-CNN \cite{ren2016faster} paper fixes this by using another convolutional network (the RPN) to generate the region proposals. This not only brings down the region proposal time from 2s to 10ms per image but also allows the region proposal stage to share layers with the following detection stages, causing an overall improvement in feature representation. In the rest of the article, “Faster R-CNN” usually refers to a detection pipeline that uses the RPN as a region proposal algorithm, and Fast R-CNN as a detector network.

\begin{figure}[!ht] 
\centering    
\includegraphics[width=\textwidth]{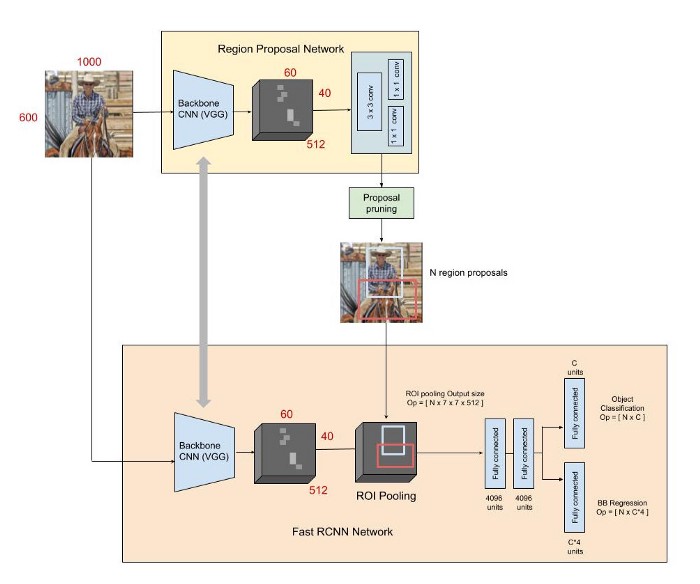}
\caption{R-CNN architecture \cite{ren2016faster}}
\label{fig:rcnn}
\end{figure}

\begin{figure}[!ht] 
\centering    
\includegraphics[width=\textwidth]{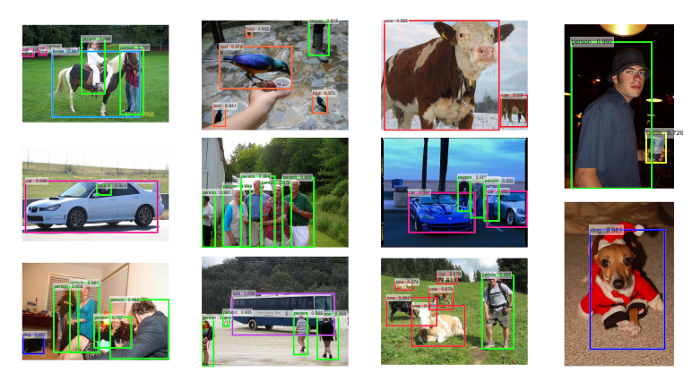}
\caption{R-CNN Sample results \cite{ren2016faster}}
\label{fig:rcnn-results}
\end{figure}

There other network architectures as CenterNet \cite{duan2019centernet} that not require the anchors, the bounding box candidates. In Sections \ref{fig:facemask-detect} and \ref{baidu} we provide more information about these Objects Detectors.\\

The NMS (Non-Maximum Supression Algorithm) unifies the ROIs (Regions of interested) generated by the Region proposal network (RPN), candidates to bounding box. The following figure shows the effect of this algotithm:

\begin{figure}[!ht] 
\centering    
\includegraphics[width=\textwidth]{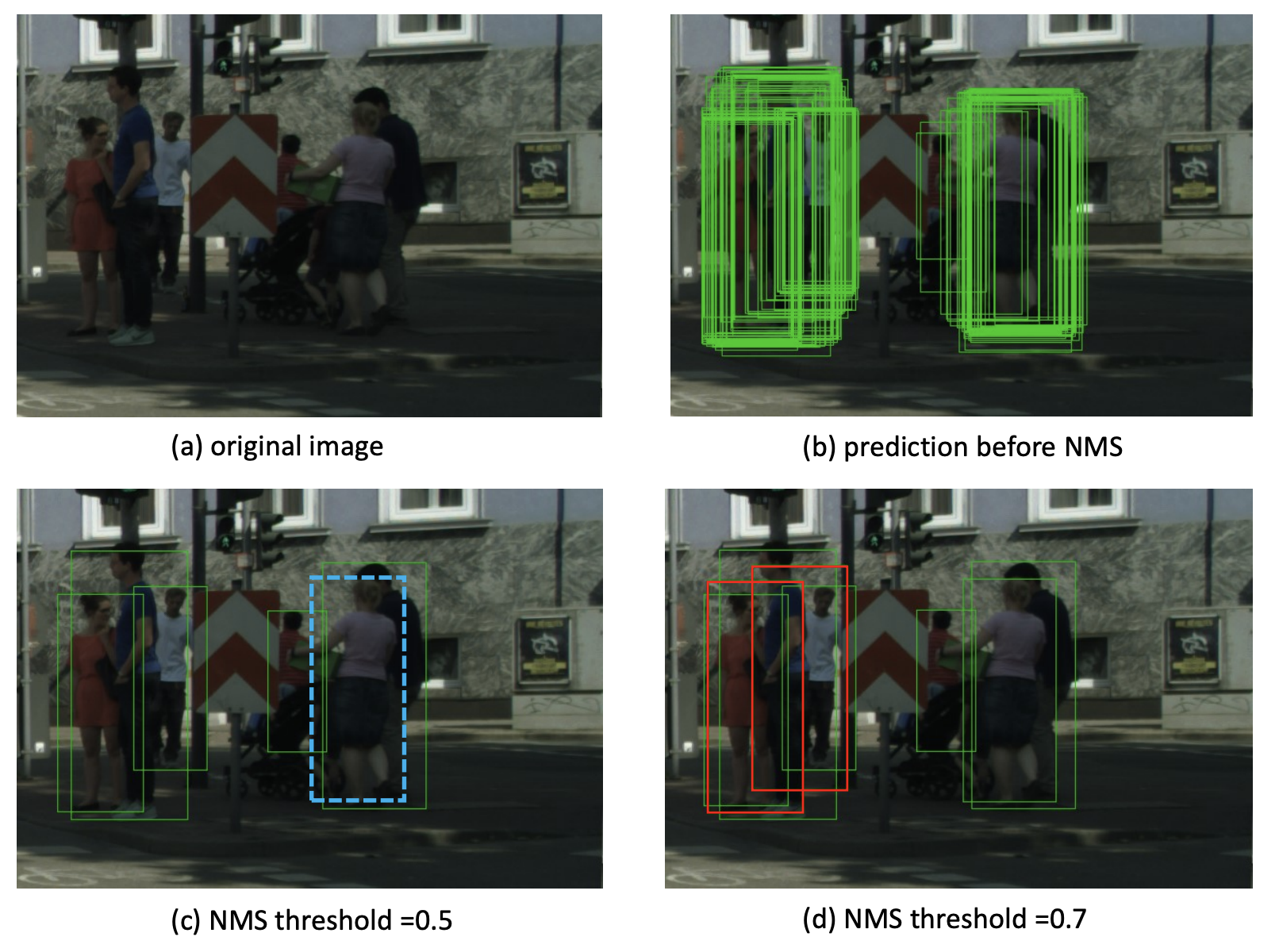}
\caption{NMS results}
\label{fig:nms}
\end{figure}

\clearpage
\section{Training and Deploying Deep Learning Models}
\label{sec:train-deploy}

\subsection{Training Strategies}

Our Deep Learning models are based on Convolutional Neural Networks (CNNs), as we explained in Section \ref{computer_vision}, the main computer vision task is ''classification``. He \textit{et.al.} introduced in the paper ''Bag of Tricks for Image Classification with Convolutional Neural Networks`` \cite{he2018bagtricks} a collection of techniques and empirically evaluate their impact on the final model performance. They also prove that ''improvement on image classification performance leads to better transfer learning performance in other application domains such as object detection and semantic segmentation`` \cite{he2018bagtricks, zhang2019bagtricks}.

\subsubsection{Transfer Learning}

''Transfer learning and domain adaptation refer to the situation where what has been learned in one setting … is exploited to improve generalization in another setting`` \cite{Goodfellow-et-al-2016-deeplearning}.

This is a concept widely used in deep learning literature, transfer learning make use of the knowledge gained while solving one problem (source domain) and applying it to a different but related problem (target domain).
For example, knowledge gained while learning to classify animals can be used to some extent to recognize particular breeds of dogs.

Usually we train the network on a large dataset (for example: ImageNet \cite{imagenet} or MS-COCO \cite{lin2015microsoft-coco-dataset}), meaning we train all the parameters of the neural network, this is what we commonly call ''training`` or ''learning`` process. Such models are also called \textit{pre-trained models}, because the are already trained for a generic task on large dataset (i.e Classification task using ImageNet or object detection using COCO).
We can reuse these models for other classification tasks, for instance, the models trained on ImageNet learn to classify cars, persons, dogs,cats, etc (A), we can take such pre-trained model as baseline and train it on a different dataset-task, for instance, classify breed of birds (B). By doing this, we are transferring the learning from task A to task B, this accelerates de learning of B in comparison with training such task from scratch \cite{tan2018survey-transfer, zhuang2020comprehensive-transfer}.
Additionally, we can \textit{Fine-Tune} models the pretrained models, if the new dataset is very small, it’s better to train only the final layers of the network to avoid overfitting, keeping all other layers fixed.

\clearpage

\subsubsection{Augmentations}
\label{sec:augmentations}

We use the \emph{Albumentations} \cite{Buslaev_2020-albus} library, this is a Python computer vision library for fast and flexible image augmentations. Albumentations efficiently implements a rich variety of image transform operations, optimized into an efficient and concise image augmentation interface for different computer vision tasks (classification, segmentation, and detection).
The library is widely used in industry, deep learning research, and open source projects. Note that this library is adapted to Tensorflow and Pytorch, and easy to integrate in data loaders.\\

The following is an example image augmentation pipeline using the library \footnote{\url{https://albumentations.ai/docs/getting_started/image_augmentation/}}

\begin{lstlisting}[language=Python]
transform = A.Compose([
    A.RandomCrop(width=256, height=256),
    A.HorizontalFlip(p=0.5),
    A.RandomBrightnessContrast(p=0.2),
])
\end{lstlisting}

In this pipeline we define 3 transformations (RandomCrop, HorizontalFlip, RandomBrightnessConstrast), the probability of occurrence $p$ of each transformation, and the order we apply them.

\begin{figure}[!h]
\includegraphics[trim={0 0 0 0 },clip, width=\textwidth]{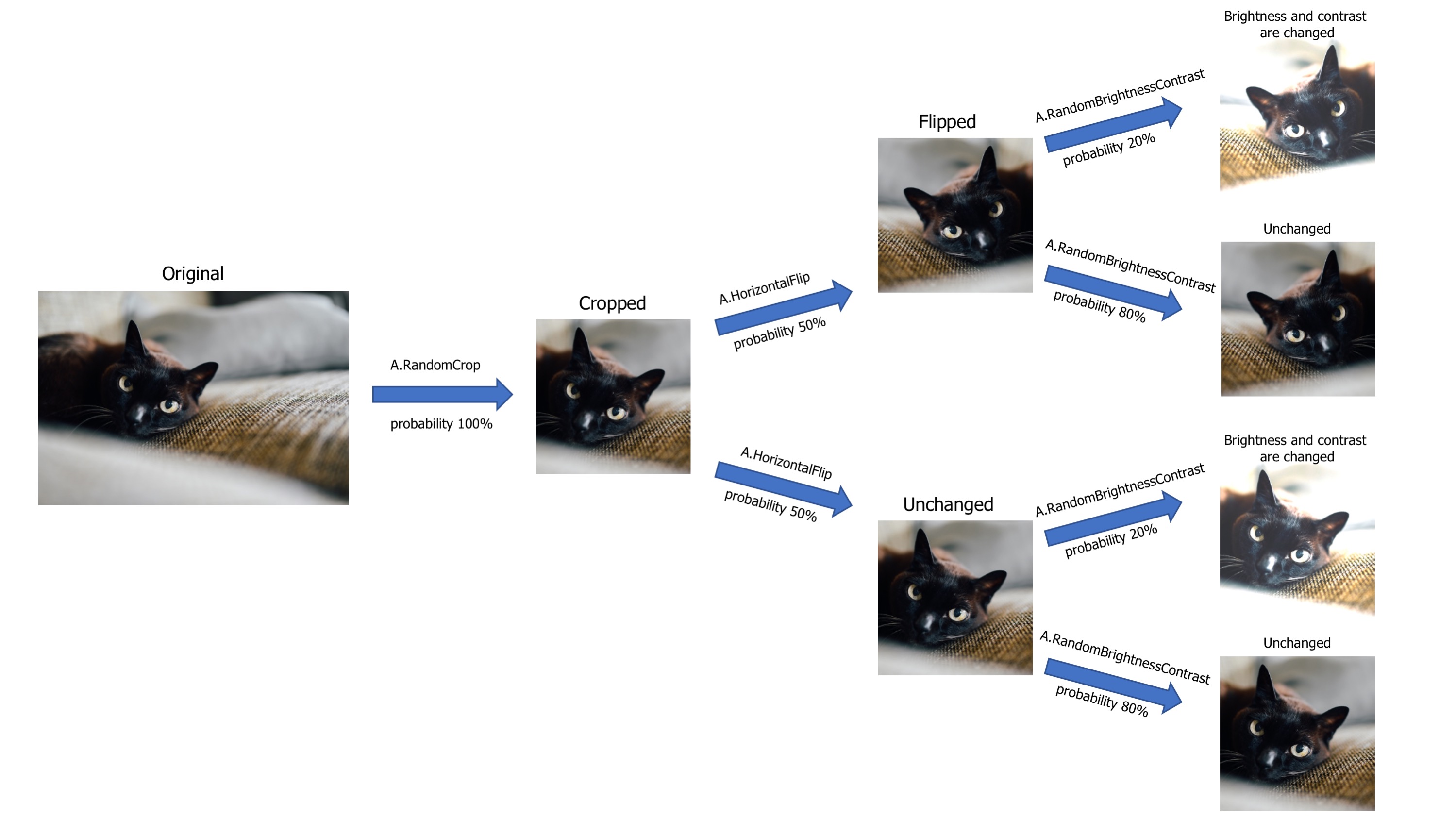}
\centering
\caption[Albumentations simple pipeline]{Albumentations simple pipeline (code above).}
\label{fig:albu-pipe}
\end{figure}

Now we can apply the proposed transformation ($transform$) to an RGB image:

\begin{lstlisting}[language=Python]

# Read an image with OpenCV and convert it to the RGB colorspace
image = cv2.imread("image.jpg")
image = cv2.cvtColor(image, cv2.COLOR_BGR2RGB)

# Augment an image
transformed = transform(image=image)
transformed_image = transformed["image"]

\end{lstlisting}

The following visualizations are examples of the possible transformations using the library.

\begin{figure}[!h]
\includegraphics[trim={0 0 0 0 },clip, width=0.9\textwidth]{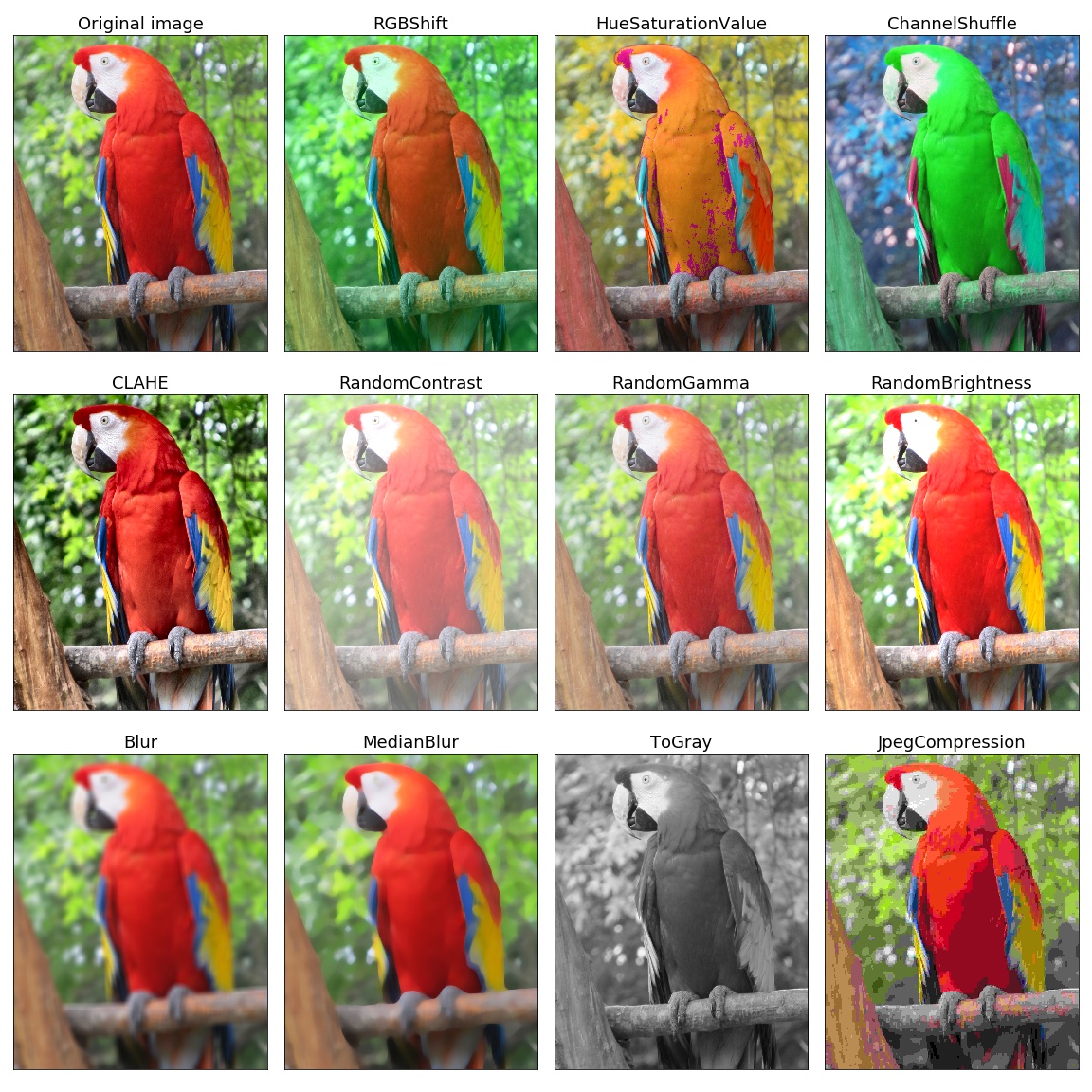}
\centering
\caption[Albumentations Demo Transformations]{Albumentations Demo Transformations.}
\label{fig:albu-demo1}
\end{figure}

Besides classical image transformations, we can also augment bounding boxes, which is extremely useful for training object detectors \footnote{\url{https://albumentations.ai/docs/getting_started/bounding_boxes_augmentation/}}. 

\begin{figure}[!h]
\includegraphics[trim={0 0 0 0 },clip, width=0.9\textwidth]{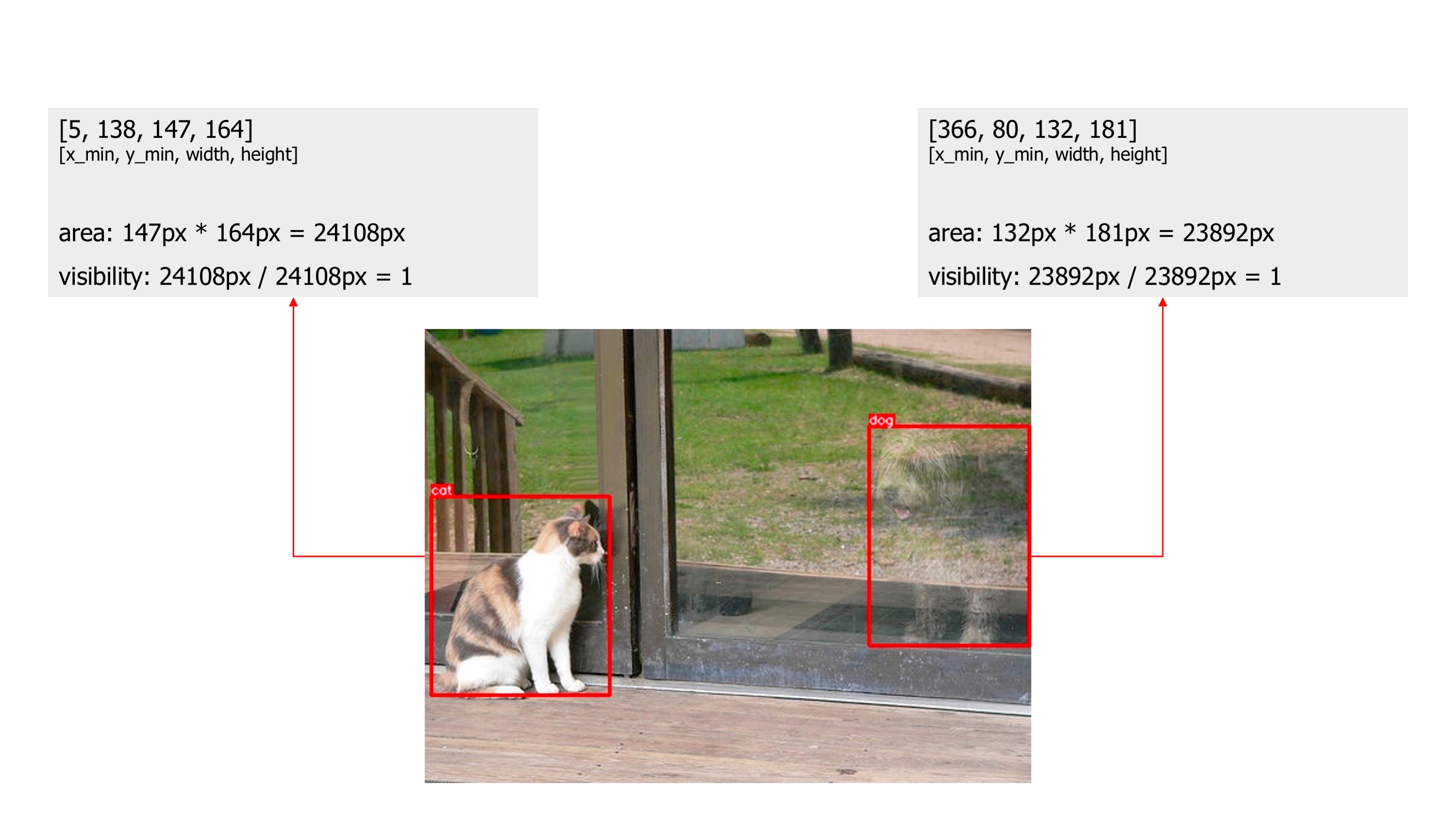}
\centering
\caption[Albumentations Bounding Boxes example]{Albumentations Bounding Boxes example.}
\label{fig:albus-2bbox}
\end{figure}

\begin{figure}[!h]
\includegraphics[trim={0 0 0 0 },clip, width=0.9\textwidth]{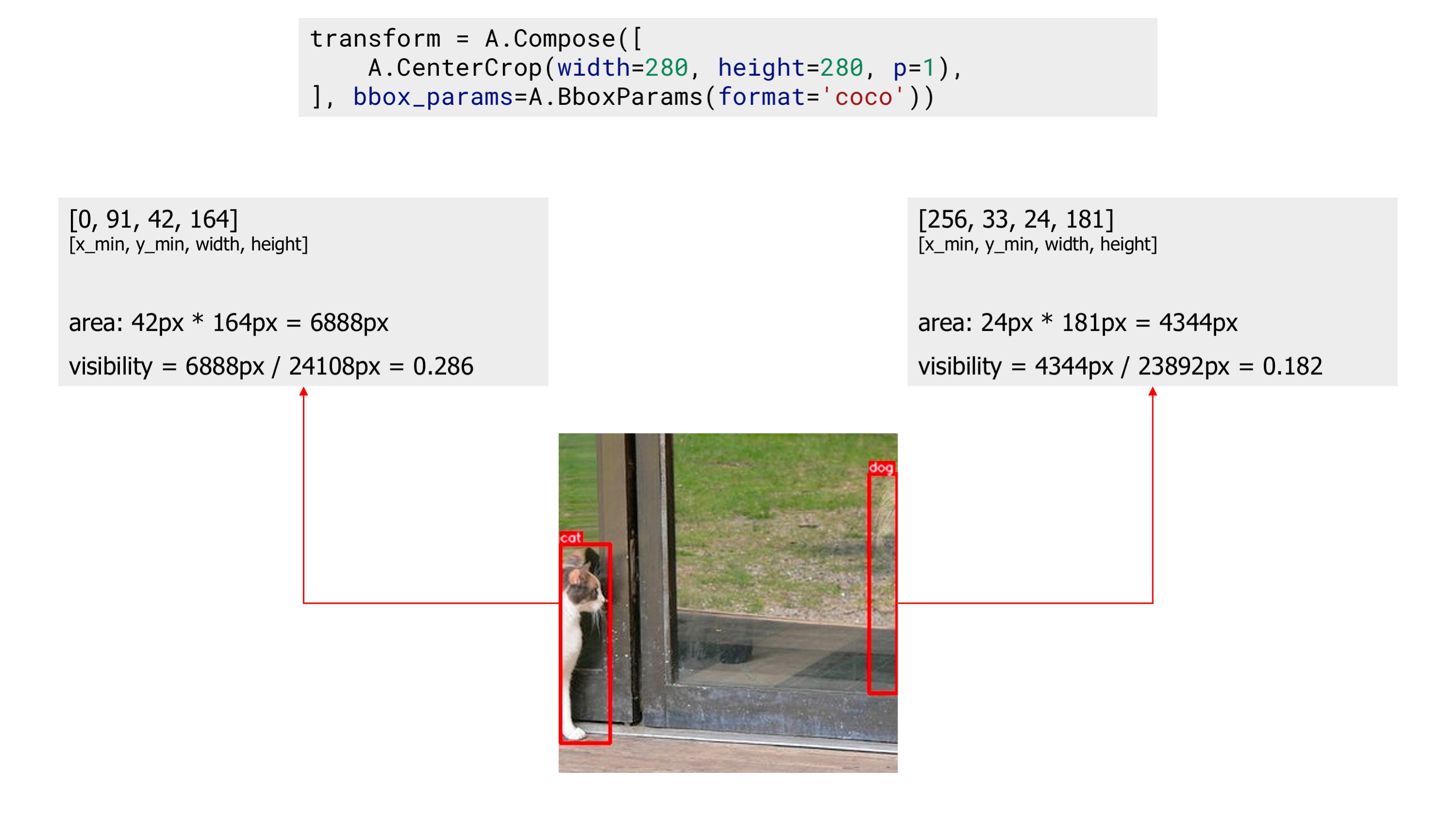}
\centering
\caption[Albumentations Bounding Boxes Augmentations example]{Albumentations Bounding Boxes Augmentations example.}
\label{fig:albus-2bbox-aug}
\end{figure}

\begin{figure}[!h]
\includegraphics[trim={0 0 0 0 },clip, width=0.9\textwidth]{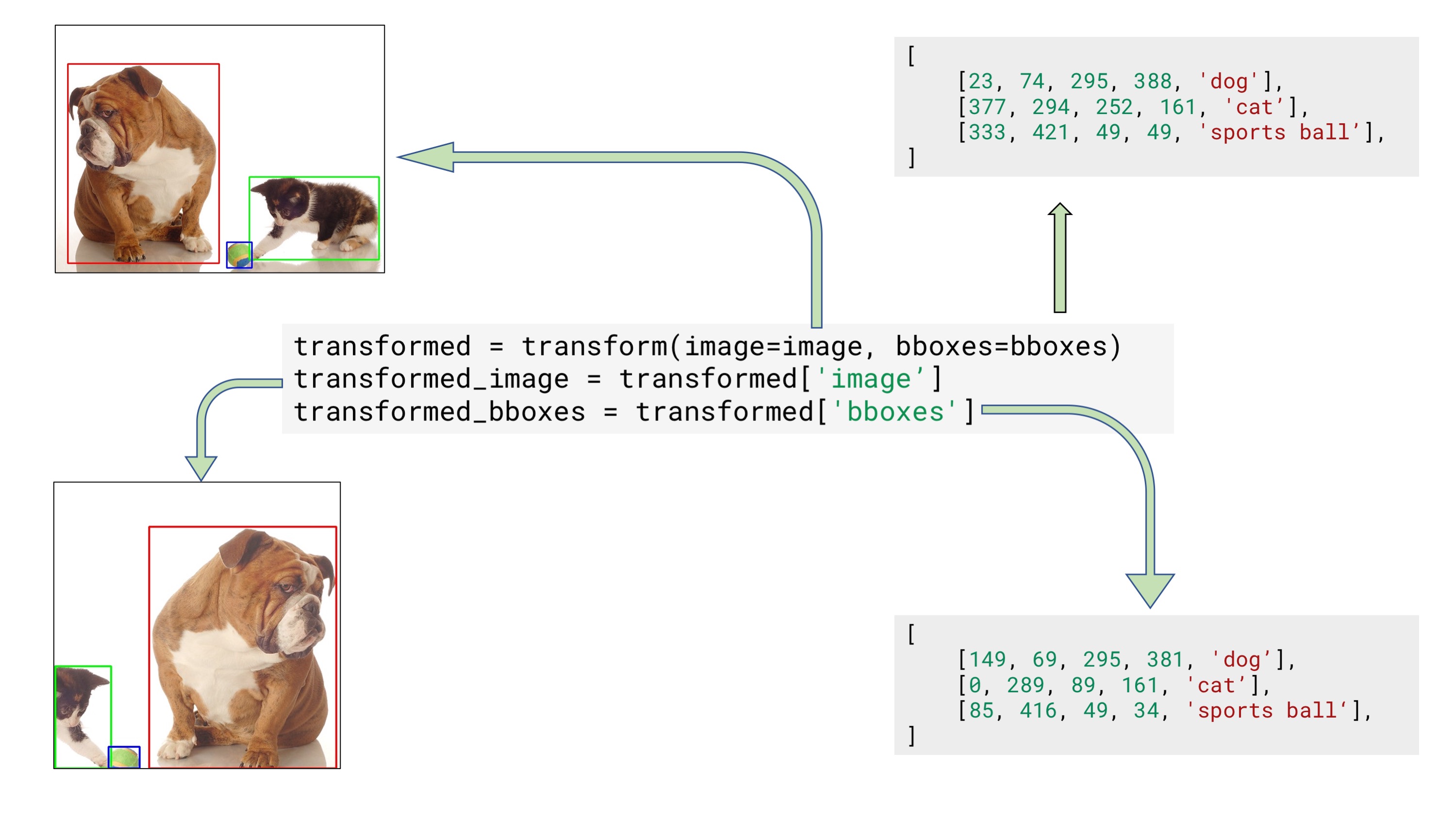}
\centering
\caption[Albumentations Bounding Boxes Augmentations example 2]{Albumentations Bounding Boxes Augmentations example 2.}
\label{fig:albu-bbox-augs2}
\end{figure}

\begin{figure}[!h]
\includegraphics[trim={0 0 0 0 },clip, width=0.7\textwidth]{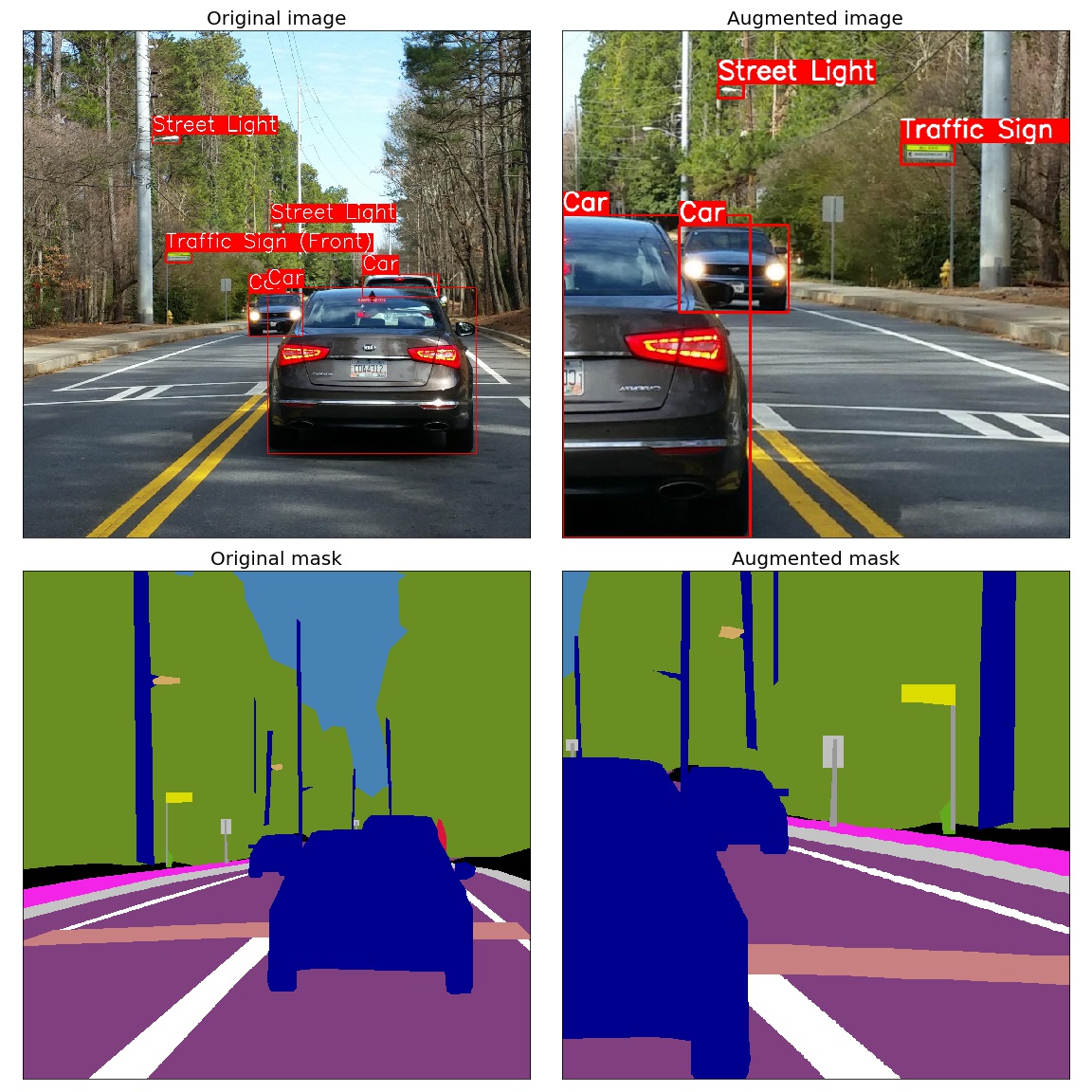}
\centering
\caption[Albumentations Demo Transformations - Detections and Segmentations]{Albumentations Demo Transformations - Detections and Segmentations.}
\label{fig:albu-demo2}
\end{figure}

\clearpage
\paragraph{Why Augmentations improve model's performance?}

\begin{figure}[!h]
\includegraphics[trim={0 0 0 0 },clip, width=\textwidth]{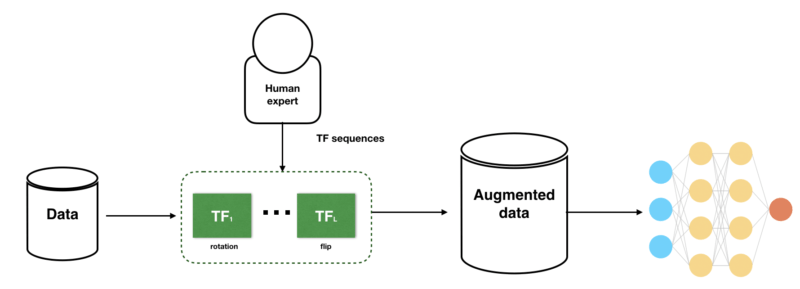}
\centering
\caption{Data Augmentations Pipeline by Stanford Class.}
\label{fig:augs-stanford}
\end{figure}

As we have seen, data augmentation or commonly called ''augmentations`` is a technique to generate virtually infinite synthetic realistic data from real data.
Since the performance of most Machine Learning models depend on quantity and diversity of data, augmentations play a crucial role on training machine learning models.
During training (on-line), we artificially increase the number of training examples (see Figure \ref{fig:augs-stanford}), this helps reducing overfitting and improves generalization, as it works as a regularization method \cite{article-augs-2019, zoph2019learning-augs, dvornik2019importance-augs, Hern_ndez_Garc_a_2018-augs}. While simple image transformations like flips can already improve predictive performance in most vision tasks, larger gains can be obtained by leveraging task-specific prior knowledge, this implies that data augmentations is another important part of the experiment design.

\clearpage
\subsection{Real-time Inference}

Training and testing models on the ''laboratoy`` is the initial part, once best models are selected, we have to deploy them on our robot. Moreover, due to computational and memory constraints of the embedded computers, models need to be low-complexity and the inference pipeline needs to be efficient in terms of IO (reading images, processing them, etc).

We use NVIDIA TensorRT, an SDK for high-performance deep learning inference that includes a deep learning inference optimizer and runtime that delivers low latency and high throughput for deep learning inference applications \footnote{\url{https://developer.nvidia.com/tensorrt}}.
Thanks to this pipeline, we only need to save the model trained on the laboratory into the board, get camera's input (RGB image) with an IO API, feed the image (represented as $h \times w \times 3$ tensor) into the model and get the output tensor. Since the JetPack SDK provides base code for the main tasks, indeed, we require minor architectural modifications.

It is interesting to mention the reasons why TensorRT is x40 faster than CPU: Mixed precision (quantizes models to INT8 while preserving accuracy), CUDA Kernel Auto-Tuning, Dynamic memory

\begin{figure}[!h]
\includegraphics[trim={0 0 0 0 },clip, width=\textwidth]{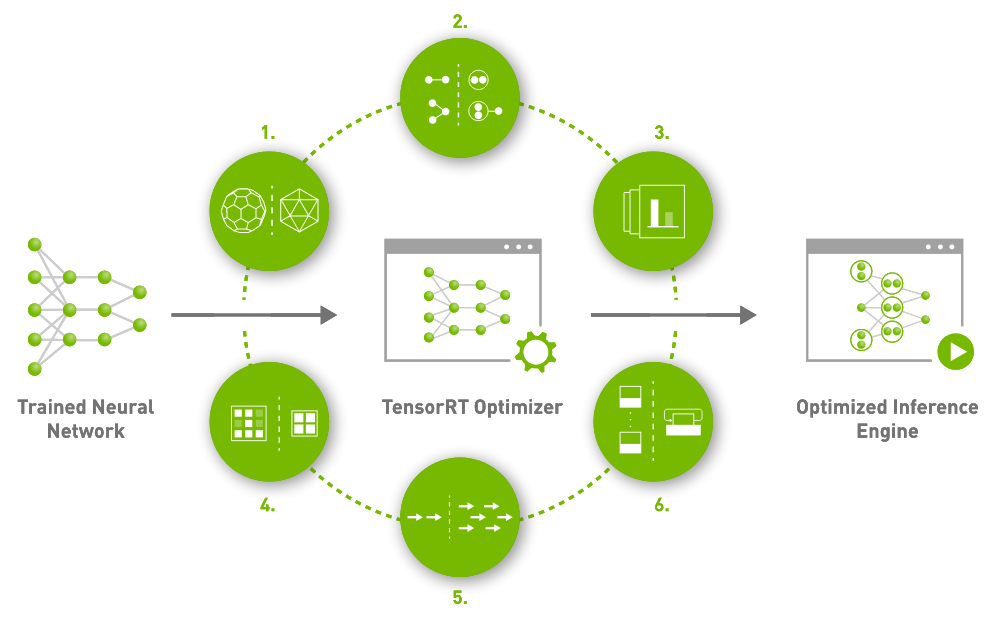}
\centering
\caption{Tensor RT pipeline. Source: NVIDIA.}
\label{fig:tensorrt}
\end{figure}


\chapter{Real-World Applications in Industry}  

\ifpdf
    \graphicspath{{Chapter6/Figs/Raster/}{Chapter6/Figs/PDF/}{Chapter6/Figs/}}
\else
    \graphicspath{{Chapter6/Figs/Vector/}{Chapter6/Figs/}}
\fi

In this Section we introduce some problems solved using the concepts introduced in Chapter \ref{computer_vision}.
These problems are focused on autonomous robots applications, and the main employed technology is Object Detection (introduced in Section \ref{sec:object-detecttion}).

\section{Vehicle Detection - Baidu Competition}
\label{baidu}

Baidu's Robotics and Autonomous Driving Lab (RAL), along with Peking University, organized a machine learning competition hosted on Kaggle \footnote{\url{https://www.kaggle.com/c/pku-autonomous-driving}}. The provided dataset consisted in more than 60,000 labeled 3D car instances from 5,000 real-world images captured from the roof of a car (6 GB of size) \cite{song2019apollocar3d}. The train-test split was 4000 and 1000 respectively. The challenge consisted in developing an algorithm to estimate the absolute pose of vehicles (6 degrees of freedom) from a single image in a real-world traffic environment.

For each car in the training set we have the pose information ground-truth formatted as strings, as follows: \begin{verbatim} model type, yaw, pitch, roll, x, y, z
\end{verbatim}

Bellow we show an example of the pose information string assuming there are two cars in the photo (car's models are 10 and 23):

\begin{verbatim}
10 0.5 0.5 0.5 0.0 0.0 0.0 23 0.25 0.25 0.25 0.5 0.4 0.7
\end{verbatim}

The authors of this dataset, the Apolloscapes Dataset, open-sourced a developer kit \footnote{\url{https://github.com/ApolloScapeAuto/dataset-api/tree/master/car_instance}} to read the images and operate with them (plots, transformations,etc).

We aim to predict the position of the cars that appear in a given image, meaning detecting the cars is not enough, we also need to provide the position estimation and the confidence score. The following is an example of a prediction for an image with ID 1d7dd9b31 and 0.99 confidence score :

\begin{verbatim}
ID_1d7dd9b31,0.5 0.5 0.5 0.0 0.0 0.0 0.99
\end{verbatim}

This is a very challenging problem because of the following reasons: 
\begin{enumerate}
    \item RGB images have resolution $2710 \times 3384$, such data size makes impossible for us to use them directly for training neural networks, therefore, we need to pre-process the images and reduce their dimension.
    \item From the mathematical point of view, this is a fundamental ill-posed problem because the 2D images representation of the 3D world, does not contain information about the angle nor depth, thus, predicting 3D information from 2D is really difficult and there is no unique (or exact) solution.
\end{enumerate}

\begin{figure}[!h] 
\centering    
\includegraphics[width=0.65\textwidth]{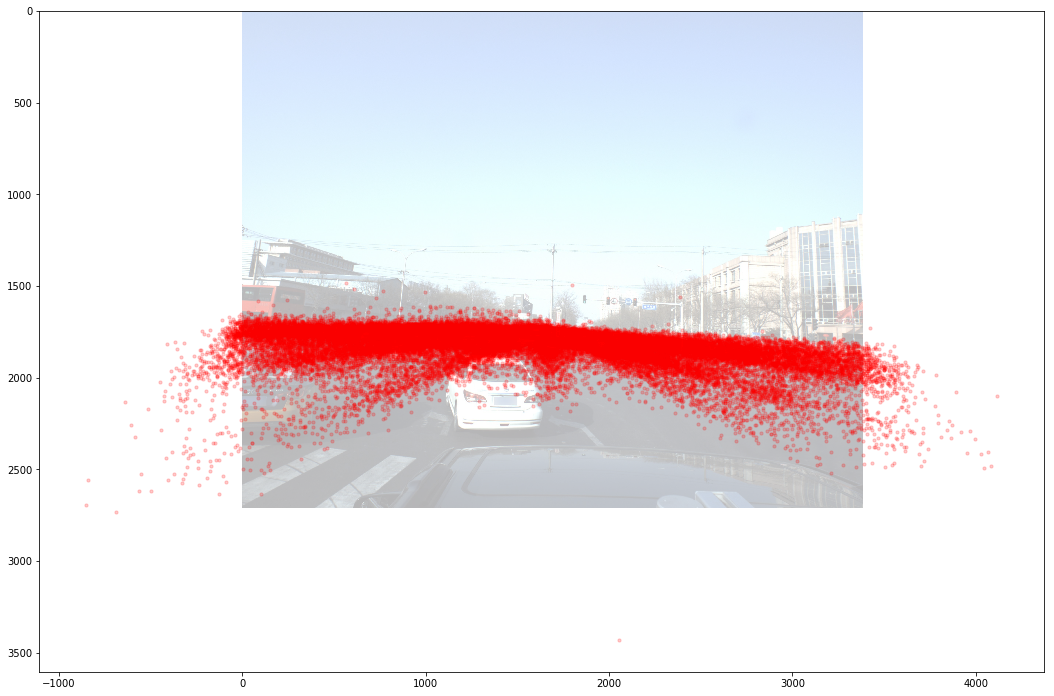}
\caption[AD Pose Projections]{Projection of all the possible 2D points (car positions) in the images.}
\label{fig:baidu-points}
\end{figure}

To solve the sub-problem 1, we plot the distribution of all possible points (positions of all the cars in the training data) as shown in Figure \ref{fig:baidu-points}. After doing this, we decide to crop the images as much as possible attending to the distribution of points, thus, all the images are cropped to a size $700 \times 1800$.

\begin{figure}[!h] 
\centering    
\includegraphics[width=0.8\textwidth]{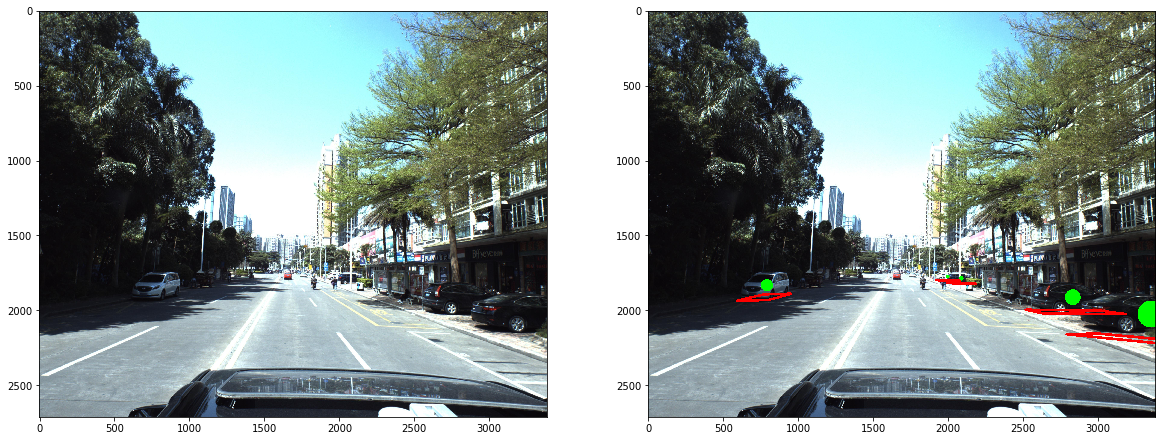}
\caption[AD Car Position Estimation - Sample 1]{AD Car Position Estimation - Sample 1}
\label{fig:add-app1}
\end{figure}

\begin{figure}[!h] 
\centering    
\includegraphics[width=0.8\textwidth]{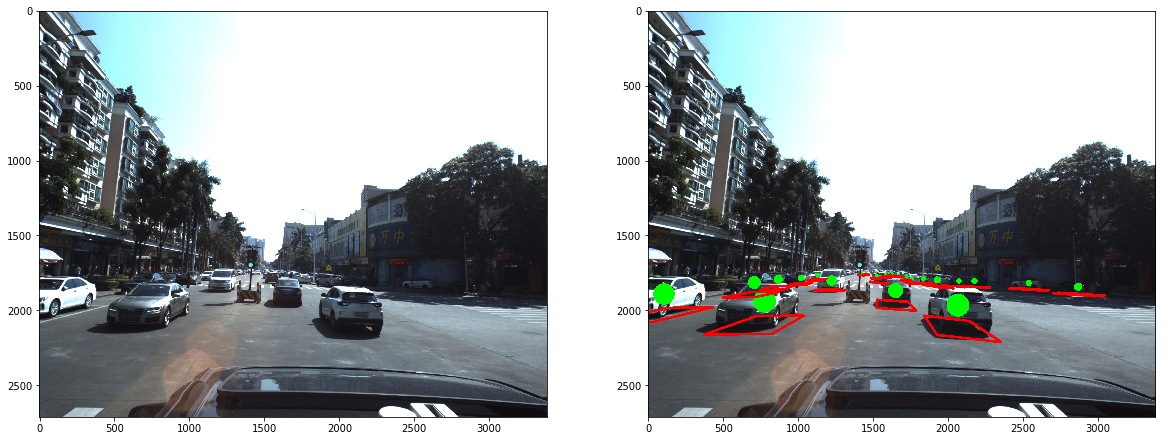}
\caption[AD Car Position Estimation - Sample 2]{AD Car Position Estimation - Sample 2.}
\label{fig:add-app2}
\end{figure}

The second sub-problem cannot be solved, for this reason we consider it a reminder about the limitation of the possible solutions.
In Figures \ref{fig:add-app1} and \ref{fig:add-app2} we show examples if the input image and predicted vehicle's position.
Note that position is given as 6DoF, therefore, a transformation is required to plot positions as 2D coordinates into the images, this transformation requires the coordinates (x,y,z) and camera information.
The code for performing such transformations is provided by the competition's host \footnote{\url{https://github.com/ApolloScapeAuto/dataset-api/tree/master/car_instance}}.

\begin{figure}[!ht] 
\centering    
\includegraphics[width=0.8\textwidth]{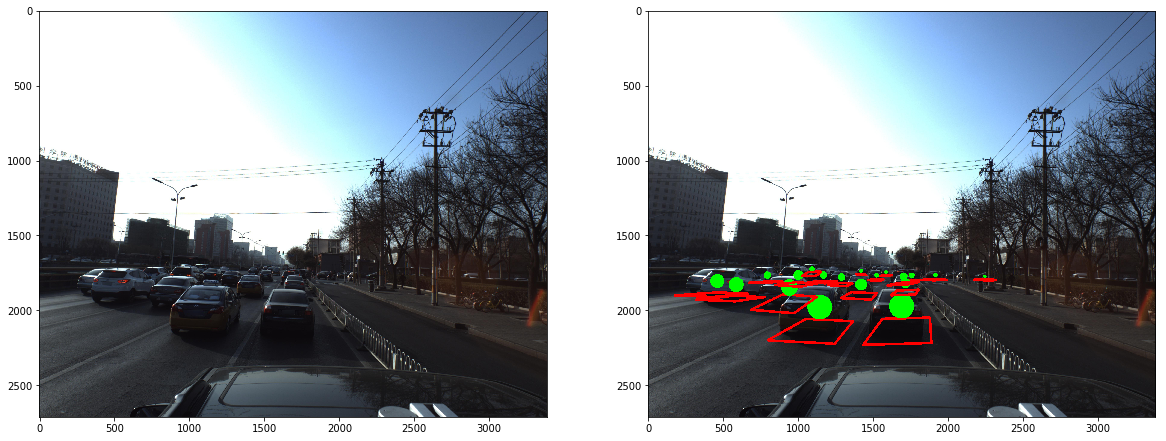}
\caption[AD Car Position Estimation - Sample 3]{AD Car Position Estimation - Sample 3}
\label{fig:add-app3}
\end{figure}

\begin{figure}[!ht] 
\centering    
\includegraphics[width=0.9\textwidth]{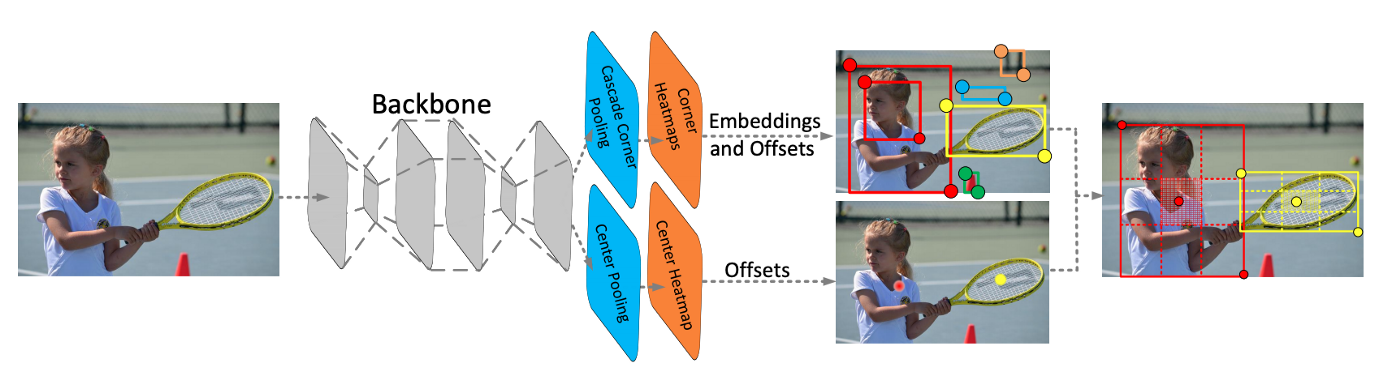}
\caption{CenterNet architecture \cite{duan2019centernet}.}
\label{fig:centernet-arch}
\end{figure}

\begin{figure}[!ht] 
\centering    
\includegraphics[width=0.9\textwidth]{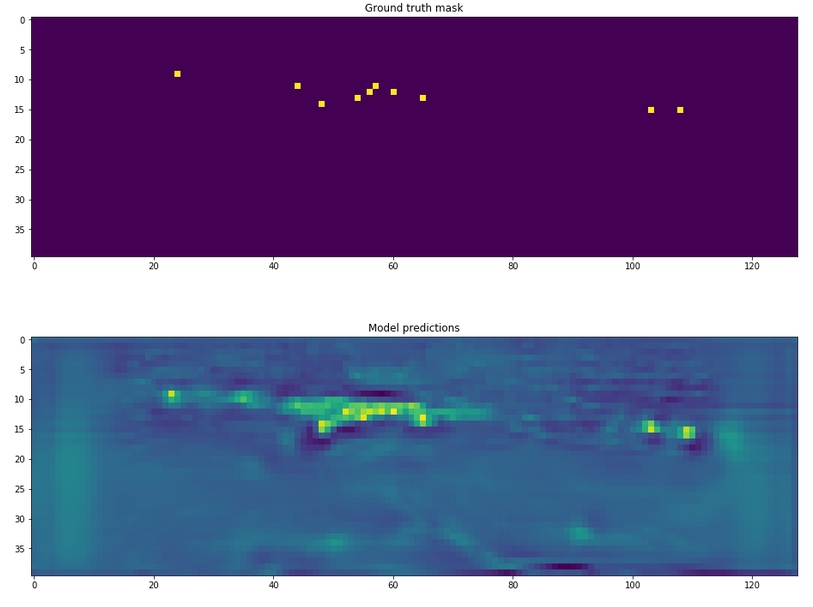}
\caption{CenterNet heatmaps (bottom) and offset point predictions (top).}
\label{fig:center-map}
\end{figure}

\begin{figure}[!h]
    \centering
    \setlength{\tabcolsep}{2.0pt}
    \begin{tabular}{c}
    \includegraphics[width=0.7\textwidth]{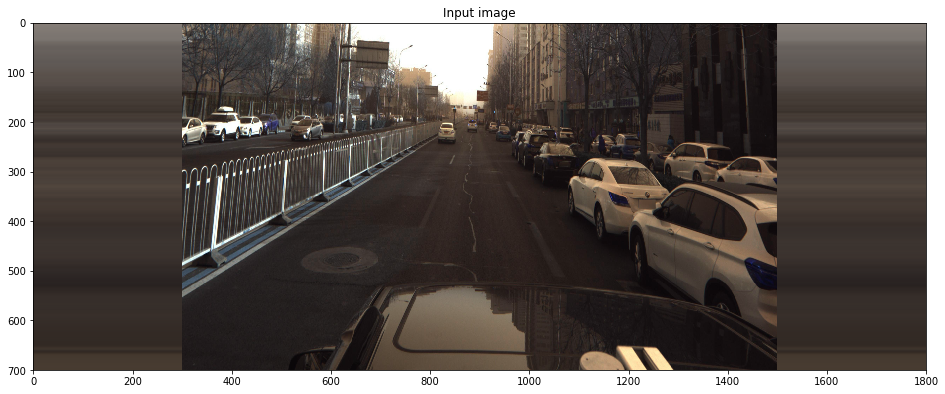} \tabularnewline
    \includegraphics[width=0.7\textwidth]{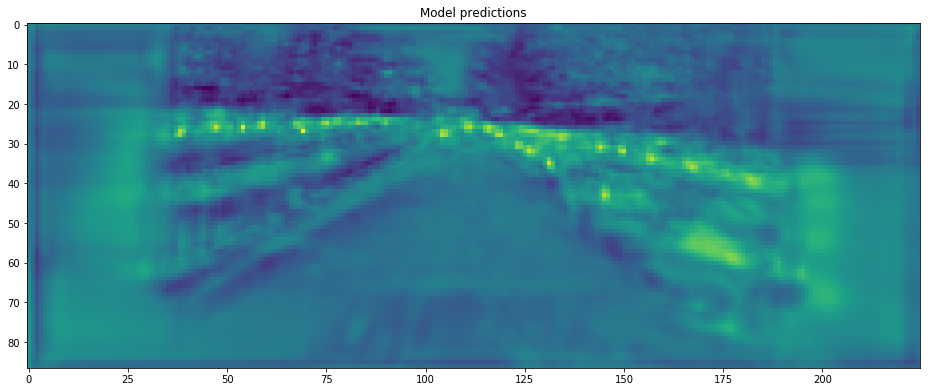} \tabularnewline
    \includegraphics[width=0.7\textwidth]{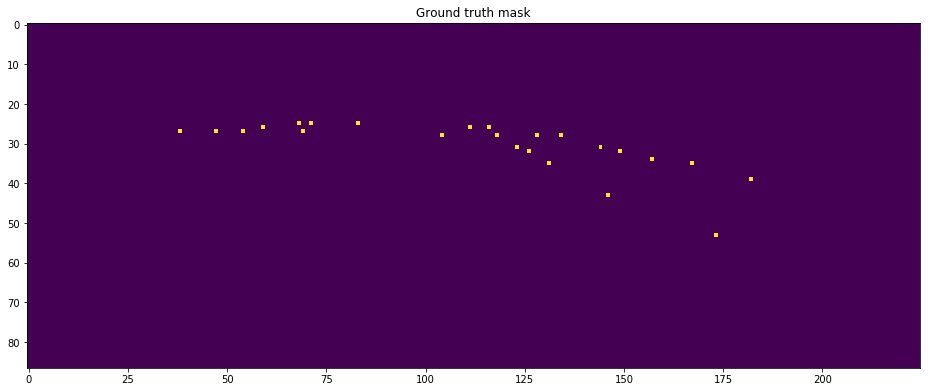} \tabularnewline
    \includegraphics[width=0.7\textwidth]{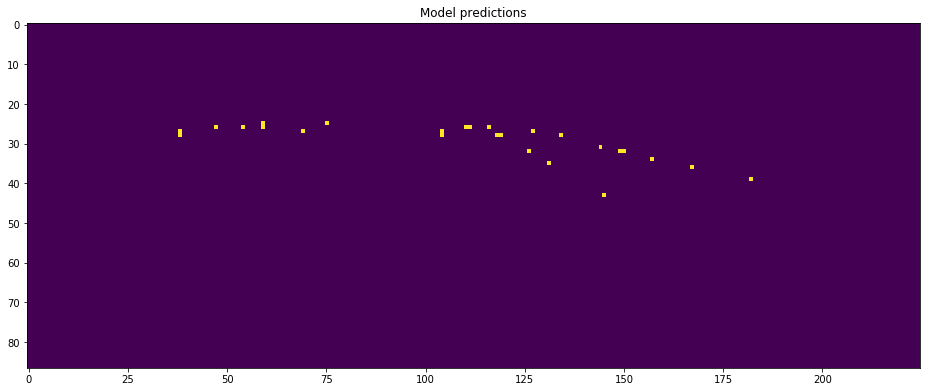} \tabularnewline
    \end{tabular}
    \caption{Visualization of the vehicles pose predictions. From top to bottom: original input image, heatmap produced by our network, gound-truth (real values), predictions (after applying threshold to the heatmap).}
    \label{fig:predictions-map}
\end{figure}

\clearpage
Our approach to solve this problem is to use a neural network for predicting centers of objects as heatmaps. This is mainly a regression problem where the model detects the vehicles in a RGB image while simultaneously regressing their 3D translation and rotation vectors \cite{6dvnet,wang20196pack-6dof-ad}.
We use as baseline the object detector network CenterNet \cite{duan2019centernet, zhou2019objects}, introduced in Section \ref{sec:object-detecttion}. The network architecture is shown in Figure \ref{fig:centernet-arch}, the backbone or feature extractor is a CNN that will extract a feature map (heatmap), in our case this CNN is a ResNet34 (can be any of its variants) \cite{resnet}, the map goes through a pooling layer and we obtain the offsets or predictions. In Figures \ref{fig:center-map}, \ref{fig:predictions-map} we show the feature maps (heatmaps) and predicted points.

The model is optimized using Adam \cite{kingma2017adam} optimizer with default parameters, and the $\textit{L}_1$ loss function for pose estimation, this is also used as performance metric together with the MSE and mean average precision (MAP) between the predicted pose information and the correct position and rotation.
For calculating the MAP, he distances between prediction and solution are less than the threshold $T$, then that prediction object is counted as a true positive (TP), if not the predicted object is counted as a false positive (FP), finally, MAP is calculated using these TP/FP determinations \footnote{\url{https://www.kaggle.com/c/pku-autonomous-driving/overview/evaluation}}.

Note that the loss function is not calculated strictly between the true position value and the model's output, we use a gaussian filter to smear the real points, generating a smooth distribution that reaches a peak at object centers, and thus, making the learning process fully differentiable. Moreover, we modify slightly CenterNet original architecture, and we use as backbone a custom UNet with a ResNet34 Encoder \cite{ronneberger2015unet} architecture as feature extractor.

\begin{figure}[!ht] 
\centering    
\includegraphics[width=\textwidth]{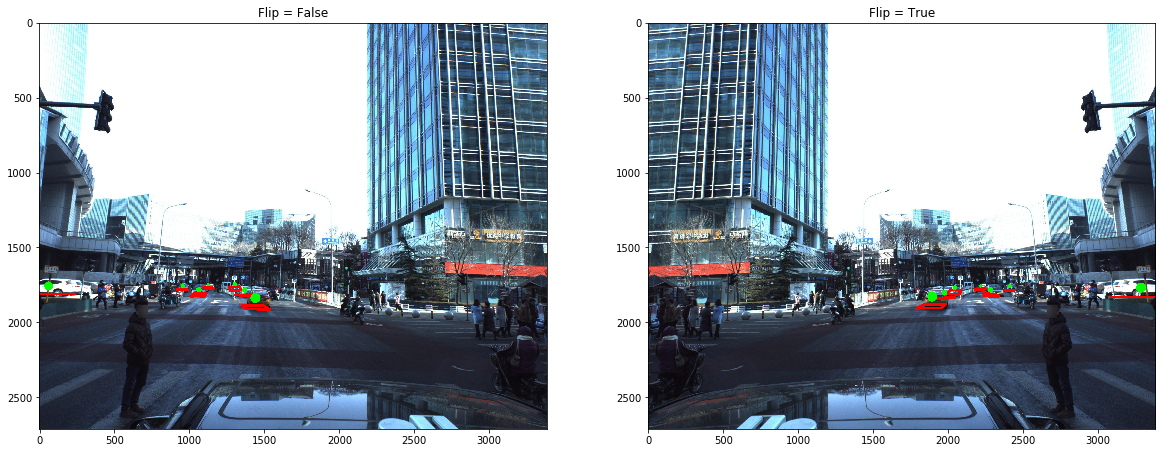}
\caption[Autonomous Driving - Flip Augmentation]{Autonomous Driving - Flip Augmentation. (Left) real image, (right) flipped image}
\label{fig:apolo-augmentation1}
\end{figure}

\begin{figure}[!ht] 
\centering    
\includegraphics[width=\textwidth]{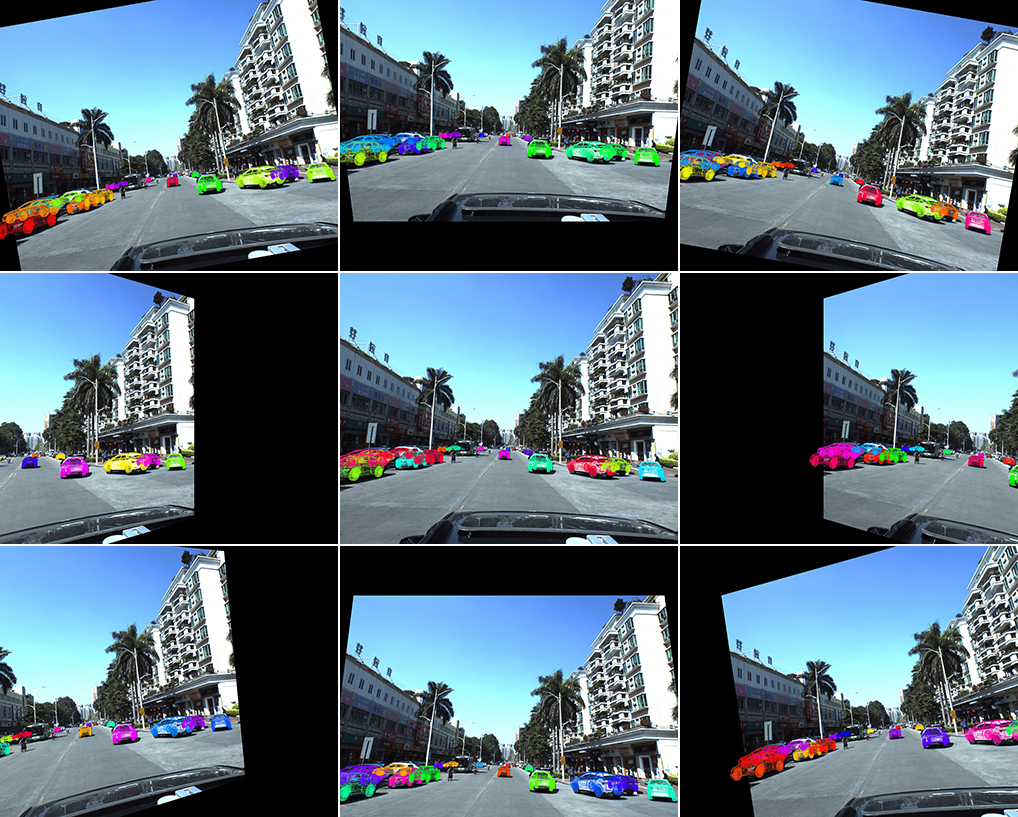}
\caption[Autonomous Driving - Rotation Augmentation]{Autonomous Driving - Rotation Augmentation. The central image is the original. Code from \url{https://www.kaggle.com/outrunner/rotation-augmentation}. }
\label{fig:apolo-augmentation2}
\end{figure}

To improve models performance, we use augmentations as introduced in Section \ref{sec:augmentations}.
The most basic augmentation that we use is the Horizontal Flip (see Figure \ref{fig:apolo-augmentation1}), however, other affine transformations and rotations can generate more diversity and imply better performance (see Figure \ref{fig:apolo-augmentation2}).

\clearpage

Finally, we show some visualizations of our predictions.

\begin{figure}[!h]
    \centering
    \setlength{\tabcolsep}{2.0pt}
    \begin{tabular}{c}
    \includegraphics[width=0.8\textwidth]{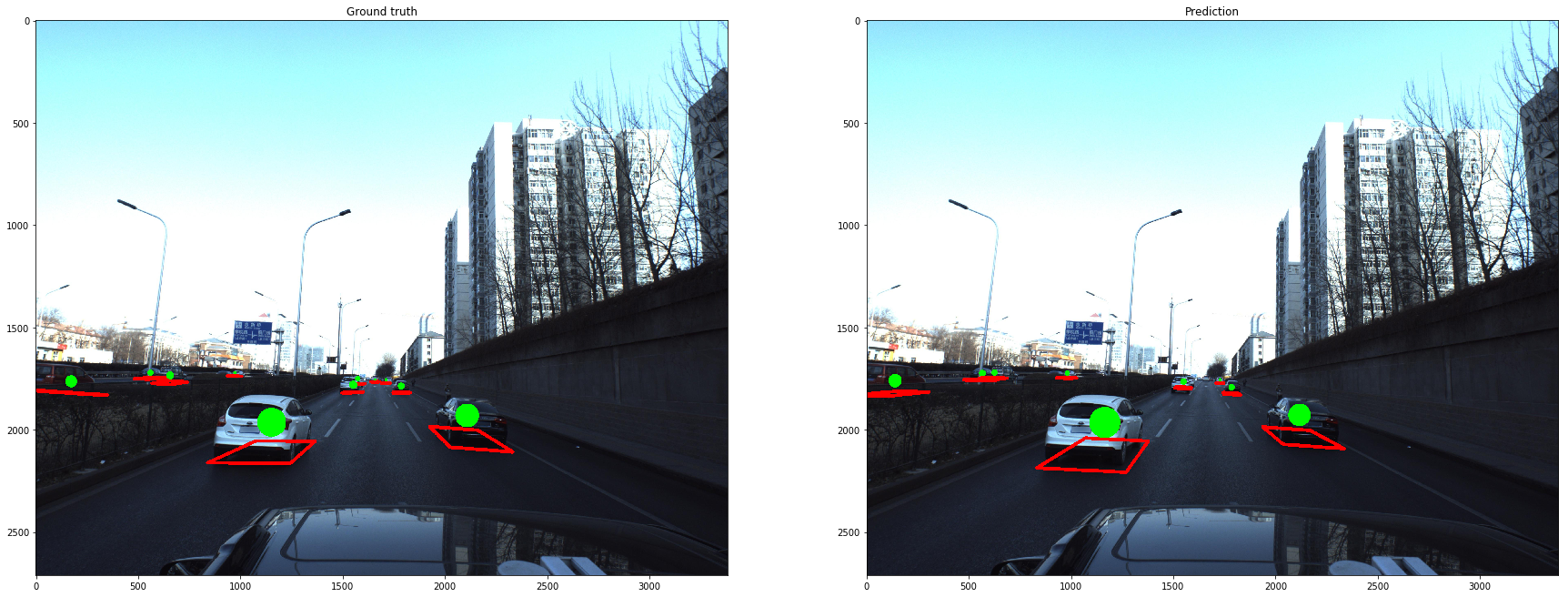} \tabularnewline
    \includegraphics[width=0.8\textwidth]{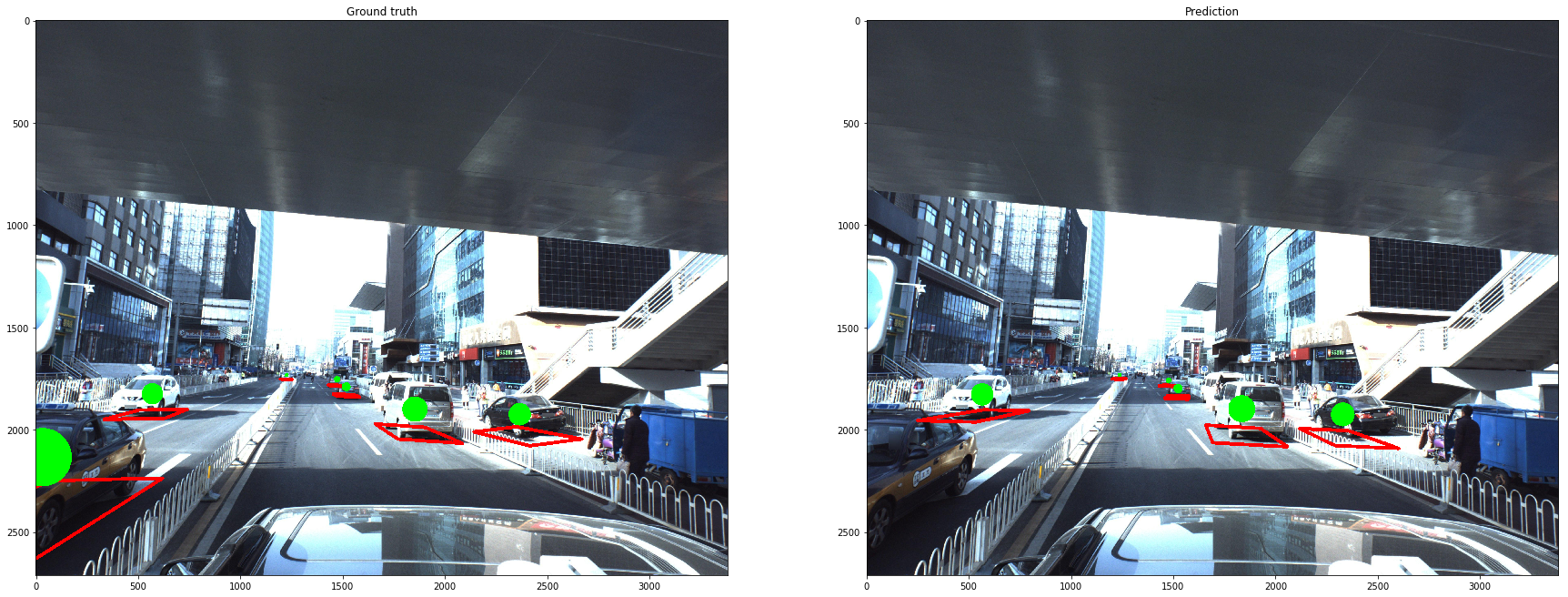} \tabularnewline
    \includegraphics[width=0.8\textwidth]{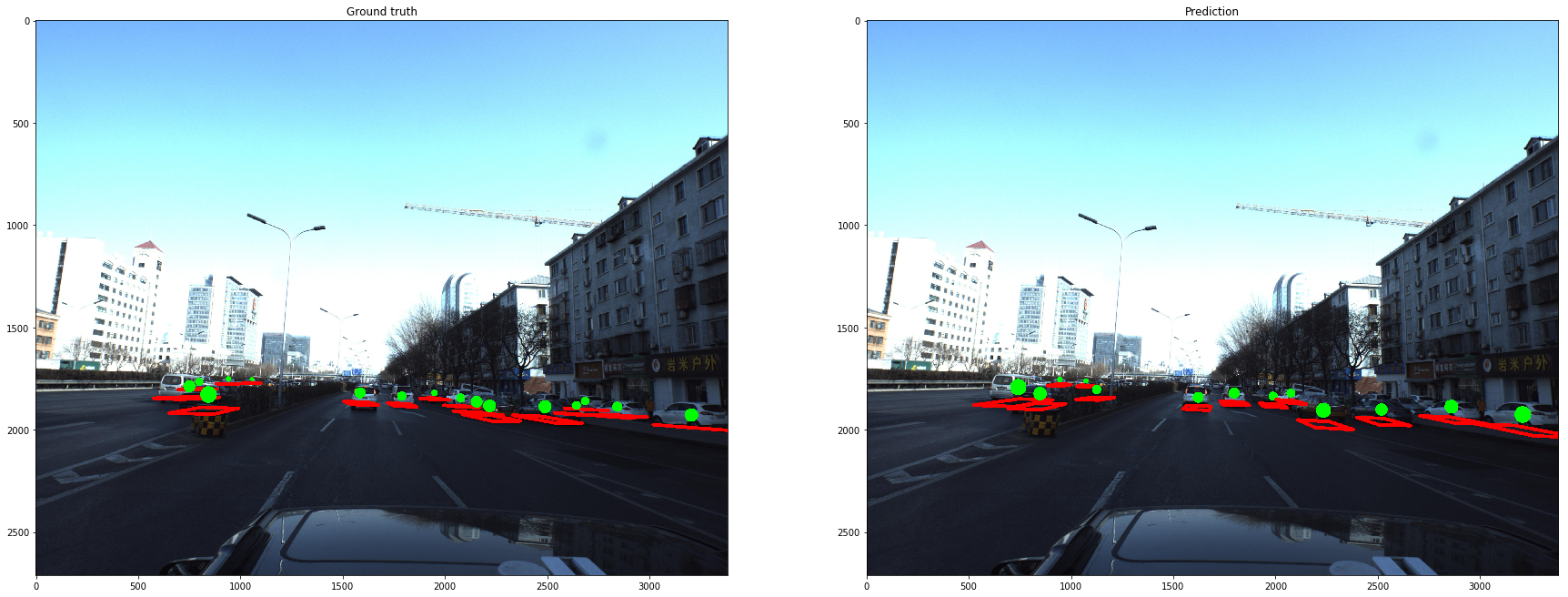} \tabularnewline
    \includegraphics[width=0.8\textwidth]{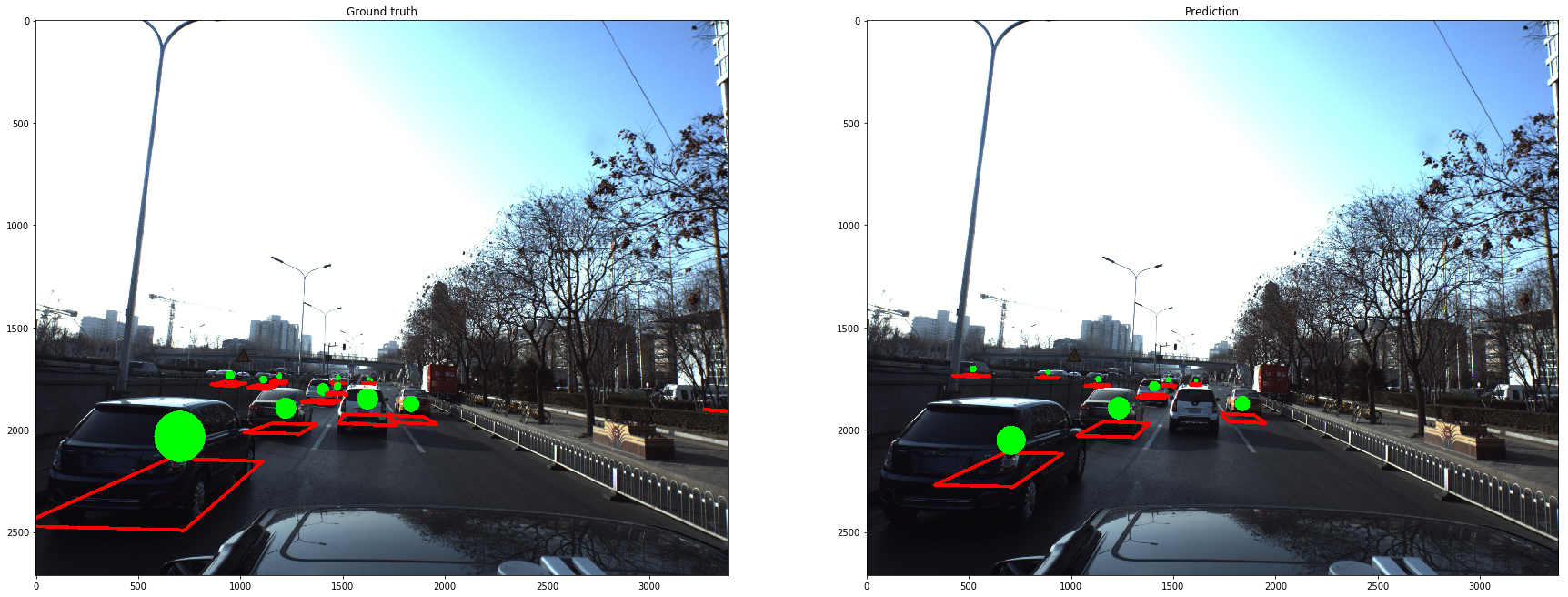} \tabularnewline
    \end{tabular}
    \caption{Visualization of vehicles (left) real position, (right) the predictions (2D projection).}
    \label{fig:predictions-cars}
\end{figure}

\clearpage

\begin{figure}[!ht] 
\centering    
\includegraphics[width=0.8\textwidth]{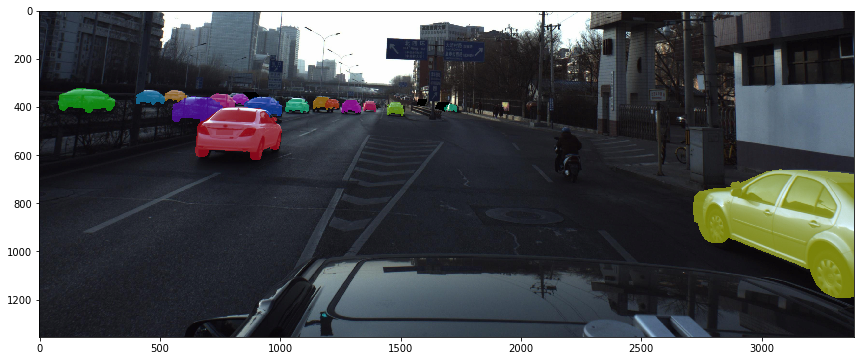}
\caption[Apollo Scape Segmentation]{Apollo Scape Vehicle Segmentation}
\label{fig:apolo-segmentation}
\end{figure}

The ApolloScape Dataset \cite{song2019apollocar3d} also includes vehicle's mask (see Figure \ref{fig:apolo-segmentation}), this allows to explore other approaches based on segmentation and not only detection. 

\begin{figure}[!ht] 
\centering    
\includegraphics[width=0.8\textwidth]{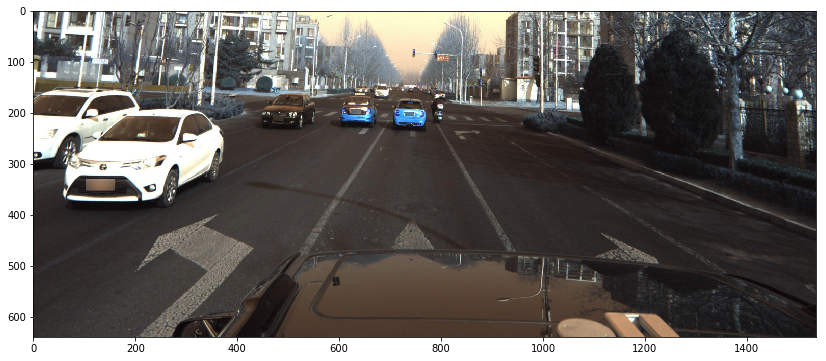}
\caption[Apollo Scape Image from Vehicle]{Apollo Scape Image from Vehicle}
\label{fig:apolo-segment1}
\end{figure}

\begin{figure}[!ht] 
\centering    
\includegraphics[width=0.8\textwidth]{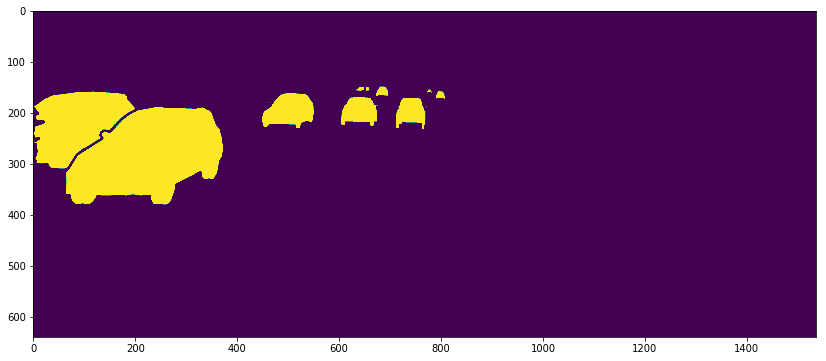}
\caption[Apollo Scape Segmentation Mask]{Apollo Scape Segmentation Mask}
\label{fig:apolo-segment2}
\end{figure}

\clearpage
\section{Maintenance and Monitoring} 

\begin{figure}[!h] 
\centering    
\includegraphics[width=0.6\textwidth]{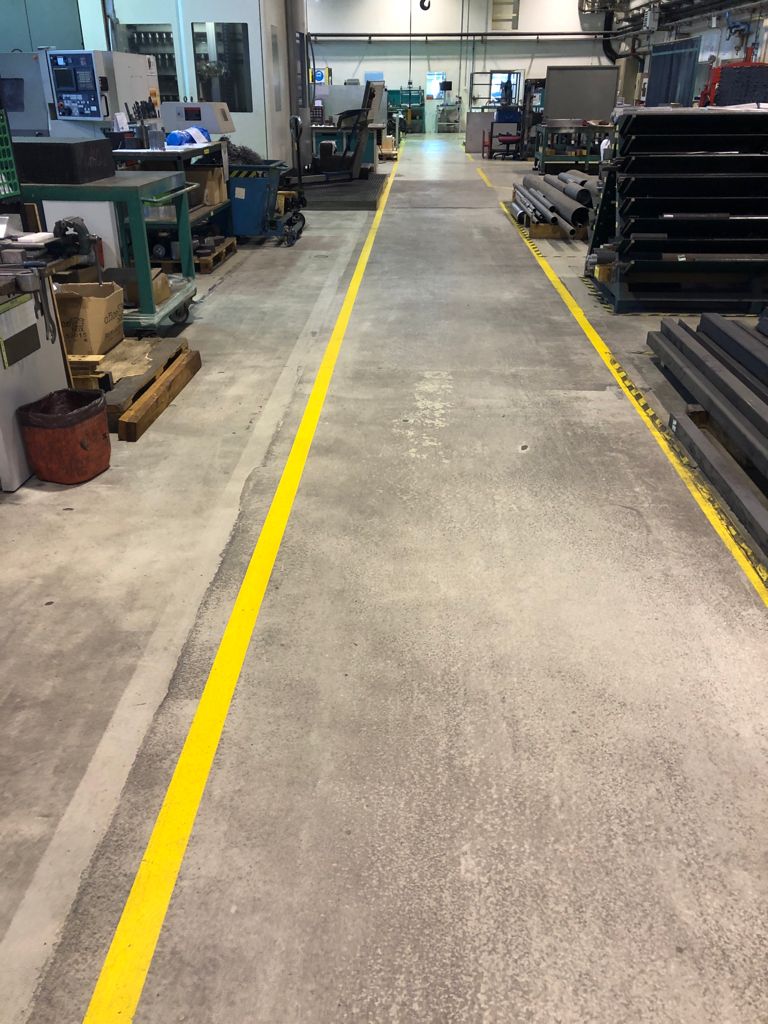}
\caption[Cidaut Use Case 1]{Cidaut Use Case 1.}
\label{fig:cidaut1}
\end{figure}

\begin{figure}[!h] 
\centering    
\includegraphics[width=\textwidth]{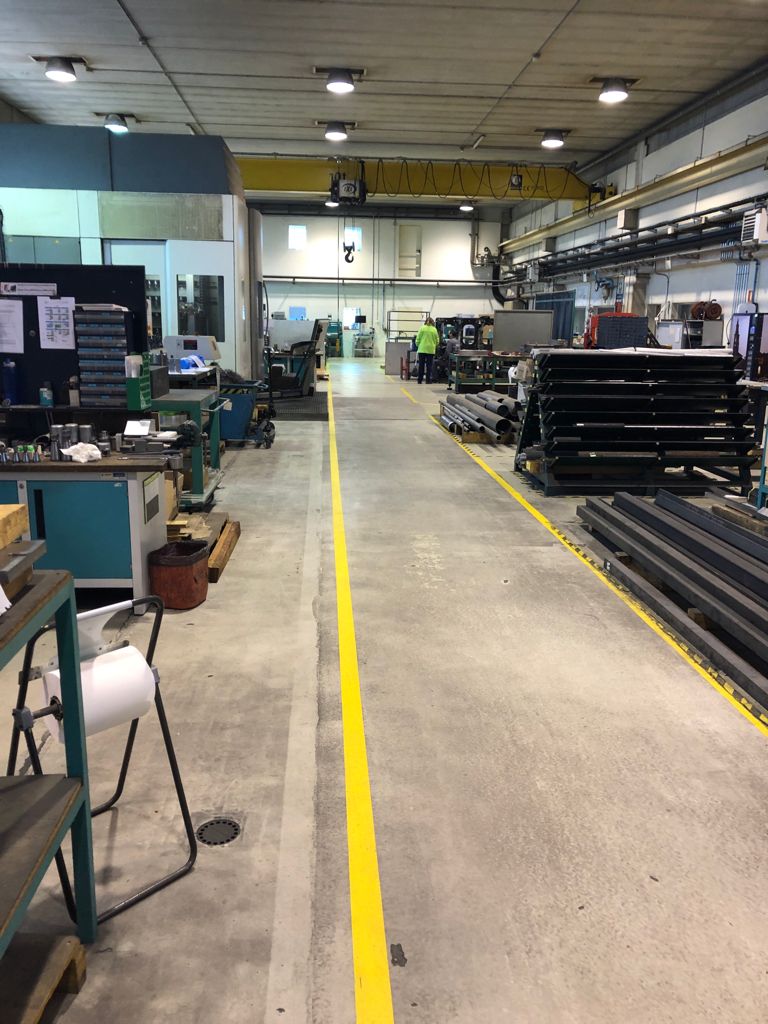}
\caption[Cidaut Use Case 2]{Cidaut Use Case 2.}
\label{fig:cidaut2}
\end{figure}

In this part of the section, we explore possible applications of the concepts mentioned in previous sections.
CIDAUT is a Spanish pioneer in Industry 4.0, for this reason, they demand AI solutions for common problems in industrial plants as: count the number of persons in the area, detect if there are goods or spare parts when they are needed, monitor temperature and air, etc.
The proposed solution in this project, based on those needs, is a small robot able to navigate with autonomy through the factory environment that we show in Figures \ref{fig:cidaut1} and \ref{fig:cidaut2}. 

As we can see in the figures, the robot would have its own lane to operate, and reference yellow lines in the floor, therefore, a simple navigation system based on lane-tracking and obstacle avoidance is enough and common for AGVs.
Because of privacy laws and Cidaut's regulations, we cannot provide more images about the real environment or applications that involve employees (i.e. persons detection).

Our inspiration to solve this problem is ANYmal from the Swiss company ANYbotics, previously introduced in Chapter \ref{introduction}. This robot can navigate with total autonomy through unknown industrial environments thanks to its mapping and localization capabilities, also can plan certain paths and check components (dashboards, pipelines, etc). From the computer vision point of view, its main features are: Object Detection and SLAM (Simultaneous Localization and Mapping), both reinforced by the robot's sensors (special camera sensor, LiDAR radar, etc).

\begin{figure}[!h] 
\centering    
\includegraphics[width=0.8\textwidth]{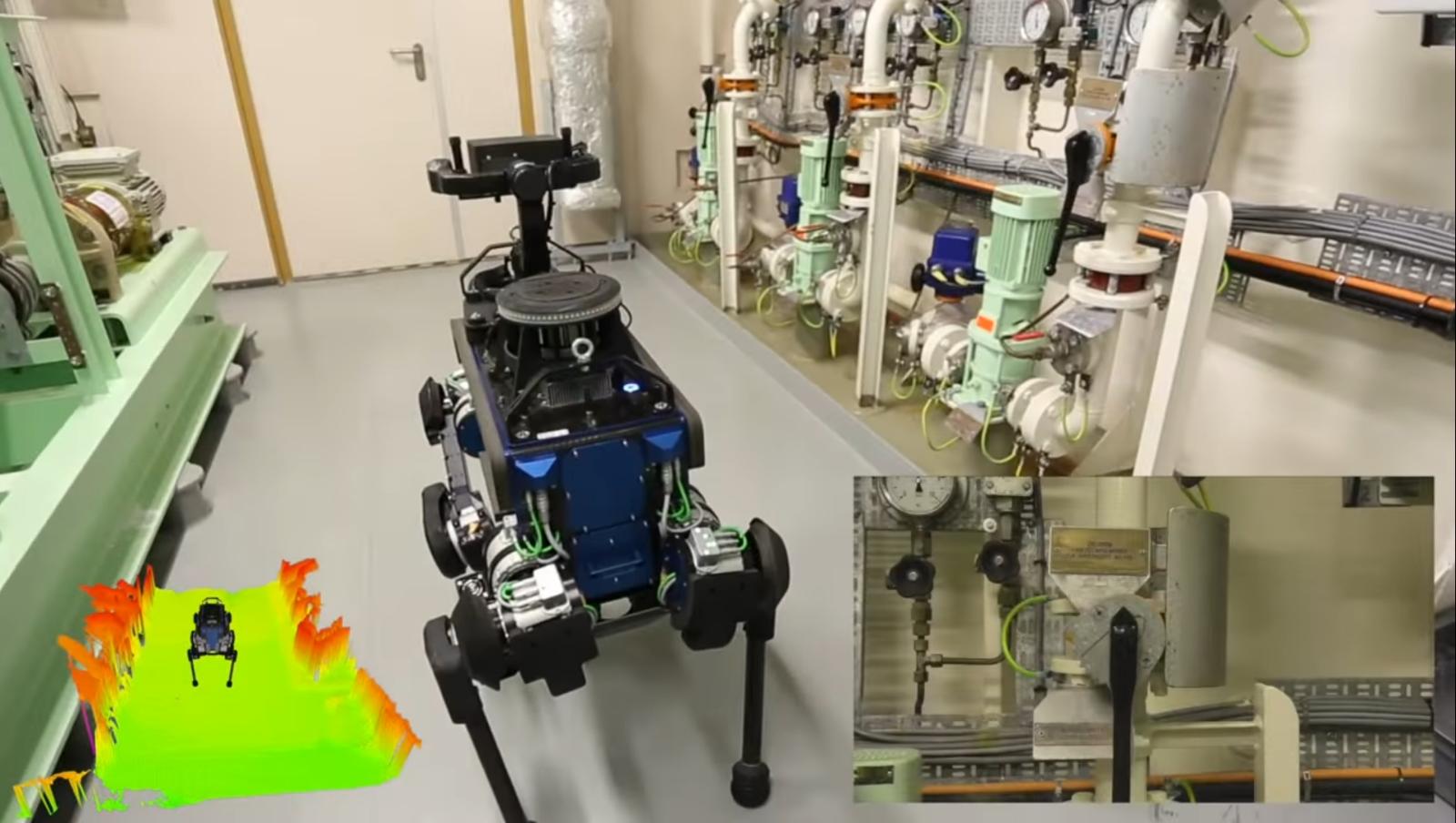}
\caption[ANYmal working - sample 1]{ANYmal working - sample 1}
\label{fig:anymal-w1}
\end{figure}

\begin{figure}[!h] 
\centering    
\includegraphics[width=0.8\textwidth]{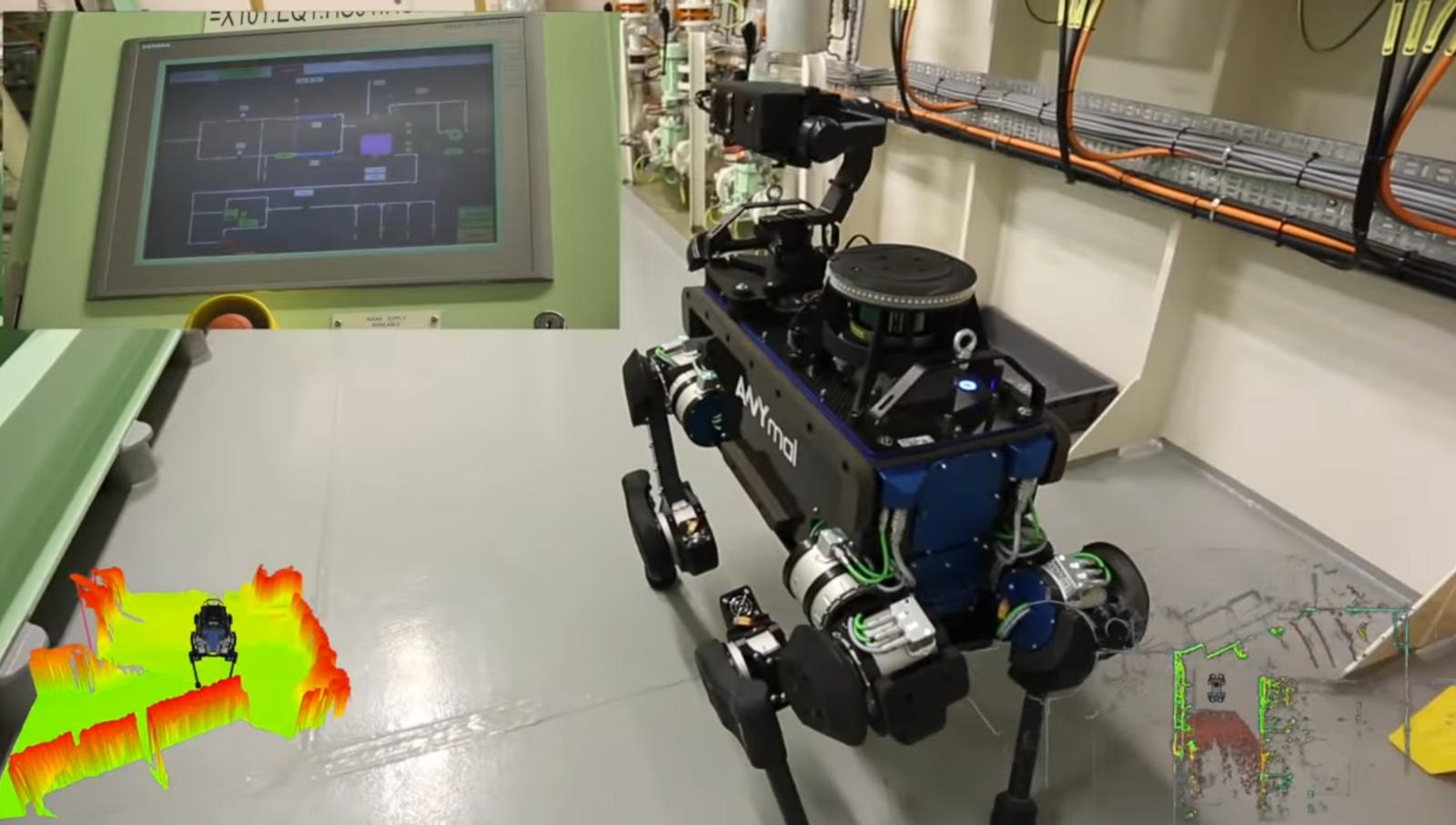}
\caption[ANYmal working - sample 2]{ANYmal working - sample 2}
\label{fig:anymal-w2}
\end{figure}

\begin{figure}[!h] 
\centering    
\includegraphics[width=0.8\textwidth]{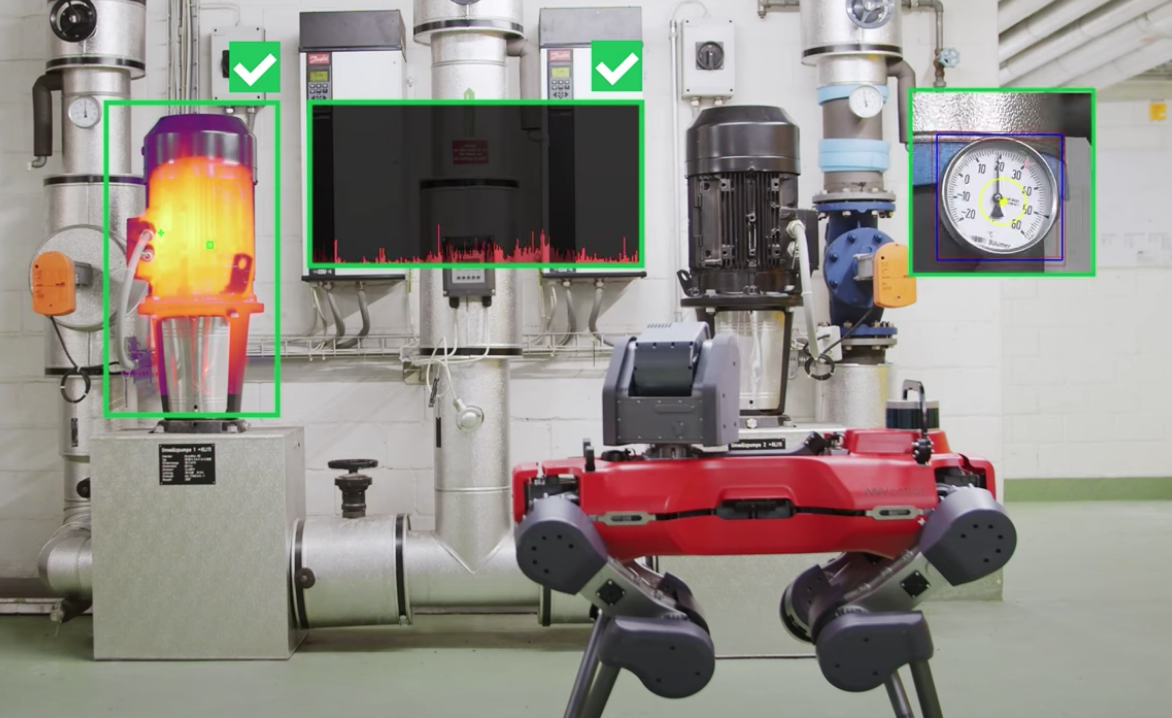}
\caption[]{ANYmal Monitoring Demo.}
\label{fig:anymal-w3}
\end{figure}

\begin{figure}[!h] 
\centering    
\includegraphics[width=0.8\textwidth]{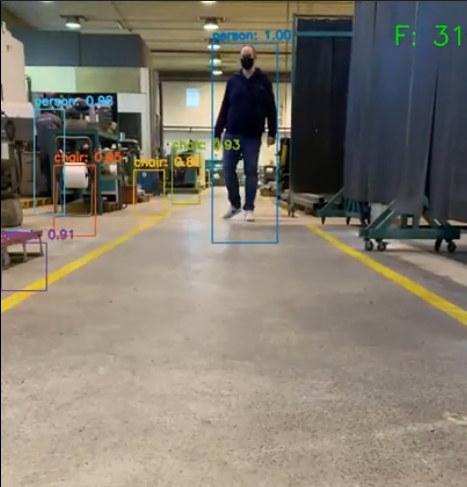}
\caption[Cidaut Detection and Monitoring]{Cidaut Detection and Monitoring results using YoloV4~\cite{bochkovskiy2020yolov4}.}
\label{fig:cidaut-result1}
\end{figure}

\begin{figure}[!h] 
\centering    
\includegraphics[width=0.8\textwidth]{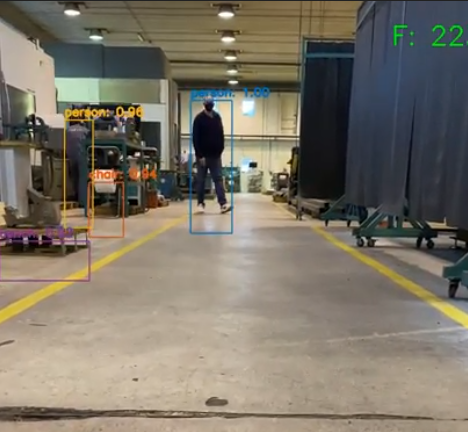}
\caption[]{Cidaut Detection and Monitoring results using CenterNet~\cite{duan2019centernet}.}
\label{fig:cidaut-result2}
\end{figure}


To ''replicate`` ANYmal performance using our DIY low-cost robot (see Appendix \ref{app:robot}) we propose a multi-task approach based on the following visual tasks:

\begin{enumerate}
    \item Objects Detection: the robot needs to detect persons, possible obstacles and key objects (i.e. industrial material to monitor). There are plenty of different ways for achieving this, in Section \ref{sec:object-detecttion} we introduce the Detection problem and in Section \ref{baidu} we show our solution to one of the hardest detection problems. The JetBot platform includes base object detector methods as SSD (Single Shot MultiBox Detector) \cite{Liu_2016-ssd-detector} (SSD-Mobilenet and SSD-Inception) pretrained on the MS-COCO Dataset \cite{lin2015microsoft-coco-dataset}. This pre-trained detectors are ready to use, since they have been training on the MS-COCO Dataset, the model is able to detect persons, animals, boxes, etc.
    The first option that we contemplate is to retrain these models on new data captured in our environment, the second option is to try SOTA detectors as YOLOv4 \cite{bochkovskiy2020yolov4, redmon2016look-yolo} or CenterNet~\cite{duan2019centernet} (previously introduced). YoloV4 offers better real-time performance than the other methods, and it is also a pre-trained model \footnote{\url{https://github.com/jkjung-avt/tensorrt_demos}}, therefore, it already has basic knowledge about persons and objects recognition. Our agent, as programmed in the official JetBot project, will react depending on the visual information usinf a neural network \footnote{\url{https://jetbot.org/master/examples/collision_avoidance.html}}. We test YoloV4 and CenterNet and show results in Figures~\ref{fig:cidaut-result1} and \ref{fig:cidaut-result2}.
    
    \item Free-space segmentation: The goal of the free-space network is to segment images into two classes of interest: drivable space and obstacles. The input of the network is a RGB image, and the output is pixel-wise segmentation.
    We use this technique as an alternative for obstacle avoidance using a monocular RGB camera sensor. The free space map determined by the segmentation model can be projected onto the real world coordinate system and used as input information for obstacle avoidance.
    
    \item Lane Tracking: as an alternative to pure path planning navigation, we use JetBot's model for lane tracking. The model is trained for regressing $x,y$ lane positions, the input is an RGB image and the output a predicted position $\hat{x}, \hat{y}$ in the lane. More specifically, the base model is a ResNet-18 \cite{resnet} as image feature extractor followed by a MLP with 2 outputs.

\end{enumerate}

This way, the robot can operate using only visual information, without requiring previous knowledge about the plant (i.e. maps). Other techniques like monocular depth estimation \cite{wang2020depthnet} where we estimate depth from 2D images are crucial for applications such as 3D scene understanding and recognition, therefore, their integration in our multi-task system would be very positive as it is another way to pseudo-measure distances and better perceive the environment.

\begin{figure}[!h] 
\centering    
\includegraphics[width=\textwidth]{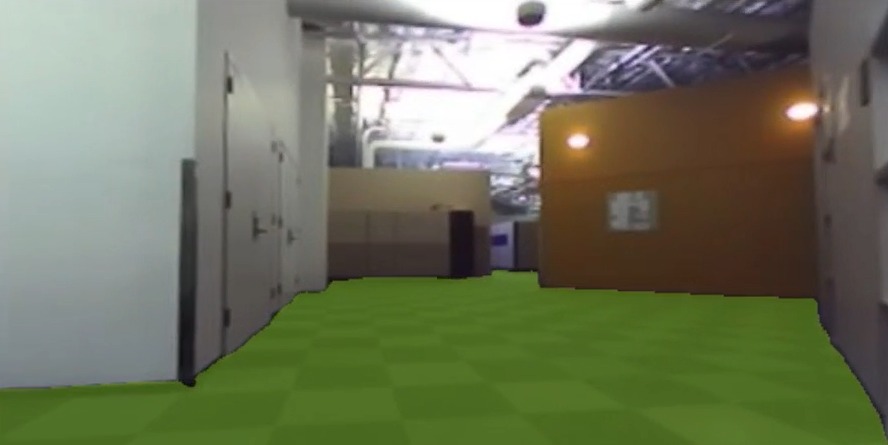}
\caption[Free-space Segmentation from NVIDIA Jetson's SDK]{Free-space Segmentation from NVIDIA Jetson's SDK}
\label{fig:free-segment-nvidia}
\end{figure}

\begin{figure}[!h] 
\centering    
\includegraphics[width=\textwidth]{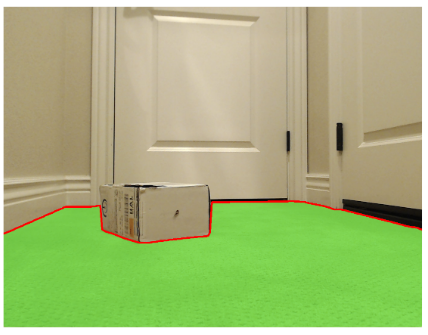}
\caption[Free-space Segmentation method]{Free-space Segmentation using network proposed by Singhani \textit{et.al} \cite{singhani2019realtime-freespace}}
\label{fig:free-segment}
\end{figure}


\begin{figure}[!h] 
\centering    
\includegraphics[width=\textwidth]{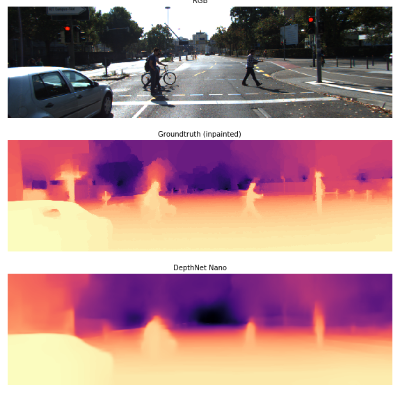}
\caption{DepthNet Nano - Monocular Depth Estimation \cite{wang2020depthnet}.}
\label{fig:depth}
\end{figure}

\clearpage

\section{Covid-19 Safety Measures in Industry} 

Covid-19 has accelerated AI integration into airports, public places and private companies. In this Section, we propose some ideas that we consider helpful for controlling Covid-19 in industry and other indoor environments.

\begin{figure}[!h] 
\centering    
\includegraphics[width=\textwidth]{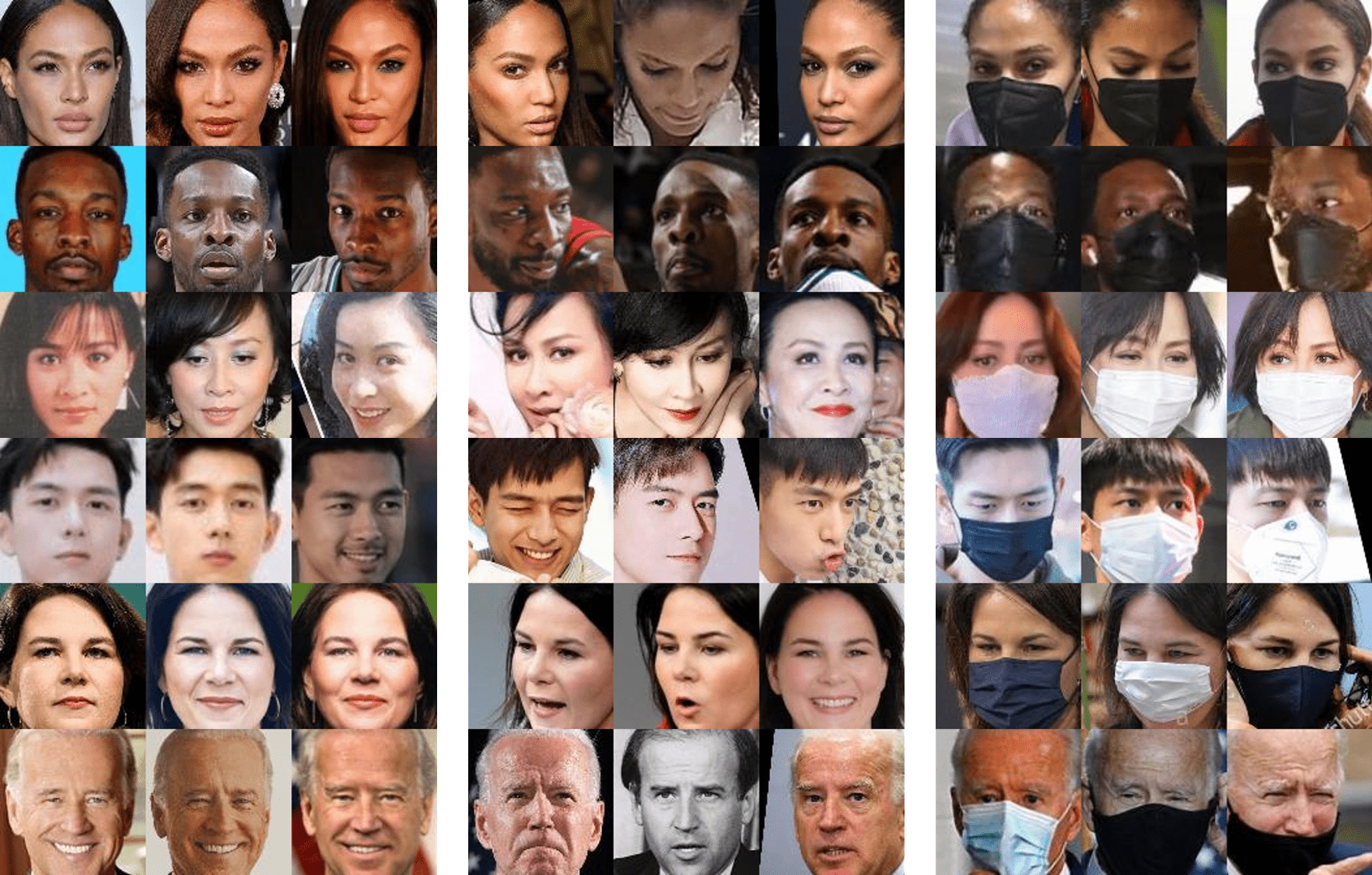}
\caption[FaceMask Challenge Test sample]{FaceMask Challenge Test sample}
\label{fig:face-test}
\end{figure}

Wearing facemask has been proved to be the most important safety measure against Covid, recently, airports have integrated automatic systems for controlling facemask wearing.
For instance, how can we create an automatic system that given a face from Figure \ref{fig:face-test} image could determine if the person is wearing a mask?

\begin{figure}[!h] 
\centering    
\includegraphics[width=\textwidth]{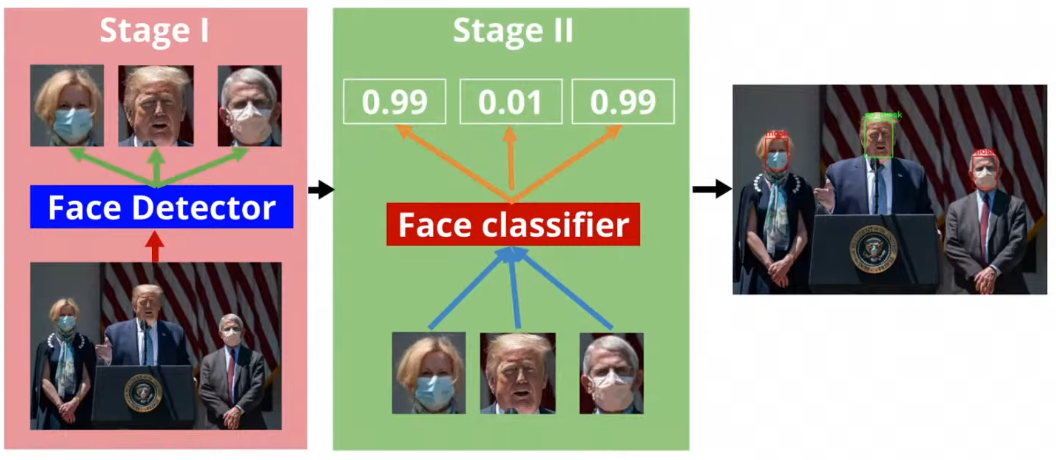}
\caption[FaceMask 2-stage approach]{FaceMask 2-stage approach: (i) we localize the faces, (ii) we classify with a different network each face separately. Source: Vladimir Iglovikov.}
\label{fig:trump}
\end{figure}

There 2 ways to solve this problem:
\begin{enumerate}
    \item 1 stage: we use a multi-task object detector network, ideally pre-trained in the face detection task as as Retinaface \cite{deng2019retinaface}. This detector can distinguish objects of 2 different classes: waring facemasks and no-facemask.
    In this approach we only train and use one model for predicting from a given image (even if the image contains multiple faces), therefore, the inference process is really fast, something important if we want to integrate this in real-time, however the performance is slightly worst than the next approach.
    
    \item 2 stages aka ''Localization and Classification``. First, we detect the faces using an specific face detector as Retinaface \cite{deng2019retinaface}, which performs pixel-wise face localisation on various scales of faces. Then, each detected face is cropped and finally, we use a binary classification network trained to predict whether there is facemask (1) or not (0). Using this approach we need to train 2 models, the inference phase is more time-consuming since 2 models must infer predictions on the image, however, it has more accuracy than the first method. See pipeline in Figure \ref{fig:trump}.
\end{enumerate}

\begin{figure}[!h] 
\centering    
\includegraphics[width=\textwidth]{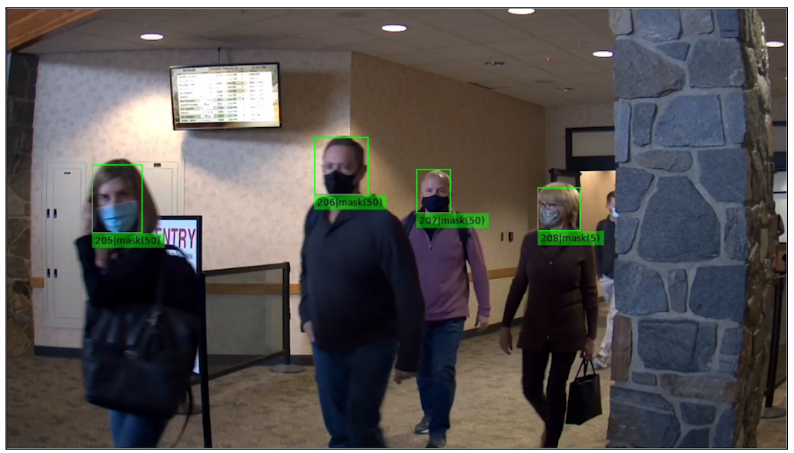}
\caption[FaceMask Detection using MASKCAM]{FaceMask Detection using MASKCAM}
\label{fig:facemask-detect}
\end{figure}

Braulio Rios and Evan Juras presented at the GTC 2021 the project ''A Mask-Detecting Smart Camera Using the Jetson Nano: The Developer Journey`` \footnote{\url{https://www.nvidia.com/en-us/on-demand/session/gtcspring21-s32588/}}. They discuss the developing process of MaskCam, a scalable and affordable Jetson Nano-based smart camera system that measures crowd face mask usage in real time, Figure \ref{fig:facemask-detect} shows MaskCam real-time results in an airport.
We use as baseline this model as baseline, as it provides great results running on the Jetson Nano Board \footnote{\url{https://github.com/bdtinc/maskcam}}. MASKCAM uses a YOLOv4-tiny model as single-shot object detector (the single model localizes and classifies multiple instances directly).
We create our own version of this model using the same YOLO-v4 implementation but training on more data and augmentations, unfortunately, there is no public benchmark to measure if our model is better than the baseline, however, our local classification accuracy is >0.95\%.

Related with this problem, we aim to compete at the ''Face Bio-metrics under COVID (Masked Face Recognition Challenge \& Workshop ICCV 2021)``. In order to achieve the best possible performance, we choose a 2-stage object detector or Localization and Classification approach.
We first classify the faces using Retinaface pre-trained model for face classification \footnote{\url{https://github.com/ternaus/retinaface}} by Vladimir Iglovikov, then following his approach, we crop each detected face and resize them into $64 \times 64$ images, the $N$ detected faces crops are classified by a second network.
In this approach the detector is called ''localizer`` and the second network, a deep CNN for binary classification in this case, is called ''classifier``. As a classifier typically we use ResNet \cite{resnet} or EfficientNet \cite{tan2020efficientnet} as SOTA networks for image classifications. In Figure \ref{fig:trump} we show a simple 2-stage pipeline.

As extension of this problem, and also useful, we can count the number of persons in the image, providing in real-time the number of persons in a room, plant, etc. This problem is usually called ''Crowd Counting `` \cite{oh2020crowd, shi2020realtime-crowd}, since it is not in the scope of this project, we simply assume that the number of detected faces is the number of persons in the place (i.e we assume there are no false positives).

\begin{figure}[!h] 
\centering    
\includegraphics[width=\textwidth]{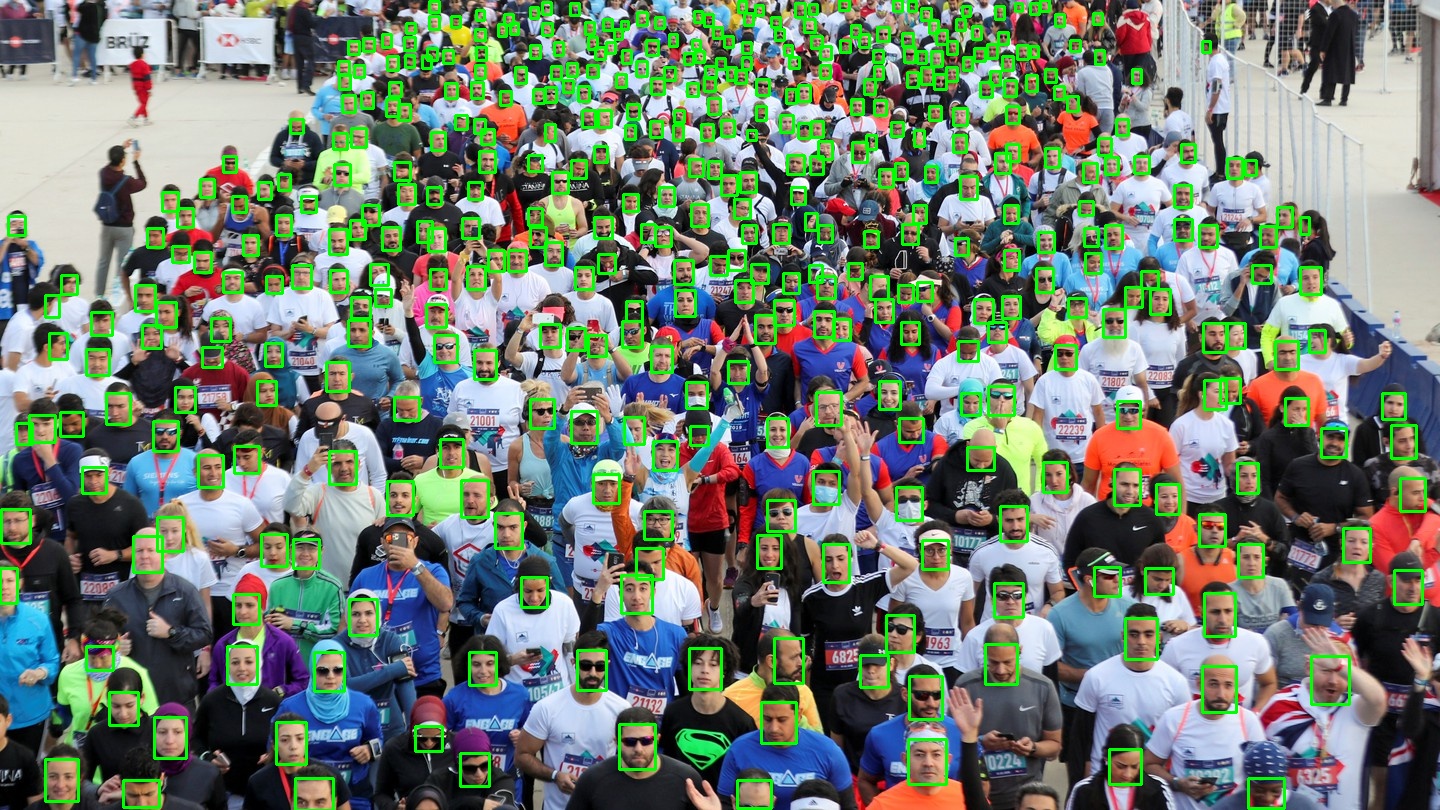}
\caption[Face Detection using Retinaface]{Face Detection using Retinaface implementation from Vladimir Iglovikov}
\label{fig:face-detect}
\end{figure}
\clearpage

\begin{figure}[!h] 
\centering    
\includegraphics[width=0.8\textwidth]{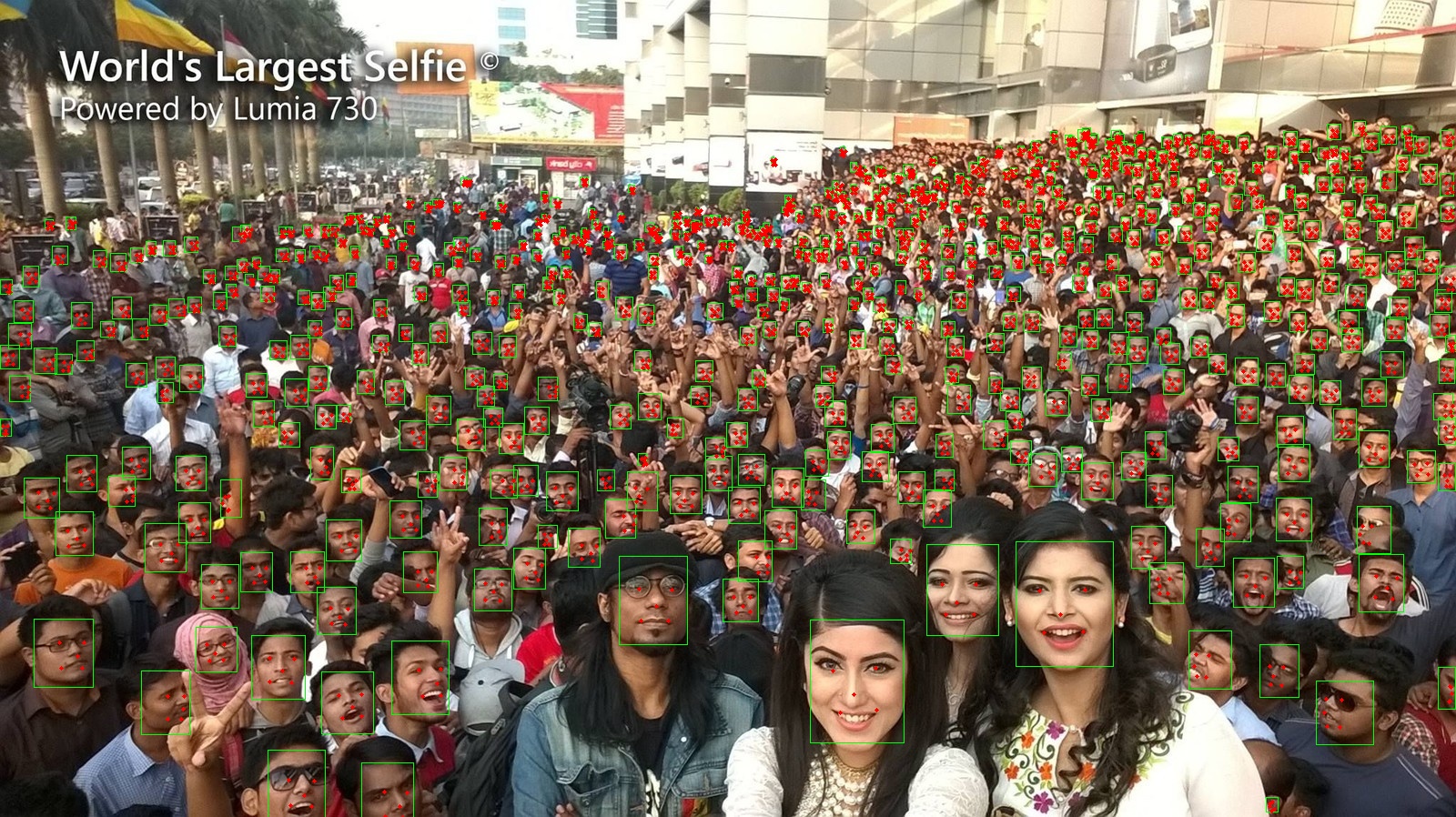}
\caption[CenterNet for face recognition]{CenterNet for face recognition.}
\label{fig:centerface}
\end{figure}

Note that we could have used the same object detector as we used for solving the autonomous driving problem in Section \ref{baidu}, CenterNet \cite{duan2019centernet}, which proves the versatility and generalization capabilities of these models if they are properly trained. The same model we used for solving a real-world self-driving car problem, can be used to solve the facemask recognition problem, Figure \ref{fig:centerface} shows an example of face classification using CenterNet \footnote{\url{https://github.com/nvlong21/Lightweight-face-detection-CenterNet}}.

\begin{figure}[!h] 
\centering    
\includegraphics[width=\textwidth]{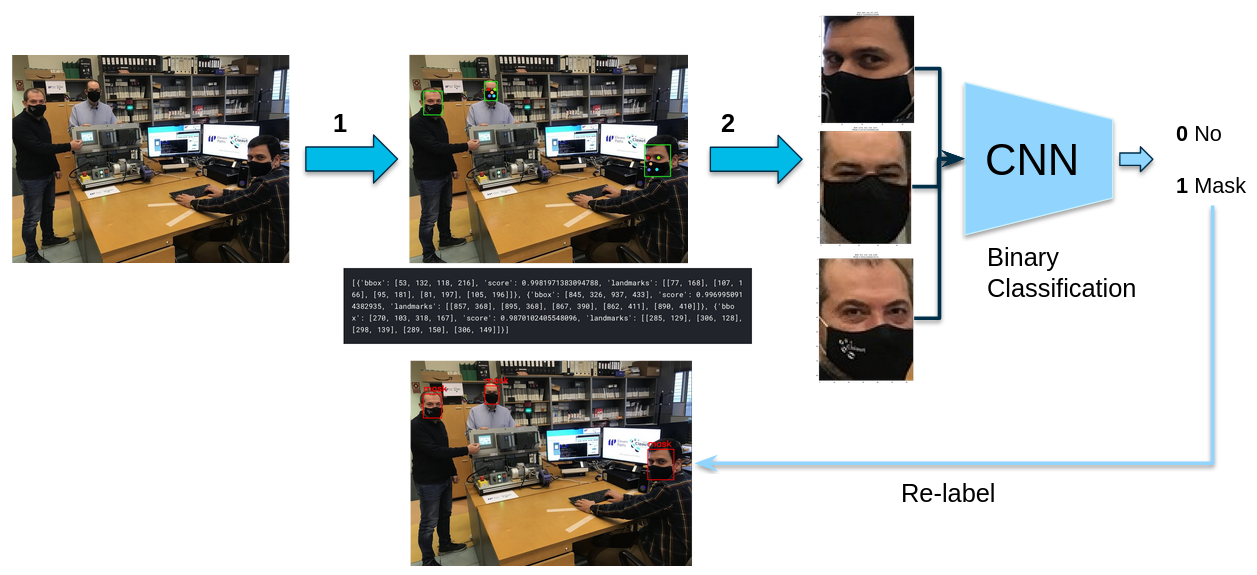}
\caption[Our 2-stage FaceMask Recognition Pipeline]{Our 2-stage FaceMask Recognition Pipeline. We use the proposed RetinaFace pre-trained detection model, and we train our own CNN (EfficientNet-B0~\cite{effnet}) as a Binary classifier that discriminates between ``mask'' and ``no mask'', using the ICCV 2021 FaceMask Dataset Challenge.}
\label{fig:our-pipe}
\end{figure}

\begin{figure}[!h] 
\centering    
\includegraphics[width=\textwidth]{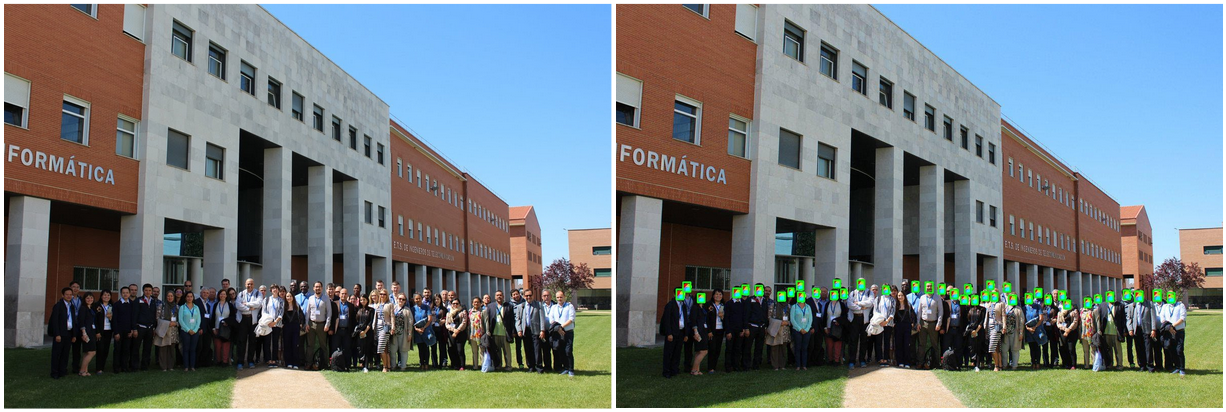}
\caption[Face-recognition Results]{Our Face Recognition results.}
\label{fig:our-pipe-results}
\end{figure}

\chapter{Conclusions and Future Work}  

\ifpdf
    \graphicspath{{Chapter7/Figs/Raster/}{Chapter7/Figs/PDF/}{Chapter7/Figs/}}
\else
    \graphicspath{{Chapter7/Figs/Vector/}{Chapter7/Figs/}}
\fi

\section{Discussion} 

Autonomous Driving and Navigation at every scale is a hard problem, even Elon Musk assumes this:

\begin{figure}[!h]
\includegraphics[width=\textwidth]{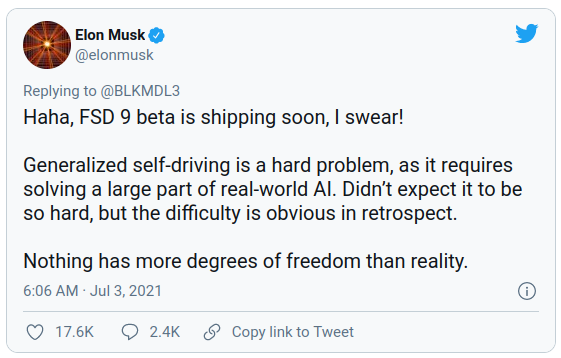}
\centering
\caption{Elon Musk twitter on Autonomous Driving.}
\label{fig:elon-ad}
\end{figure}

Academics who study AI and computer vision have long said that creating a fully self-driving (level 5) automobile would take many years, perhaps decades.

Currently, the approach that provides some decent level of autonomy (i.e. the driver does not require to do anything) assumes a constrained world with perfect illumination and weather, wide roads, etc; most companies do over-engineer here, using cars with expensive sensors as multiple LiDARs and other sophisticated sensors and mapping the entire ''world``. In this scenario, the system only needs to worry about moving objects (cars, pedestrians,etc.).

The \textbf{question} is: can a model trained in perfect conditions in California, on well-known mapped roads, generalize and work on cloudy non-delimited roads in Utah?

Some scientist like Yann LeCun believe the real solution to this problem is a combination of deep learning and model-based reinforcement learning \cite{kiran2021deep-rl-ad} that allows the system to adapt and overcome uncertainty.

George Hotz, CEO of \url{comma.ai}, understand this problem as 3 different problems:
\begin{itemize}
    \item Static driving problem: We assume we are the only car in the world-environment. This problem can be solved 100\% with mapping and localization using LiDAR sensors as localizers for constantly tracking our position. This problem is not realistic since assumes an unrealistic simplification of a constrained world.
    
    \item Dynamic: Multiple unpredictable agents and non-mapped conditions. We need to detect other cars, pedestrians, signals, etc in real-time and react properly. Note that many of those agents are not static, but dynamic.
    To solve this problem we need localization, mapping, and models about other agents behaviour (object detectors, motion predictors, etc).
    
    \item Counterfactual: Is an extension of the ''dynamic`` problem, here we consider our interaction and influence on the environment. This problem needs, besides the other solutions, deep reinforcement learning \cite{kiran2021deep-rl-ad, Folkers_2019-deep-rl-ad, osinski2020simulationbased-deep-rl-ad} to learn world dynamics and adapt.
\end{itemize}

At small-scale, we consider that the 3 main problems are: inference , deal with adversarial conditions, and lack of training data.

\paragraph{Inference}

Most of the developed methods for computer vision tasks like Object Detection and Segmentation are too complex and cannot be used in real-time applications.
A reasonable image resolution of $512\times512$ pixels at 25 FPS require very efficient memory and computation models \cite{kouris2019approximate-time-cons-ad,redmon2016look-yolo}. Regular GPUs cannot be used as regular embedded systems in small-scale robots, despite they are used in self-driving cars, its power consumption is not efficient and problematic.
Multi-task networks enable to perform different tasks using a single model, avoiding memory and IO bottlenecks in comparison with many different task-specific models running in parallel. 

A future alternative might be Binarized Neural Networks (BNNs) \cite{NIPS2016_d8330f85-bnn, courbariaux2016binarized-bnn}, this networks have binary weights (+1, -1) and activations at run-time. A normal neural network uses 32-bit floating point, but a binarized network uses 1-bit numbers (we encode +1 as 1, and -1 as 0), this way, memory size is reduced, and bitwise operations improve the power efficiency. BNNs can reduce power consumption by more than 27 times while keeping competitive perfomrance \cite{NIPS2016_d8330f85-bnn}. Some high-level tasks as object detection are starting to be solved using this efficient technology, an example is BiDet \cite{Wang_2020_CVPR-bidet}, a binarized neural network learning method for efficient object detection.

\paragraph{Adversarial Conditions}

Adverse weather and illumination conditions (e.g. fog, dust, rain, low light, nighttime and shadows) create visibility problems, without visual information this systems tend to collapse. Many outdoor applications such as autonomous cars and delivery robots are required to operate securely on adversarial scenarios (bad weather, dark areas, occluded sensor, etc.). While progress is being made in this direction, the performance of current vision algorithms is still mainly benchmarked under ''utopic`` conditions like good weather and favorable lighting, this is one of main reasons why models fail to generalize and adapt to edge or adversarial scenarios. Even the top-performing models suffer a severe performance degradation under adverse conditions as Hendrycks \textit{et.al.} proved \cite{hendrycks2019benchmarking-robustness}.

A possible solution to this problem will be based on strong data-augmentation techniques and richer data simulation. Heavy tail and edge events can help us to explain model's fragility, and thus, improve its performance.

\begin{figure}[!h]
\includegraphics[width=\textwidth]{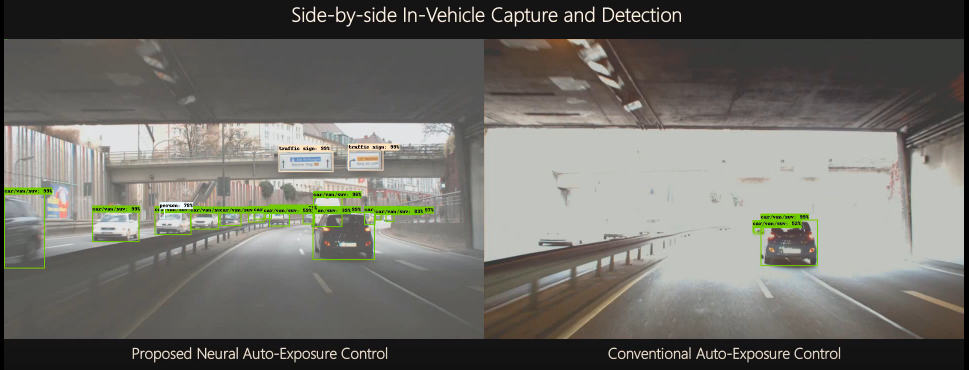}
\centering
\caption[Autonomous Driving Adversarial Contidions]{Example of an adversarial condition for an Autonomous Vehicle \cite{onzon2021neural}. The camera sensor is overexposed, therefore the system loses spatial information and there is an object detection performance degradation.}
\label{fig:adversarial1}
\end{figure}

\begin{figure}[!h]
\includegraphics[width=\textwidth]{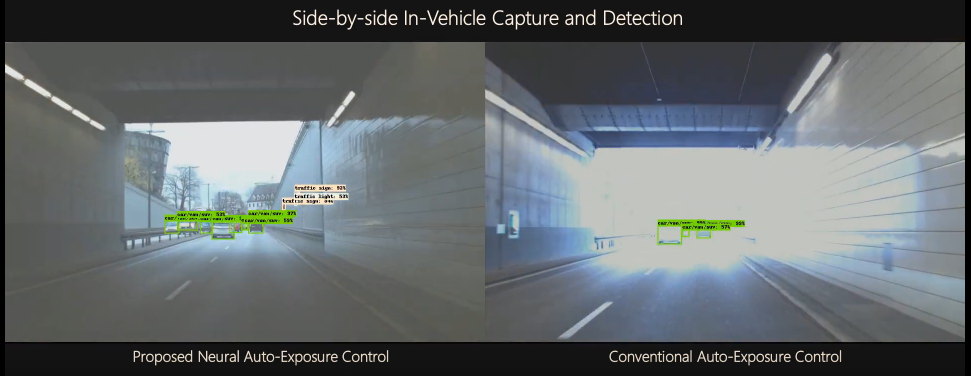}
\centering
\caption[]{Example of an adversarial condition (camera sensor overexposed) for an Autonomous Vehicle \cite{onzon2021neural}.}
\label{fig:adversarial2}
\end{figure}

\begin{figure}[!h]
\includegraphics[width=\textwidth]{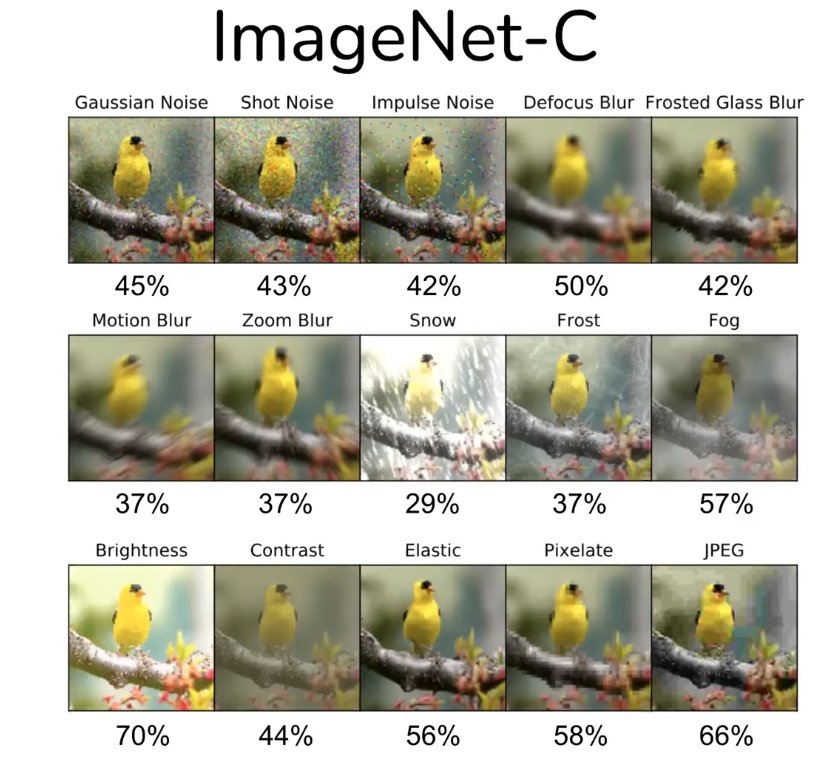}
\centering
\caption[Example of performance degradation at adversarial conditions]{A base ResNet-50 \cite{resnet} network performance degradation during test-time classification depending on the corruption \cite{hendrycks2019benchmarking-robustness}.}
\label{fig:adversarial}
\end{figure}

\paragraph{Scarce Training Data}

Andrej Karpathy, Director of AI at Tesla, mentioned at CVPR 2021 that Tesla has 3,000,000,000 miles of data collected using their AutoPilot, this is traduced into: 1.5 petabytes of data, 6 billion labeled objects, 1 million 10s videos (8 cameras, 36 FPS) including wide variety of scenarios. Yet, this is not enough to create a full self-driving car. Either small-scale or real-scale autonomous robots, require: (i) large amount of data (videos), (ii) clean and labeled data, (iii) diversity, edge cases are key.

Capturing data is a time-consuming and expensive process, on the other hand, models trained on simulators and synthetic data usually fail to generalize on real-world scenarios \cite{Dosovitskiy17-carla, Li_2019-aads-simulator, savva2017minos-indoor-simulator, Li_2019-aads-simulator}.
Indeed, the most expensive part is data labelling, which is done by third-party companies, for this reason, this problem can be solved using auto-labelling; models trained with historic data generate offline predictions new captured data, the predictions from this model ensemble are used are labels (ground-truth). 

Note that large enough and efficient neural networks are needed to train on such amount of data properly.

\begin{figure}[]
\includegraphics[width=\textwidth]{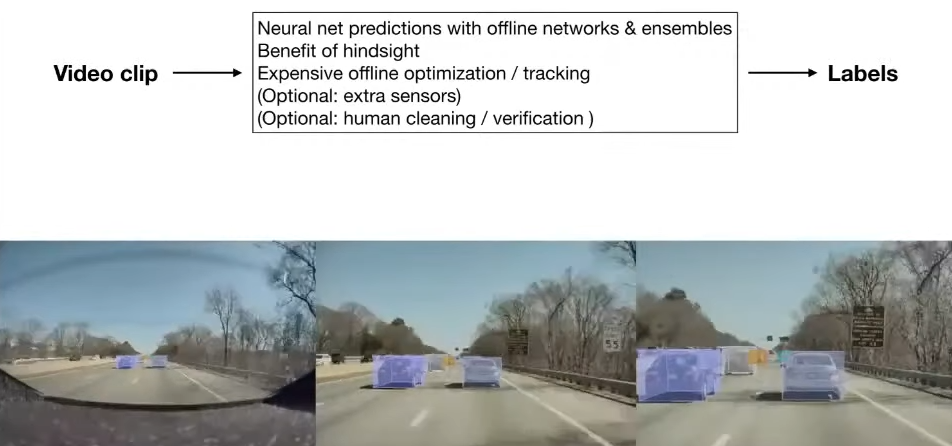}
\centering
\caption[Tesla Auto-Labelling technique]{Tesla Auto-Labelling technique presented by Andrej Karpathy at CVPR'21 WAD.}
\label{fig:tesla-autolabel}
\end{figure}


\clearpage

\section{Future Work} 

We would like to maintain the robotic solutions (JetBot or JetRover) and improve the AI solution by: (i) increasing the number of tasks the robot can do, (ii) use new technologies to improve existing tasks (i.e. object detection).
We believe that multi-task networks are the best option to add new tasks and improve efficiency.\\

In the field of \textbf{motion}, we can explore recent work for learning adaptable policies for legged and wheeled robots \cite{song2020rapidly, Magana_2019_fast_adapt, Luo2020ModelIA-jetson}, this is crucial for robots to operate autonomously in complex and quickly changing environments. 
The idea proposed by Kumar \textit{et.al.} Rapid Motor Adaptation (RMA) \cite{kumar2021rma} allows legged robots to adapt in real-time to unseen scenarios like changing terrains, changing payloads, deformable surfaces, weather, etc. 
This new RMA algorithm is trained completely in simulation without using any domain knowledge like reference trajectories and is easy to deploy on the robot. Despite the paper is focused on legged robots, this approach can be useful for delivery robots and other kinds of wheeled robots, and thus, our robot would be able to adapt its speed and traction depending on the scenario in the warehouse or industry plant (slippery, grass, dust, concrete, sand, etc.)

Besides motion and other hardware improvements, we focus on the ''intelligence`` of the robot, we aim to improve its perception by developing new computer vision applications.

\paragraph{3D Understanding: Scene Reconstruction and 3D Detection.}

Reconstructing the 3D world around us is a long standing goal in computer vision research, for this reason, 3D reconstruction \cite{monocular-scene-reconst-2009, indoor-reconst-2015, 3d-recons-eccv-2020, murez2020atlas} is considered one of the most fundamental tasks for visual perception systems. The problem definition is: to reconstruct a complete and accurate 3D environment structure and geometry from only images (2D observations) captured by conventional monocular RGB cameras.
This task is very challenging from a mathematical point of view, the visual input (RGB image) is the result of a projection mapping from the 3D world onto a lower-dimensional space, a 2D image, and this mapping is not bijective, it cannot be inverted. 
Other approaches reconstruct scenes by fusing depth measurements from special sensors like RGB-D cameras or LiDAR, into 3D models \cite{indoor-reconst-2015}. Despite these sensors are very effective, we do not use this approach because they require special  hardware and more computing power, making them more expensive.\\


\begin{figure}[]
\includegraphics[width=\textwidth]{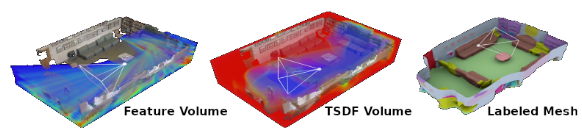}
\centering
\caption{Atlas \cite{murez2020atlas} method overview.}
\label{fig:atlas}
\end{figure}

\begin{figure}[]
\includegraphics[width=\textwidth]{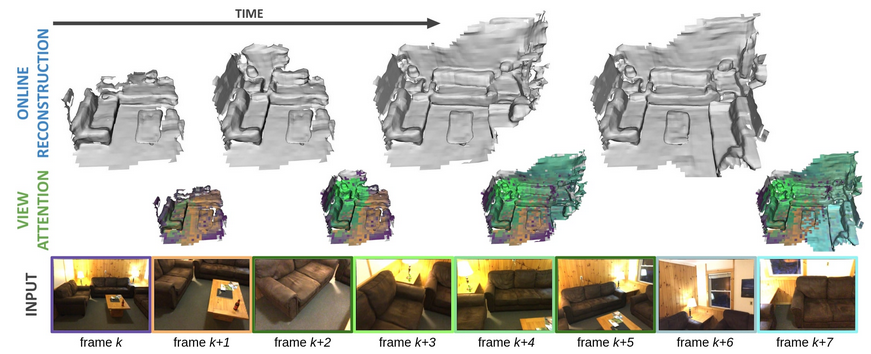}
\centering
\caption{TransformerFusion \cite{2021transformerfusion} reconstructs scenes online taking a RGB video as input.}
\label{fig:3d-transformer}
\end{figure}

\begin{figure}[]
\includegraphics[width=\textwidth]{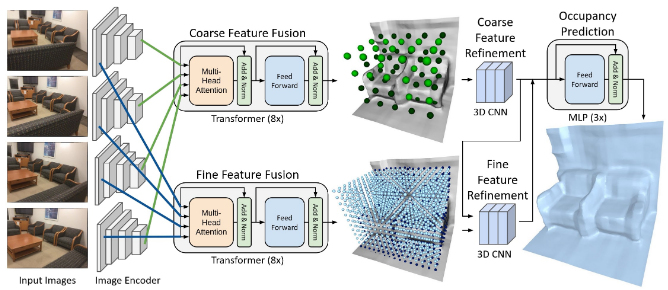}
\centering
\caption{TransformerFusion \cite{2021transformerfusion} method overview using a transformer architecture \cite{transformers-2017}.}
\label{fig:3d-transformer-method}
\end{figure}

State-of-the-art approaches use monocular cameras to take RGB images (single or multi). Denninger \textit{et.al.} \cite{3d-recons-eccv-2020} propose a novel approach to infer full volumetric reconstructions from a single viewport, even occluded areas. Yet, due to the model complexity, we cannot integrate it as a real-time application in the robot.

Atlas \cite{murez2020atlas} is an end-to-end method that can reconstruct scenes from a set of posed RGB images.
This approach performs a direct regression from 2D to 3D without relying on an intermediate representation of depth maps prior to estimating a full 3D model of a scene. They extract features from $N$ images with pose information using 2D CNNs, these features are back-projected and accumulated into a voxel volume using the camera pose estimation. After accumulating the $\mathcal{D}$-dimensional $N$ features, a 3D CNN refines the accumulated features and predicts the output. 
We consider to use this method offline for indoor scene reconstruction of small environments within the factory.

Most recent method, TransformerFusion \cite{2021transformerfusion}, from an input monocular RGB video (2D images stream), the video frames are processed by a Transformer network \cite{transformers-2017} that fuses the observations into a volumetric feature grid representing the scene; this feature grid is then decoded into an implicit 3D scene representation (see Figure \ref{fig:3d-transformer-method}). This approach outperforms state-of-the-art multi-view stereo depth estimation methods, fully-convolutional 3D reconstruction approaches, and LSTM-based recurrent networks for video sequence fusion. This approach is design for an online reconstruction (see Figure \ref{fig:3d-transformer}), adapting model complexity and input size (low-resolution images) will allow to use this model as a real-time application integrated in our robot.

\begin{figure}[t]
\includegraphics[width=\textwidth]{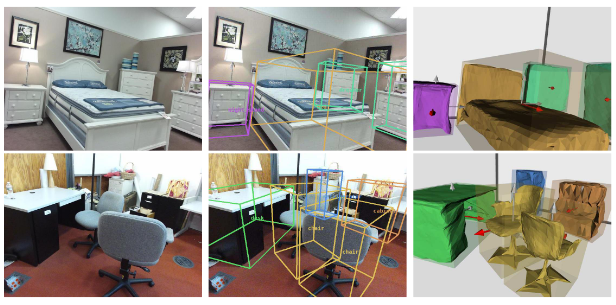}
\centering
\caption[Total3DUnderstanding]{Total3DUnderstanding \cite{nie2020total3dunderstanding} predicts, from a single image (left), the contextual knowledge as the camera pose and 3D object bounding boxes (middle) and reconstruct object meshes (right).}
\label{fig:total3d}
\end{figure}

Integrating this technology into the robot would allow us to reconstruct 3D indoor environments (e.g. warehouses, factory plants, etc.) from RGB images. Nowadays, this is a useful feature as many applications such as autonomous driving and augmented reality, rely heavily upon accurate 3D reconstructions of the surrounding environment.
In addition, geometric scene reconstructions form the basis for 3D scene understanding, supporting tasks such as 3D object detection, semantic, and instance segmentation \cite{nie2020total3dunderstanding} (See Figure \ref{fig:total3d}).\\

The most common 3D task in the visual perception field is \textbf{3D Object Detection} \cite{Chen_2016_CVPR-3DOD_AD, mousavian20173d, liu2019deep, qin2020monogrnet, weng2019monocular, ku2019monocular}.
In Section \ref{sec:object-detecttion} we explained the 2D case where models predict 4 degrees of freedom (DoF) axis-aligned bounding boxes with center (x, y) and size (w, h). This idea can be extended to 3D bounding boxes, which usually have have 7 DoF: 3D physical size (w, h, l), 3D center location (x, y, z) and yaw. Alternatively, 6DoF pose can 

\begin{figure}[t]
\includegraphics[width=\textwidth]{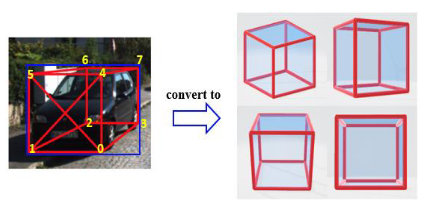}
\centering
\caption{4 example viewpoints for the projection of 3D cuboid \cite{fang20193d}.}
\label{fig:2d-3d-projection}
\end{figure}

Both, 2D and 3D Object Detectors take as input a single RGB image (typically captured using a monocular camera). However, extending 2D priors to 3D predictions is a very challenging task because it is fundamentally ill-posed as the critical depth information is not present in the RGB image (see Figure \ref{fig:2d-3d-projection}).

\begin{figure}[t]
\includegraphics[width=\textwidth]{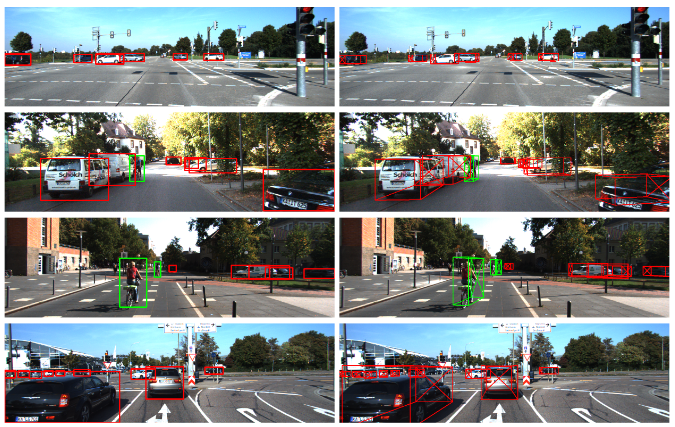}
\centering
\caption[3D Object Detection example]{Examples of the 2D detection boxes and the estimated 3D projections, (red) cars and (green) cyclists. \cite{mousavian20173d}}
\label{fig:3d-od-ad}
\end{figure}

By extending prediction to 3D, we can capture size, position and orientation in the world, leading to a variety of applications in robotics, self-driving vehicles, and augmented reality. Some of these applications are shown in Figures \ref{fig:total3d}, \ref{fig:3d-od-ad}.

\begin{figure}[]
\includegraphics[width=\textwidth]{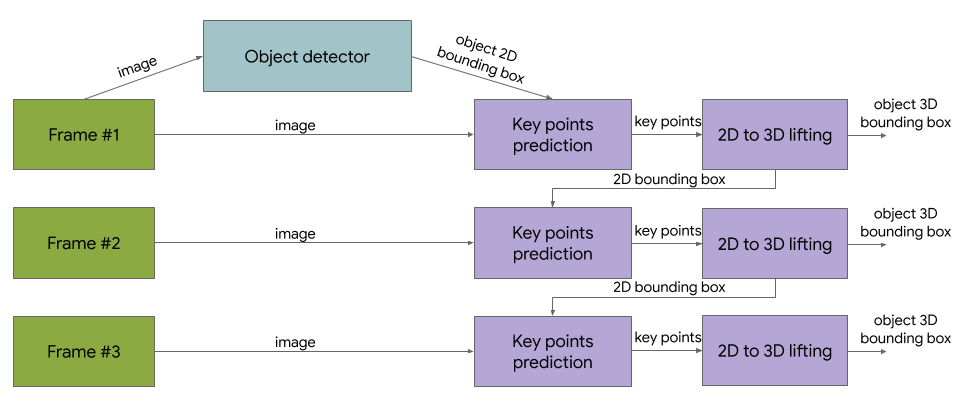}
\centering
\caption{MediaPipe Objectron 2-stage Network Architecture \cite{objectron2021}.}
\label{fig:mediapipe-2stage}
\end{figure}

\begin{figure}[]
\includegraphics[width=\textwidth]{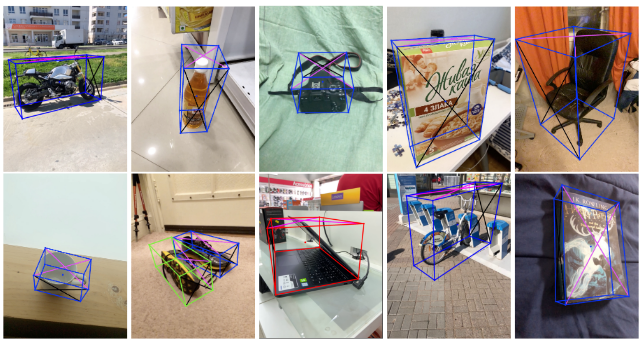}
\centering
\caption{MediaPipe Objectron Dataset examples \cite{objectron2021}.}
\label{fig:mediapipe-viz}
\end{figure}

We are experimenting how to integrate in our robot the MediaPipe Objectron \cite{objectron2021}, a mobile real-time 3D object detection pipeline. This pipeline is single-stage detector model that predicts the pose and physical size of an object from a single RGB image (See Figure \ref{fig:mediapipe-viz}) The model backbone has an encoder-decoder architecture, built upon MobileNetv2 \cite{howard2017mobilenets, sandler2019mobilenetv2}. We choose this model because it is light enough to run real-time on mobile devices at 20-30 FPS and high-quality images.
MediaPipe also proposes a two-stage pipeline illustrated in Figure \ref{fig:mediapipe-2stage}. The first stage uses a conventional 2D object detector to find the 2D bounding boxes of the objects. The second stage takes each detected object (box) and estimates the 3D bounding box. At the same time, the information from the $i$-frame is re-used to predict the $i+1$-frame, such that the object detector does not need to run every frame.

\begin{figure}[!h]
\includegraphics[width=\textwidth]{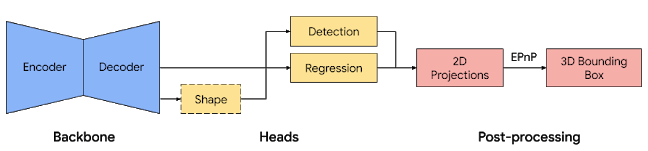}
\centering
\caption{MediaPipe Objectron Network Architecture \cite{objectron2021}.}
\label{fig:mediapipe-net}
\end{figure}

\clearpage
Additionally, we also started to investigate LiDAR sensors for Raspberry Pi and Jetson Nano, the integration of radar sensor onto the pure visual-based system would allow to better localize and map indoor environments, for instance, using SLAM technology (Simultaneous localization and mapping) \cite{tateno2017cnnslam, chen2020survey-slam, bloesch2019codeslam, Czarnowski_2020-deepfactors-slam} for constructing or updating a map of an unknown environment while simultaneously keeping track of our agent's location within it.

\begin{figure}[!h]
\includegraphics[width=\textwidth]{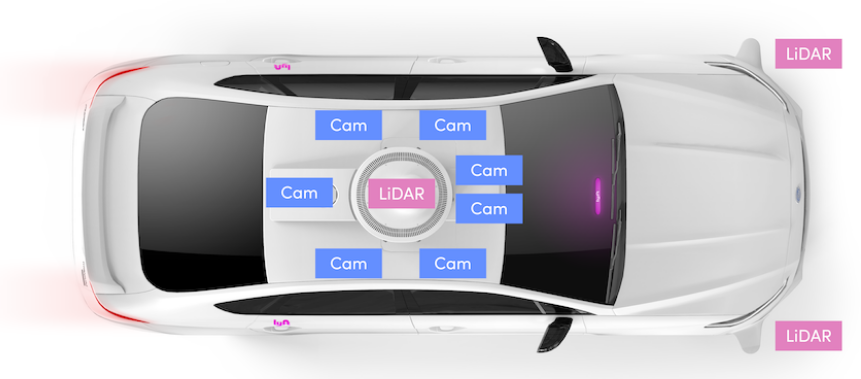}
\centering
\caption{Lyft's car sensor distribution.}
\label{fig:lyft-car}
\end{figure}

\begin{figure}[!h]
\includegraphics[trim={0 0 0 14cm },clip, width=0.8\textwidth]{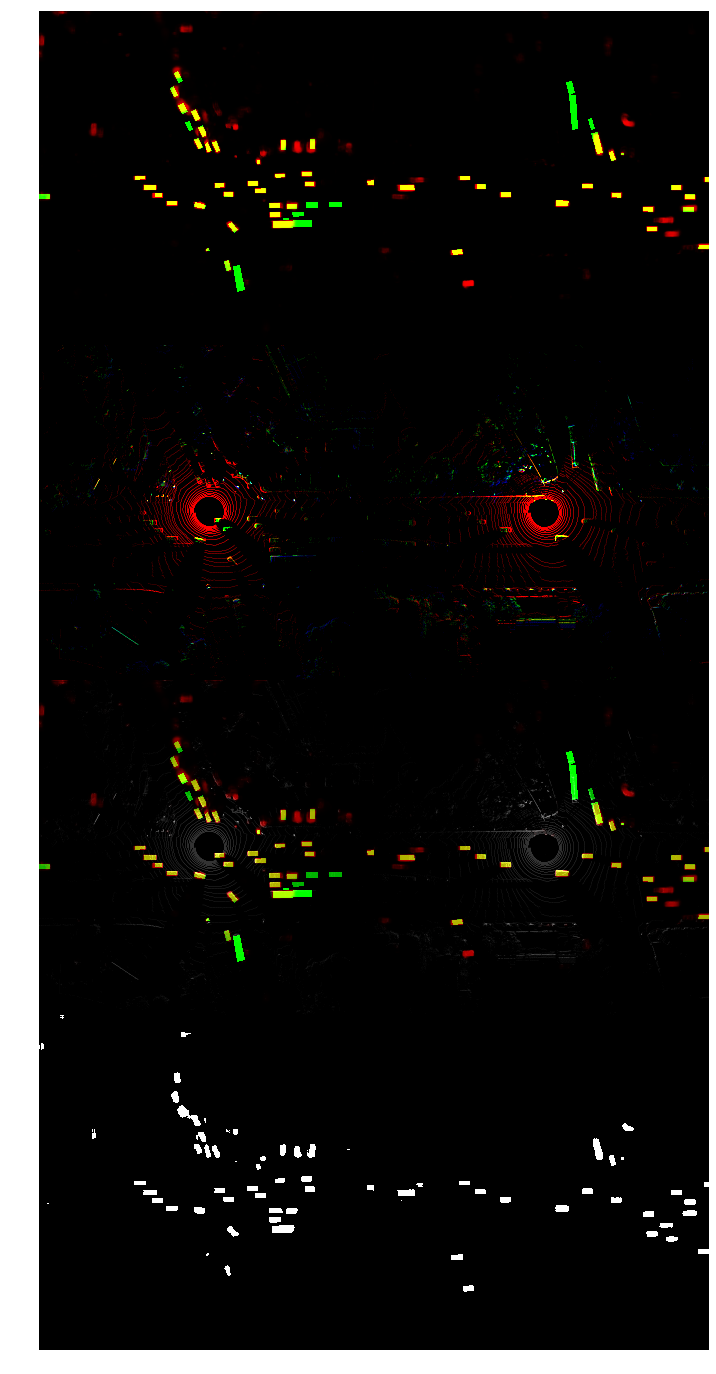}
\centering
\caption[3D Object Detection on LiDAR pointcloud]{We show the predictions for 2 different timestamps of the same scene (left and right). From top to bottom: Pointcloud from the LiDAR sensor, predicted cars (yellow) and ground-truth (green), in white color the predictions (boxes where the model predicts a car).}
\label{fig:lyft-preds}
\end{figure}

Our first experiment consisted on 3D detection on dense point clouds captured using LiDAR. We used Lyft \cite{lyft-dataset}, nuScenes and KITTI datasets~\cite{Caesar_2020_CVPR-nuscenes, Geiger2013IJRRkitti} for autonomous driving as they are common benchmarks for testing and developing 3D detection models. Our solution was ranked 38th of 546 teams at the ''Lyft 3D Object Detection for Autonomous Vehicles Challenge 2019`` \footnote{\url{https://www.kaggle.com/c/3d-object-detection-for-autonomous-vehicles/}}, held at NeurIPS 2019. The following figures are some data visualizations from the self-driving car cameras and LiDAR sensor's point clouds. We show in Figure~\ref{fig:lyft-car} the autonomous vehicle's sensor distribution. Figure~\ref{fig:lyft-preds} shows the detected cars in the pointcloud using our model (yellow) and the gound-truth, the real position of the cars (green).

We train a custom U-Net~\cite{ronneberger2015unet} fully convolutional neural network to predict whether a car or other object is present for every pixel in a birds eye view of the world centered on the car. We can then threshold this probability map and fit boxes around each of the detections.
Train the model took 10hrs using a Tesla P100 GPU, with inference times under 30ms per image. We provide more details about the architecture and training at the supplementary material (e.g. code).

\begin{figure}[!h]
\includegraphics[width=\textwidth]{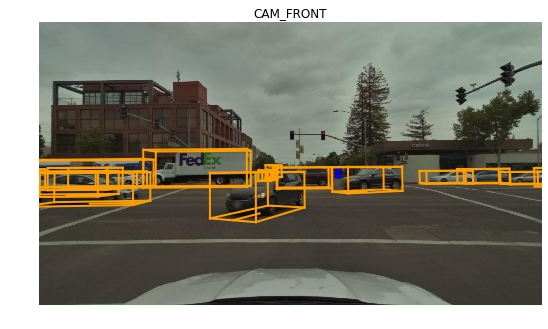}
\centering
\caption{Front-camera visualization with cars' 3D detection boxes.}
\label{fig:lyft-3dbox}
\end{figure}

\begin{figure}[!h]
\includegraphics[width=\textwidth]{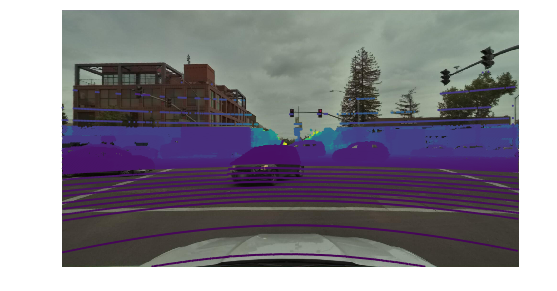}
\centering
\caption{Projection of the LiDAR PointCloud into front-camera image.}
\label{fig:lyft-cloud}
\end{figure}

\begin{figure}[!h]
\includegraphics[width=\textwidth]{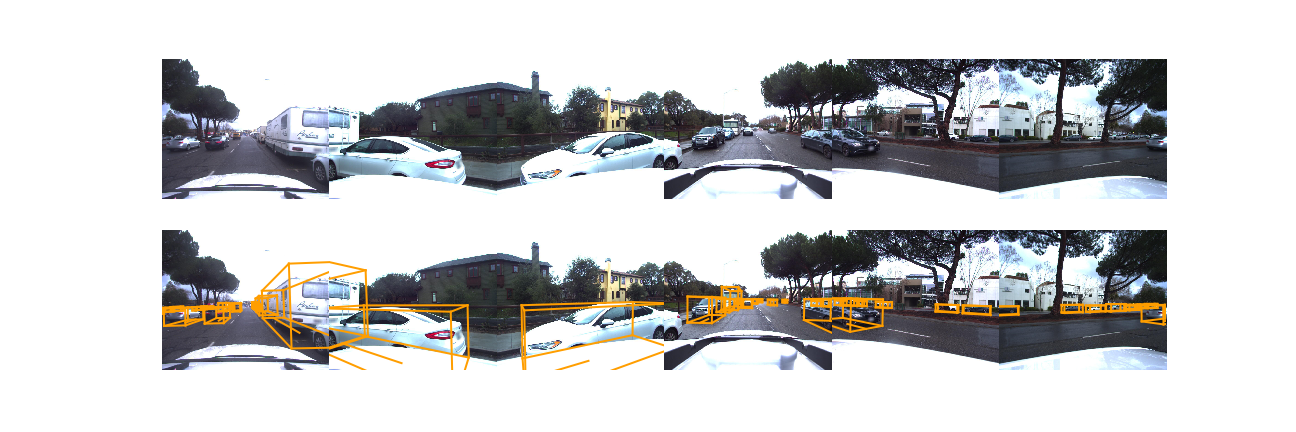}
\centering
\caption{Multi-view from the different camera sensors.}
\label{fig:lyft-cloud2}
\end{figure}

\begin{figure}[!h]
\includegraphics[width=\textwidth]{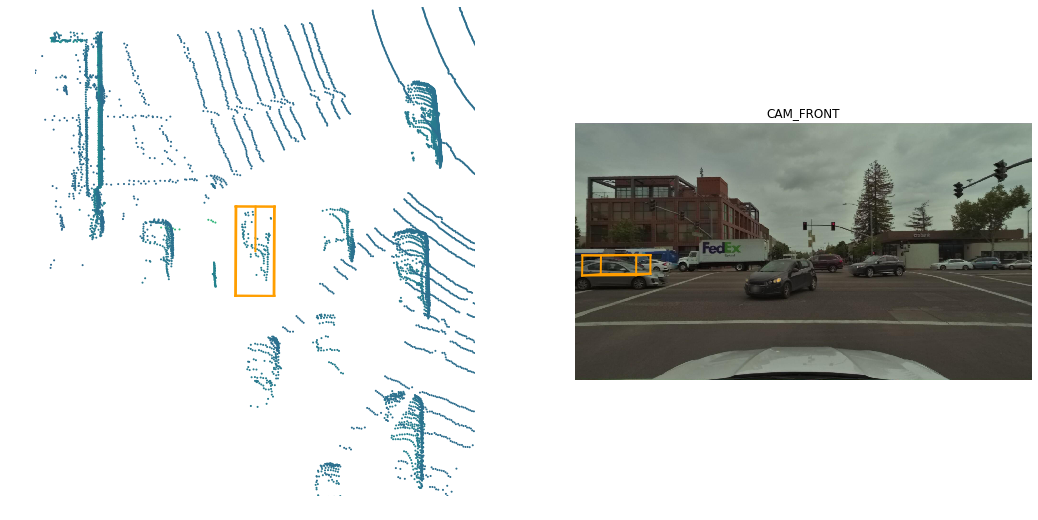}
\centering
\caption{LiDAR pointcloud annotation and corresponding front-camera 3D detection.}
\label{fig:lyft-cloud3}
\end{figure}

\begin{figure}[!h]
\includegraphics[width=\textwidth]{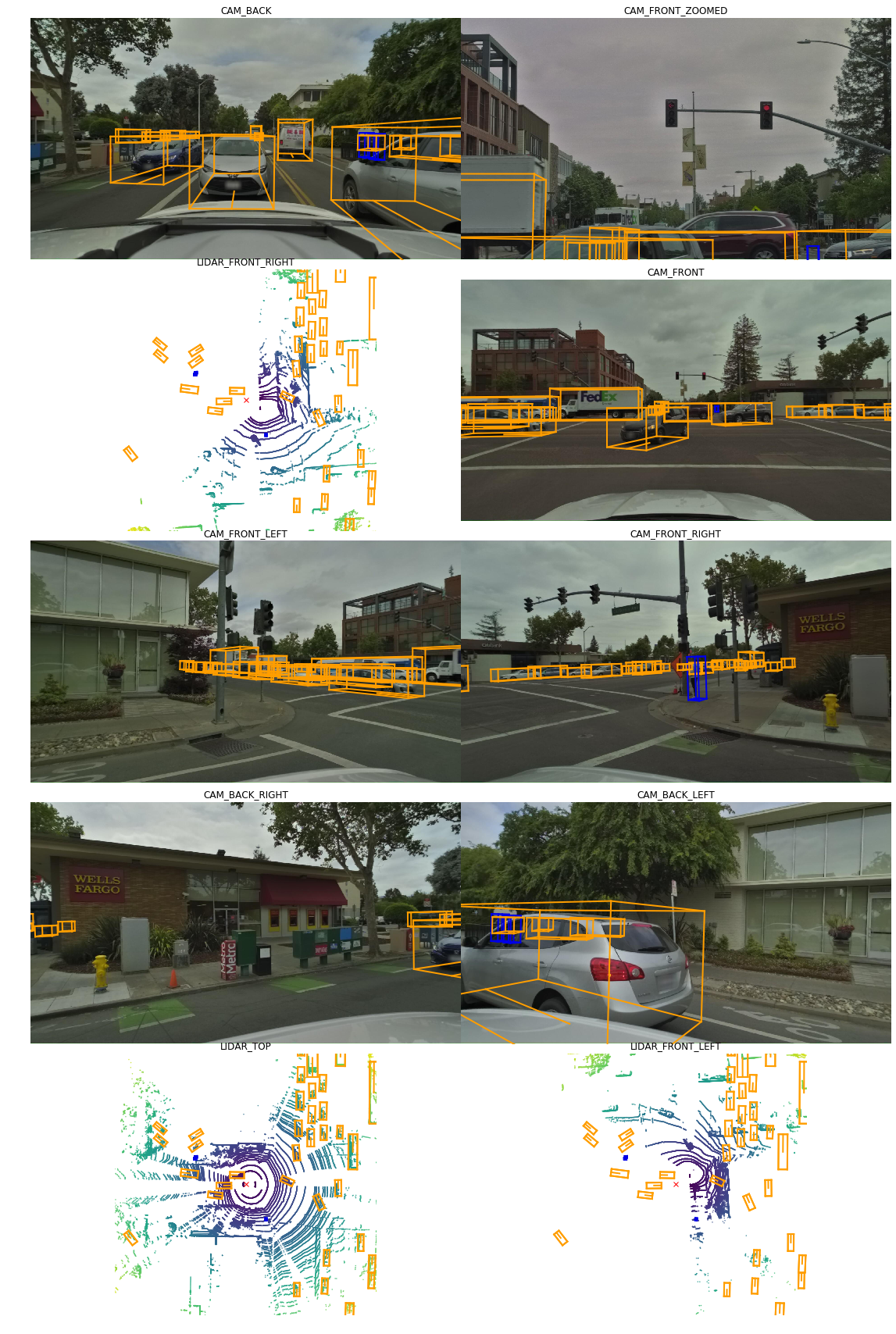}
\centering
\caption[Autonomous Driving annotated scene from Lyft]{Example of an annotated scene (6 cameras' pov, 3 radar maps) generated by us.}
\label{fig:lyft-comp}
\end{figure}

\clearpage

\section*{Conclusions} 

\epigraph{If you think of standardization as the best that you know today, but which is to be improved tomorrow; you get somewhere.}{Henry Ford}

The Industry acceleration needs of the innovative AI research contributions, this allows us to explore a new point of view about autonomous robots and their applications and impact in industrial environments, and consequently, in society. In order to develop vision-based deep learning solutions, this work has addressed experimentation, research and review of state-of-the-art deep learning models, and two partial objectives were proposed: (i) develop a prototype low-cost robotic solution and (ii) propose Industry solutions.
A starting robotic solution was developed from scratch and tested in laboratory, also industrial applications were proposed and modeled for their integration in real environments.
\\

To conclude, we believe this work has reviewed the state-of-the-art and the most important questions regarding autonomous robots and autonomous driving, we have explored the similarities between real-scale cars and small robots, and we have proposed some approaches to solve real-world problems.







{\small
\bibliographystyle{ieee_fullname}
\cleardoublepage
\bibliography{references}

\begin{thebibliography}{100}\itemsep=-1pt

\bibitem{objectron2021}
Adel Ahmadyan, Liangkai Zhang, Jianing Wei, Artsiom Ablavatski, and Matthias
  Grundmann.
\newblock Objectron: A large scale dataset of object-centric videos in the wild
  with pose annotations, 2020.

\bibitem{copilot}
Open AI.
\newblock {GitHub Copilot is powered by Codex, the new AI system created by
  OpenAI.}
\newblock \url{https://copilot.github.com/}, 2021.
\newblock [Online; accessed 19-June-2021].

\bibitem{bloesch2019codeslam}
Michael Bloesch, Jan Czarnowski, Ronald Clark, Stefan Leutenegger, and
  Andrew~J. Davison.
\newblock Codeslam - learning a compact, optimisable representation for dense
  visual slam, 2019.

\bibitem{bochkovskiy2020yolov4}
Alexey Bochkovskiy, Chien-Yao Wang, and Hong-Yuan~Mark Liao.
\newblock Yolov4: Optimal speed and accuracy of object detection, 2020.

\bibitem{2021transformerfusion}
Aljaž Božič, Pablo Palafox, Justus Thies, Angela Dai, and Matthias Nießner.
\newblock Transformerfusion: Monocular rgb scene reconstruction using
  transformers, 2021.

\bibitem{brown2020language-gpt3}
Tom~B. Brown, Benjamin Mann, Nick Ryder, Melanie Subbiah, Jared Kaplan,
  Prafulla Dhariwal, Arvind Neelakantan, Pranav Shyam, Girish Sastry, Amanda
  Askell, Sandhini Agarwal, Ariel Herbert-Voss, Gretchen Krueger, Tom Henighan,
  Rewon Child, Aditya Ramesh, Daniel~M. Ziegler, Jeffrey Wu, Clemens Winter,
  Christopher Hesse, Mark Chen, Eric Sigler, Mateusz Litwin, Scott Gray,
  Benjamin Chess, Jack Clark, Christopher Berner, Sam McCandlish, Alec Radford,
  Ilya Sutskever, and Dario Amodei.
\newblock Language models are few-shot learners, 2020.

\bibitem{Buslaev_2020-albus}
Alexander Buslaev, Vladimir~I. Iglovikov, Eugene Khvedchenya, Alex Parinov,
  Mikhail Druzhinin, and Alexandr~A. Kalinin.
\newblock Albumentations: Fast and flexible image augmentations.
\newblock {\em Information}, 11(2):125, Feb 2020.

\bibitem{Caesar_2020_CVPR-nuscenes}
Holger Caesar, Varun Bankiti, Alex~H. Lang, Sourabh Vora, Venice~Erin Liong,
  Qiang Xu, Anush Krishnan, Yu Pan, Giancarlo Baldan, and Oscar Beijbom.
\newblock nuscenes: A multimodal dataset for autonomous driving.
\newblock In {\em Proceedings of the IEEE/CVF Conference on Computer Vision and
  Pattern Recognition (CVPR)}, June 2020.

\bibitem{chen2020survey-slam}
Changhao Chen, Bing Wang, Chris~Xiaoxuan Lu, Niki Trigoni, and Andrew Markham.
\newblock A survey on deep learning for localization and mapping: Towards the
  age of spatial machine intelligence, 2020.

\bibitem{Chen_2016_CVPR-3DOD_AD}
Xiaozhi Chen, Kaustav Kundu, Ziyu Zhang, Huimin Ma, Sanja Fidler, and Raquel
  Urtasun.
\newblock Monocular 3d object detection for autonomous driving.
\newblock In {\em Proceedings of the IEEE Conference on Computer Vision and
  Pattern Recognition (CVPR)}, June 2016.

\bibitem{indoor-reconst-2015}
Sungjoon Choi, Qian-Yi Zhou, and Vladlen Koltun.
\newblock Robust reconstruction of indoor scenes.
\newblock In {\em 2015 IEEE Conference on Computer Vision and Pattern
  Recognition (CVPR)}, pages 5556--5565, 2015.

\bibitem{courbariaux2016binarized-bnn}
Matthieu Courbariaux, Itay Hubara, Daniel Soudry, Ran El-Yaniv, and Yoshua
  Bengio.
\newblock Binarized neural networks: Training deep neural networks with weights
  and activations constrained to +1 or -1, 2016.

\bibitem{Czarnowski_2020-deepfactors-slam}
Jan Czarnowski, Tristan Laidlow, Ronald Clark, and Andrew~J. Davison.
\newblock Deepfactors: Real-time probabilistic dense monocular slam.
\newblock {\em IEEE Robotics and Automation Letters}, 5(2):721–728, Apr 2020.

\bibitem{deng2019retinaface}
Jiankang Deng, Jia Guo, Yuxiang Zhou, Jinke Yu, Irene Kotsia, and Stefanos
  Zafeiriou.
\newblock Retinaface: Single-stage dense face localisation in the wild, 2019.

\bibitem{3d-recons-eccv-2020}
Maximilian Denninger and Rudolph Triebel.
\newblock 3d scene reconstruction from a single viewport.
\newblock In Andrea Vedaldi, Horst Bischof, Thomas Brox, and Jan-Michael Frahm,
  editors, {\em Computer Vision -- ECCV 2020}, pages 51--67, Cham, 2020.
  Springer International Publishing.

\bibitem{Dosovitskiy17-carla}
Alexey Dosovitskiy, German Ros, Felipe Codevilla, Antonio Lopez, and Vladlen
  Koltun.
\newblock {CARLA}: {An} open urban driving simulator.
\newblock In {\em Proceedings of the 1st Annual Conference on Robot Learning},
  pages 1--16, 2017.

\bibitem{duan2019centernet}
Kaiwen Duan, Song Bai, Lingxi Xie, Honggang Qi, Qingming Huang, and Qi Tian.
\newblock Centernet: Keypoint triplets for object detection, 2019.

\bibitem{dvornik2019importance-augs}
Nikita Dvornik, Julien Mairal, and Cordelia Schmid.
\newblock On the importance of visual context for data augmentation in scene
  understanding, 2019.

\bibitem{spotmini}
Boston Dynamics.
\newblock {SpotMini Robot}.
\newblock \url{https://www.bostondynamics.com/spot}, 2021.
\newblock [Online; accessed 19-June-2021].

\bibitem{monocular-scene-reconst-2009}
Erik Einhorn, Christof Schröter, and Horst-Michael Gross.
\newblock Monocular scene reconstruction for reliable obstacle detection and
  robot navigation.
\newblock pages 7--12, 01 2009.

\bibitem{fang20193d}
Jiaojiao Fang, Lingtao Zhou, and Guizhong Liu.
\newblock 3d bounding box estimation for autonomous vehicles by cascaded
  geometric constraints and depurated 2d detections using 3d results, 2019.

\bibitem{fedus2021switch-language-trill}
William Fedus, Barret Zoph, and Noam Shazeer.
\newblock Switch transformers: Scaling to trillion parameter models with simple
  and efficient sparsity, 2021.

\bibitem{florensa2018automatic-goal-rl}
Carlos Florensa, David Held, Xinyang Geng, and Pieter Abbeel.
\newblock Automatic goal generation for reinforcement learning agents, 2018.

\bibitem{Folkers_2019-deep-rl-ad}
Andreas Folkers, Matthias Rick, and Christof Buskens.
\newblock Controlling an autonomous vehicle with deep reinforcement learning.
\newblock {\em 2019 IEEE Intelligent Vehicles Symposium (IV)}, Jun 2019.

\bibitem{Fran_ois_Lavet_2018_intro_rl}
Vincent François-Lavet, Peter Henderson, Riashat Islam, Marc~G. Bellemare, and
  Joelle Pineau.
\newblock An introduction to deep reinforcement learning.
\newblock {\em Foundations and Trends® in Machine Learning},
  11(3-4):219–354, 2018.

\bibitem{Geiger2013IJRRkitti}
Andreas Geiger, Philip Lenz, Christoph Stiller, and Raquel Urtasun.
\newblock Vision meets robotics: The kitti dataset.
\newblock {\em International Journal of Robotics Research (IJRR)}, 2013.

\bibitem{Goodfellow-et-al-2016-deeplearning}
Ian Goodfellow, Yoshua Bengio, and Aaron Courville.
\newblock {\em Deep Learning}.
\newblock MIT Press, 2016.
\newblock \url{http://www.deeplearningbook.org}.

\bibitem{resnet}
Kaiming He, Xiangyu Zhang, Shaoqing Ren, and Jian Sun.
\newblock Deep residual learning for image recognition, 2015.

\bibitem{he2018bagtricks}
Tong He, Zhi Zhang, Hang Zhang, Zhongyue Zhang, Junyuan Xie, and Mu Li.
\newblock Bag of tricks for image classification with convolutional neural
  networks, 2018.

\bibitem{hendrycks2019benchmarking-robustness}
Dan Hendrycks and Thomas Dietterich.
\newblock Benchmarking neural network robustness to common corruptions and
  perturbations, 2019.

\bibitem{Hermann2015-industry40}
Mario Hermann, Tobias Pentek, and Boris Otto.
\newblock {Design Principles for Industrie 4.0 Scenarios: A Literature Review}.
\newblock {\em Technische Universitat Dortmund}, 1(1):4--16, 2015.

\bibitem{Hern_ndez_Garc_a_2018-augs}
Alex Hernández-García and Peter König.
\newblock Further advantages of data augmentation on convolutional neural
  networks.
\newblock {\em Lecture Notes in Computer Science}, page 95–103, 2018.

\bibitem{howard2017mobilenets}
Andrew~G. Howard, Menglong Zhu, Bo Chen, Dmitry Kalenichenko, Weijun Wang,
  Tobias Weyand, Marco Andreetto, and Hartwig Adam.
\newblock Mobilenets: Efficient convolutional neural networks for mobile vision
  applications, 2017.

\bibitem{NIPS2016_d8330f85-bnn}
Itay Hubara, Matthieu Courbariaux, Daniel Soudry, Ran El-Yaniv, and Yoshua
  Bengio.
\newblock Binarized neural networks.
\newblock In D. Lee, M. Sugiyama, U. Luxburg, I. Guyon, and R. Garnett,
  editors, {\em Advances in Neural Information Processing Systems}, volume~29.
  Curran Associates, Inc., 2016.

\bibitem{oh2020crowd}
Min hwan Oh, Peder~A. Olsen, and Karthikeyan~Natesan Ramamurthy.
\newblock Crowd counting with decomposed uncertainty, 2020.

\bibitem{ioffe2015batch}
Sergey Ioffe and Christian Szegedy.
\newblock Batch normalization: Accelerating deep network training by reducing
  internal covariate shift, 2015.

\bibitem{Jiao_2019-objectdetc-survey}
Licheng Jiao, Fan Zhang, Fang Liu, Shuyuan Yang, Lingling Li, Zhixi Feng, and
  Rong Qu.
\newblock A survey of deep learning-based object detection.
\newblock {\em IEEE Access}, 7:128837–128868, 2019.

\bibitem{jouppi2017indatacenter-tpu}
Norman~P. Jouppi, Cliff Young, Nishant Patil, David Patterson, Gaurav Agrawal,
  Raminder Bajwa, Sarah Bates, Suresh Bhatia, Nan Boden, Al Borchers, Rick
  Boyle, Pierre luc Cantin, Clifford Chao, Chris Clark, Jeremy Coriell, Mike
  Daley, Matt Dau, Jeffrey Dean, Ben Gelb, Tara~Vazir Ghaemmaghami, Rajendra
  Gottipati, William Gulland, Robert Hagmann, C.~Richard Ho, Doug Hogberg, John
  Hu, Robert Hundt, Dan Hurt, Julian Ibarz, Aaron Jaffey, Alek Jaworski,
  Alexander Kaplan, Harshit Khaitan, Andy Koch, Naveen Kumar, Steve Lacy, James
  Laudon, James Law, Diemthu Le, Chris Leary, Zhuyuan Liu, Kyle Lucke, Alan
  Lundin, Gordon MacKean, Adriana Maggiore, Maire Mahony, Kieran Miller, Rahul
  Nagarajan, Ravi Narayanaswami, Ray Ni, Kathy Nix, Thomas Norrie, Mark
  Omernick, Narayana Penukonda, Andy Phelps, Jonathan Ross, Matt Ross, Amir
  Salek, Emad Samadiani, Chris Severn, Gregory Sizikov, Matthew Snelham, Jed
  Souter, Dan Steinberg, Andy Swing, Mercedes Tan, Gregory Thorson, Bo Tian,
  Horia Toma, Erick Tuttle, Vijay Vasudevan, Richard Walter, Walter Wang, Eric
  Wilcox, and Doe~Hyun Yoon.
\newblock In-datacenter performance analysis of a tensor processing unit, 2017.

\bibitem{996-sobel}
N. Kanopoulos, N. Vasanthavada, and R.L. Baker.
\newblock Design of an image edge detection filter using the sobel operator.
\newblock {\em IEEE Journal of Solid-State Circuits}, 23(2):358--367, 1988.

\bibitem{mit-racecar-icra-2019}
Sertac Karaman, Owen Guldner, Abhishek Agarwal, and Winter Guerra.
\newblock {MIT RACECAR at ICRA 2019}.
\newblock \url{https://mit-racecar.github.io/}, 2019.
\newblock [Online Available; accessed 19-June-2021].

\bibitem{lyft-dataset}
R. Kesten, M. Usman, J. Houston, T. Pandya, K. Nadhamuni, A. Ferreira, M. Yuan,
  B. Low, A. Jain, P. Ondruska, S. Omari, S. Shah, A. Kulkarni, A. Kazakova, C.
  Tao, L. Platinsky, W. Jiang, and V. Shet.
\newblock Level 5 perception dataset 2020.
\newblock \url{https://level-5.global/level5/data/}, 2019.

\bibitem{kingma2017adam}
Diederik~P. Kingma and Jimmy Ba.
\newblock Adam: A method for stochastic optimization, 2017.

\bibitem{kiran2021deep-rl-ad}
B~Ravi Kiran, Ibrahim Sobh, Victor Talpaert, Patrick Mannion, Ahmad A.~Al
  Sallab, Senthil Yogamani, and Patrick Pérez.
\newblock Deep reinforcement learning for autonomous driving: A survey, 2021.

\bibitem{sensor-fusion-ad}
Jelena Kocić, Nenad Jovičić, and Vujo Drndarević.
\newblock Sensors and sensor fusion in autonomous vehicles.
\newblock In {\em 2018 26th Telecommunications Forum (TELFOR)}, pages 420--425,
  2018.

\bibitem{kouris2019approximate-time-cons-ad}
Alexandros Kouris, Stylianos~I. Venieris, Michail Rizakis, and Christos-Savvas
  Bouganis.
\newblock Approximate lstms for time-constrained inference: Enabling fast
  reaction in self-driving cars, 2019.

\bibitem{imagenet}
Alex Krizhevsky, Ilya Sutskever, and Geoffrey~E. Hinton.
\newblock Imagenet classification with deep convolutional neural networks.
\newblock NIPS'12, page 1097–1105, Red Hook, NY, USA, 2012. Curran Associates
  Inc.

\bibitem{imagenet-classification-cnn}
Alex Krizhevsky, Ilya Sutskever, and Geoffrey~E. Hinton.
\newblock Imagenet classification with deep convolutional neural networks.
\newblock {\em Commun. ACM}, 60(6):84–90, May 2017.

\bibitem{ku2019monocular}
Jason Ku, Alex~D. Pon, and Steven~L. Waslander.
\newblock Monocular 3d object detection leveraging accurate proposals and shape
  reconstruction, 2019.

\bibitem{kumar2021rma}
Ashish Kumar, Zipeng Fu, Deepak Pathak, and Jitendra Malik.
\newblock Rma: Rapid motor adaptation for legged robots, 2021.

\bibitem{LeCun1999-cnns}
Yann LeCun, Patrick Haffner, L{\'e}on Bottou, and Yoshua Bengio.
\newblock {\em Object Recognition with Gradient-Based Learning}, pages
  319--345.
\newblock Springer Berlin Heidelberg, Berlin, Heidelberg, 1999.

\bibitem{Li_2019-aads-simulator}
W. Li, C.~W. Pan, R. Zhang, J.~P. Ren, Y.~X. Ma, J. Fang, F.~L. Yan, Q.~C.
  Geng, X.~Y. Huang, H.~J. Gong, and et al.
\newblock Aads: Augmented autonomous driving simulation using data-driven
  algorithms.
\newblock {\em Science Robotics}, 4(28):eaaw0863, Mar 2019.

\bibitem{lin2015microsoft-coco-dataset}
Tsung-Yi Lin, Michael Maire, Serge Belongie, Lubomir Bourdev, Ross Girshick,
  James Hays, Pietro Perona, Deva Ramanan, C.~Lawrence Zitnick, and Piotr
  Dollár.
\newblock Microsoft coco: Common objects in context, 2015.

\bibitem{liu2019deep}
Lijie Liu, Jiwen Lu, Chunjing Xu, Qi Tian, and Jie Zhou.
\newblock Deep fitting degree scoring network for monocular 3d object
  detection, 2019.

\bibitem{Liu_2016-ssd-detector}
Wei Liu, Dragomir Anguelov, Dumitru Erhan, Christian Szegedy, Scott Reed,
  Cheng-Yang Fu, and Alexander~C. Berg.
\newblock Ssd: Single shot multibox detector.
\newblock {\em Lecture Notes in Computer Science}, page 21–37, 2016.

\bibitem{Luo2020ModelIA-jetson}
Yan Luo, Abdeslam Boularias, and Mridul Aanjaneya.
\newblock Model identification and control of a low-cost wheeled mobile robot
  using differentiable physics.
\newblock {\em ArXiv}, abs/2009.11465, 2020.

\bibitem{Magana_2019_fast_adapt}
Octavio Antonio~Villarreal Magana, Victor Barasuol, Marco Camurri, Luca
  Franceschi, Michele Focchi, Massimiliano Pontil, Darwin~G. Caldwell, and
  Claudio Semini.
\newblock Fast and continuous foothold adaptation for dynamic locomotion
  through cnns.
\newblock {\em IEEE Robotics and Automation Letters}, 4(2):2140–2147, Apr
  2019.

\bibitem{ml-methodologies}
D. Michie.
\newblock {Methodologies from Machine Learning in Data Analysis and Software}.
\newblock {\em The Computer Journal}, 34(6):559--565, 12 1991.

\bibitem{mousavian20173d}
Arsalan Mousavian, Dragomir Anguelov, John Flynn, and Jana Kosecka.
\newblock 3d bounding box estimation using deep learning and geometry, 2017.

\bibitem{murez2020atlas}
Zak Murez, Tarrence van As, James Bartolozzi, Ayan Sinha, Vijay Badrinarayanan,
  and Andrew Rabinovich.
\newblock Atlas: End-to-end 3d scene reconstruction from posed images, 2020.

\bibitem{muller2021openbot-jetson}
Matthias Müller and Vladlen Koltun.
\newblock Openbot: Turning smartphones into robots, 2021.

\bibitem{andrewng-plan}
Andrew Ng.
\newblock {Andrew Ng’s 5 Step Framework to Plan AI Projects Effectively}.
\newblock \url{https://read.deeplearning.ai/the-batch/issue-81/}, 2021.
\newblock [Online; accessed 19-June-2021].

\bibitem{nie2020total3dunderstanding}
Yinyu Nie, Xiaoguang Han, Shihui Guo, Yujian Zheng, Jian Chang, and Jian~Jun
  Zhang.
\newblock Total3dunderstanding: Joint layout, object pose and mesh
  reconstruction for indoor scenes from a single image, 2020.

\bibitem{onzon2021neural}
Emmanuel Onzon, Fahim Mannan, and Felix Heide.
\newblock Neural auto-exposure for high-dynamic range object detection.
\newblock In {\em Proceedings of the IEEE/CVF Conference on Computer Vision and
  Pattern Recognition}, 2021.

\bibitem{osinski2020simulationbased-deep-rl-ad}
Błażej Osiński, Adam Jakubowski, Piotr Miłoś, Paweł Zięcina, Christopher
  Galias, Silviu Homoceanu, and Henryk Michalewski.
\newblock Simulation-based reinforcement learning for real-world autonomous
  driving, 2020.

\bibitem{duckietown-icra}
Liam Paull, Jacopo Tani, Heejin Ahn, Javier Alonso-Mora, Luca Carlone, Michal
  Cap, Yu~Fan Chen, Changhyun Choi, Jeff Dusek, Yajun Fang, Daniel Hoehener,
  Shih-Yuan Liu, Michael Novitzky, Igor~Franzoni Okuyama, Jason Pazis, Guy
  Rosman, Valerio Varricchio, Hsueh-Cheng Wang, Dmitry Yershov, Hang Zhao,
  Michael Benjamin, Christopher Carr, Maria Zuber, Sertac Karaman, Emilio
  Frazzoli, Domitilla Del~Vecchio, Daniela Rus, Jonathan How, John Leonard, and
  Andrea Censi.
\newblock Duckietown: An open, inexpensive and flexible platform for autonomy
  education and research.
\newblock In {\em 2017 IEEE International Conference on Robotics and Automation
  (ICRA)}, pages 1497--1504, 2017.

\bibitem{qin2020monogrnet}
Zengyi Qin, Jinglu Wang, and Yan Lu.
\newblock Monogrnet: A geometric reasoning network for monocular 3d object
  localization, 2020.

\bibitem{redmon2016look-yolo}
Joseph Redmon, Santosh Divvala, Ross Girshick, and Ali Farhadi.
\newblock You only look once: Unified, real-time object detection, 2016.

\bibitem{ren2016faster}
Shaoqing Ren, Kaiming He, Ross Girshick, and Jian Sun.
\newblock Faster r-cnn: Towards real-time object detection with region proposal
  networks, 2016.

\bibitem{ronneberger2015unet}
Olaf Ronneberger, Philipp Fischer, and Thomas Brox.
\newblock U-net: Convolutional networks for biomedical image segmentation,
  2015.

\bibitem{Rumelhart1986LearningRB-backprop}
D. Rumelhart, Geoffrey~E. Hinton, and Ronald~J. Williams.
\newblock Learning representations by back-propagating errors.
\newblock {\em Nature}, 323:533--536, 1986.

\bibitem{ai-modern-approach-russell}
Stuart Russell and Peter Norvig.
\newblock {\em Artificial Intelligence: A Modern Approach}.
\newblock Prentice Hall Press, USA, 3rd edition, 2009.

\bibitem{sandler2019mobilenetv2}
Mark Sandler, Andrew Howard, Menglong Zhu, Andrey Zhmoginov, and Liang-Chieh
  Chen.
\newblock Mobilenetv2: Inverted residuals and linear bottlenecks, 2019.

\bibitem{savva2017minos-indoor-simulator}
Manolis Savva, Angel~X. Chang, Alexey Dosovitskiy, Thomas Funkhouser, and
  Vladlen Koltun.
\newblock Minos: Multimodal indoor simulator for navigation in complex
  environments, 2017.

\bibitem{sazanovich2020imitation-ducky}
Mikita Sazanovich, Konstantin Chaika, Kirill Krinkin, and Aleksei Shpilman.
\newblock Imitation learning approach for ai driving olympics trained on
  real-world and simulation data simultaneously, 2020.

\bibitem{shi2020realtime-crowd}
Xiaowen Shi, Xin Li, Caili Wu, Shuchen Kong, Jing Yang, and Liang He.
\newblock A real-time deep network for crowd counting, 2020.

\bibitem{article-augs-2019}
Connor Shorten and Taghi Khoshgoftaar.
\newblock A survey on image data augmentation for deep learning.
\newblock {\em Journal of Big Data}, 6, 07 2019.

\bibitem{vgg}
Karen Simonyan and Andrew Zisserman.
\newblock Very deep convolutional networks for large-scale image recognition,
  2015.

\bibitem{singhani2019realtime-freespace}
Anish Singhani.
\newblock Real-time freespace segmentation on autonomous robots for detection
  of obstacles and drop-offs, 2019.

\bibitem{song2019apollocar3d}
Xibin Song, Peng Wang, Dingfu Zhou, Rui Zhu, Chenye Guan, Yuchao Dai, Hao Su,
  Hongdong Li, and Ruigang Yang.
\newblock Apollocar3d: A large 3d car instance understanding benchmark for
  autonomous driving.
\newblock In {\em Proceedings of the IEEE Conference on Computer Vision and
  Pattern Recognition}, pages 5452--5462, 2019.

\bibitem{song2020rapidly}
Xingyou Song, Yuxiang Yang, Krzysztof Choromanski, Ken Caluwaerts, Wenbo Gao,
  Chelsea Finn, and Jie Tan.
\newblock Rapidly adaptable legged robots via evolutionary meta-learning, 2020.

\bibitem{srinivasa2019mushr}
Siddhartha~S. Srinivasa, Patrick Lancaster, Johan Michalove, Matt Schmittle,
  Colin Summers, Matthew Rockett, Joshua~R. Smith, Sanjiban Chouhury,
  Christoforos Mavrogiannis, and Fereshteh Sadeghi.
\newblock {MuSHR}: A low-cost, open-source robotic racecar for education and
  research.
\newblock {\em CoRR}, abs/1908.08031, 2019.

\bibitem{JMLR:v15:srivastava14a-dropout}
Nitish Srivastava, Geoffrey Hinton, Alex Krizhevsky, Ilya Sutskever, and Ruslan
  Salakhutdinov.
\newblock Dropout: A simple way to prevent neural networks from overfitting.
\newblock {\em Journal of Machine Learning Research}, 15(56):1929--1958, 2014.

\bibitem{tan2018survey-transfer}
Chuanqi Tan, Fuchun Sun, Tao Kong, Wenchang Zhang, Chao Yang, and Chunfang Liu.
\newblock A survey on deep transfer learning, 2018.

\bibitem{effnet}
Mingxing Tan and Quoc Le.
\newblock {E}fficient{N}et: Rethinking model scaling for convolutional neural
  networks.
\newblock In Kamalika Chaudhuri and Ruslan Salakhutdinov, editors, {\em
  Proceedings of the 36th International Conference on Machine Learning},
  volume~97 of {\em Proceedings of Machine Learning Research}, pages
  6105--6114. PMLR, 09--15 Jun 2019.

\bibitem{tan2020efficientnet}
Mingxing Tan and Quoc~V. Le.
\newblock Efficientnet: Rethinking model scaling for convolutional neural
  networks, 2020.

\bibitem{tani2020integrated-ducky}
Jacopo Tani, Andrea~F. Daniele, Gianmarco Bernasconi, Amaury Camus, Aleksandar
  Petrov, Anthony Courchesne, Bhairav Mehta, Rohit Suri, Tomasz Zaluska,
  Matthew~R. Walter, Emilio Frazzoli, Liam Paull, and Andrea Censi.
\newblock Integrated benchmarking and design for reproducible and accessible
  evaluation of robotic agents, 2020.

\bibitem{duckitown}
Jacopo Tani, Liam Paull, Maria~T. Zuber, Daniela Rus, Jonathan How, John
  Leonard, and Andrea Censi.
\newblock Duckietown: An innovative way to teach autonomy.
\newblock In Dimitris Alimisis, Michele Moro, and Emanuele Menegatti, editors,
  {\em Educational Robotics in the Makers Era}, pages 104--121, Cham, 2017.
  Springer International Publishing.

\bibitem{tateno2017cnnslam}
Keisuke Tateno, Federico Tombari, Iro Laina, and Nassir Navab.
\newblock Cnn-slam: Real-time dense monocular slam with learned depth
  prediction, 2017.

\bibitem{transformers-2017}
Ashish Vaswani, Noam Shazeer, Niki Parmar, Jakob Uszkoreit, Llion Jones,
  Aidan~N. Gomez, undefinedukasz Kaiser, and Illia Polosukhin.
\newblock Attention is all you need.
\newblock In {\em Proceedings of the 31st International Conference on Neural
  Information Processing Systems}, NIPS'17, page 6000–6010, Red Hook, NY,
  USA, 2017. Curran Associates Inc.

\bibitem{wang20196pack-6dof-ad}
Chen Wang, Roberto Martín-Martín, Danfei Xu, Jun Lv, Cewu Lu, Li Fei-Fei,
  Silvio Savarese, and Yuke Zhu.
\newblock 6-pack: Category-level 6d pose tracker with anchor-based keypoints,
  2019.

\bibitem{wang2020depthnet}
Linda Wang, Mahmoud Famouri, and Alexander Wong.
\newblock Depthnet nano: A highly compact self-normalizing neural network for
  monocular depth estimation, 2020.

\bibitem{wang2019benchmarking}
Yu~Emma Wang, Gu-Yeon Wei, and David Brooks.
\newblock Benchmarking tpu, gpu, and cpu platforms for deep learning, 2019.

\bibitem{Wang_2020_CVPR-bidet}
Ziwei Wang, Ziyi Wu, Jiwen Lu, and Jie Zhou.
\newblock Bidet: An efficient binarized object detector.
\newblock In {\em Proceedings of the IEEE/CVF Conference on Computer Vision and
  Pattern Recognition (CVPR)}, June 2020.

\bibitem{weng2019monocular}
Xinshuo Weng and Kris Kitani.
\newblock Monocular 3d object detection with pseudo-lidar point cloud, 2019.

\bibitem{6dvnet}
Di Wu, Zhaoyong Zhuang, Canqun Xiang, Wenbin Zou, and Xia Li.
\newblock 6d-vnet: End-to-end 6dof vehicle pose estimation from monocular rgb
  images.
\newblock In {\em 2019 IEEE/CVF Conference on Computer Vision and Pattern
  Recognition Workshops (CVPRW)}, pages 1238--1247, 2019.

\bibitem{xu2015empirical-relu}
Bing Xu, Naiyan Wang, Tianqi Chen, and Mu Li.
\newblock Empirical evaluation of rectified activations in convolutional
  network, 2015.

\bibitem{zhang2020empowering-jetson}
Jing Zhang and Dacheng Tao.
\newblock Empowering things with intelligence: A survey of the progress,
  challenges, and opportunities in artificial intelligence of things, 2020.

\bibitem{zhang2019bagtricks}
Zhi Zhang, Tong He, Hang Zhang, Zhongyue Zhang, Junyuan Xie, and Mu Li.
\newblock Bag of freebies for training object detection neural networks, 2019.

\bibitem{Zhong2017-industry40}
Ray~Y. Zhong, Xun Xu, Eberhard Klotz, and Stephen~T. Newman.
\newblock {Intelligent Manufacturing in the Context of Industry 4.0: A Review}.
\newblock {\em Engineering}, 3(5):616--630, 2017.

\bibitem{zhou2019objects}
Xingyi Zhou, Dequan Wang, and Philipp Krähenbühl.
\newblock Objects as points, 2019.

\bibitem{zhuang2020comprehensive-transfer}
Fuzhen Zhuang, Zhiyuan Qi, Keyu Duan, Dongbo Xi, Yongchun Zhu, Hengshu Zhu, Hui
  Xiong, and Qing He.
\newblock A comprehensive survey on transfer learning, 2020.

\bibitem{zilly2019ai-nips-ducky}
Julian Zilly, Jacopo Tani, Breandan Considine, Bhairav Mehta, Andrea~F.
  Daniele, Manfred Diaz, Gianmarco Bernasconi, Claudio Ruch, Jan Hakenberg,
  Florian Golemo, A.~Kirsten Bowser, Matthew~R. Walter, Ruslan Hristov, Sunil
  Mallya, Emilio Frazzoli, Andrea Censi, and Liam Paull.
\newblock The ai driving olympics at neurips 2018, 2019.

\bibitem{zoph2019learning-augs}
Barret Zoph, Ekin~D. Cubuk, Golnaz Ghiasi, Tsung-Yi Lin, Jonathon Shlens, and
  Quoc~V. Le.
\newblock Learning data augmentation strategies for object detection, 2019.

\end{thebibliography}
}





\afterpage{\blankpage}
\afterpage{\blankpage}

\end{document}